\documentclass[11pt]{article}
\usepackage{amstext,amssymb,amsmath,epsf,graphics}

\newtheorem{thm}{Theorem}
\newtheorem{lemma}{Lemma}
\newtheorem{coro}{Corollary}

\newcommand{\elementwise}[1]{\mbox{\scriptsize$\circ{#1}$}}
\newcommand{\elementwisen}[1]{\mbox{$\circ{#1}$}}
\newcommand{\kmea}[1]{\mbox{$|\!\!\wr{#1}\wr\!\!|$}}
\newcommand{\kmeas}[1]{\mbox{$|\!\!\wr{#1}\wr\!\!|_k$}}
\newcommand{\kmeask}[1]{\mbox{$|\!\!\wr{#1}\wr\!\!|_k^k$}}
\newcommand{\kmeasq}[1]{\mbox{$|\!\!\wr{#1}\wr\!\!|_k^q$}}

\newcommand{\sgn}{\mbox{${\rm sgn}$}}

\renewcommand{\geq}{\geqslant}
\renewcommand{\leq}{\leqslant}

\newcommand{\bm}[1]{\boldsymbol #1}
\newcommand{\bepsilon}{\boldsymbol \epsilon}

\newcommand{\balpha}{\boldsymbol{\alpha}}
\newcommand{\bbeta}{\boldsymbol{\beta}}

\newcommand{\Real}{\mbox{$\mathbb{R}$}}

\begin{document}

\title{\bf Deterministic Stretchy Regression}

\author{\textbf{\small Kar-Ann Toh$^1$, Lei Sun$^2$ and Zhiping Lin$^3$ }\\
{\footnotesize $^1$School of Electrical and Electronic Engineering}\\
{\footnotesize Yonsei University, Seoul, Korea\ 03722} \\
{\footnotesize $^2$School of Information and Electronics} \\
{\footnotesize Beijing Institute of Technology Beijing, PR China, 100081} \\
{\footnotesize $^3$School of Electrical and Electronic Engineering} \\
{\footnotesize Nanyang Technological University, Singapore 639798} \\
{\footnotesize Emails: $^1$katoh@yonsei.ac.kr,
$^2$sunlei@bit.edu.cn, and $^3$ezplin@ntu.edu.sg} }

\date{\footnotesize Drafted August 2015, Revised September 2017}
\maketitle

\begin{abstract}
An extension of the regularized least-squares in which the estimation parameters are
stretchable is introduced and studied in this paper. The solution of this ridge
regression with stretchable parameters is given in primal and dual spaces and in
closed-form. Essentially, the proposed solution stretches the covariance computation by a
power term, thereby compressing or amplifying the estimation parameters. To maintain the
computation of power root terms within the real space, an input transformation is
proposed. The results of an empirical evaluation in both synthetic and real-world data
illustrate that the proposed method is effective for compressive learning with
high-dimensional data.
\end{abstract}

\section{Introduction}\label{sec:introduction}

\subsection{Background}

\label{page1}

In the past few decades, linear prediction models were seen to be among the most popular
choices for data fitting and pattern recognition. They remain as part of the most
important tools {in various scientific applications today}. {Consider a set of $M$
training observations $(\bm{x}_i,y_i)$, $i=1,...,M$. The value $y_i\in\Real$ is regarded
as an \emph{output} or \emph{response} associated with the \emph{input features}
$\bm{x}_i\in\Real^d$ of the system to be learned.} Based on these observations, a
generalized linear prediction model \cite{Duda1} can be written as
\begin{equation}\label{eqn_linear_model_p}
    g(\bm{x}_i,\balpha) = \alpha_0 + \sum^{D-1}_{j=1} p_j(\bm{x}_i) \alpha_j
    = \bm{p}_i^T\balpha, \ \ i=1,...,M,
\end{equation}
where $\bm{p}_i:=[1,p_1(\bm{x}_i),...,p_{D-1}(\bm{x}_i)]^T$ maps the original input
features $\bm{x}_i\in\Real^{d}$ onto the transformed features $\bm{p}_i\in\Real^{D}$ with
{corresponding} weight coefficient vector $\balpha=[\alpha_0,\alpha_1,...,$
$\alpha_{D-1}]^T\in\Real^{D}$. Expression \eqref{eqn_linear_model_p} can be seen as an
expansion in the basis $\{\bm{p}_i\}_{i=1}^M$. Popular choices for such basis are
\emph{Gaussian} \cite{Poggio1}, \emph{Sigmoid} \cite{Bishop100}, \emph{Polynomial}
\cite{Campbell1,Toh39} {or} \emph{Random Projection} \cite{Toh77} functions. For the
given set of $M$ data samples, the {associated} multiple column vectors
$\bm{p}_1,\cdots,\bm{p}_M$ can be stacked as ${\bf P}=[\bm{p}_1,...,\bm{p}_M]^T$ and the
above generalized linear prediction model can be compactly written as
\begin{equation}\label{eqn_linear_model_BigP}
    {\bf g}(\bm{x},\bm{\alpha}) = {\bf P}\bm{\alpha},
\end{equation}
where ${\bf g}=[g(\bm{x}_1,\bm{\alpha}),...,g(\bm{x}_M,\bm{\alpha})]^T$ contains the
corresponding output predictions of the given samples.

Based on the Gaussian random design model, one seeks to reconstruct the unknown weight
coefficients ($\balpha$) from a set of noisy measurements $y_i=g(\bm{x}_i,\balpha)+e_i$,
$i=1,...,M$. Frequently, the random noise $e_i$ is assumed to have independent and
identically distributed random entries with mean 0 and variance $\sigma^2$. From the
perspective of cost minimization, a \emph{Least Squares} (LS) regression can be applied
to minimize the discrepancy between the model output and the target. The LS regression is
based on the magnitudes of $e_i=y_i-g(\bm{\alpha},\bm{x}_i)$ in the \emph{Sum of Squared
Errors} (SSE) sense:
\begin{eqnarray}
    \textrm{SSE}(\bm{\alpha}) &=& {\frac{1}{2}\sum^{M}_{i=1}e_i^2} =
     \frac{1}{2}\|{\bf y}-{\bf P}\bm{\alpha}\|^2_2,\label{eqn_SSE1}
\end{eqnarray}
where ${\bf y}=[y_1,...,y_M]^T$ and $\|\cdot\|_2$ denotes the $\ell^2$-norm of a vector.

In order to deal with possible ill-conditioning when solving the minimum SSE problem, a
popular and yet effective choice is to include {an} $\ell^2$-norm related penalty term
(i.e., $\|\bm{\alpha}\|_2^2:=\sum_{j=0}^{D-1}\alpha_j^2=\bm{\alpha}^T\bm{\alpha}$) while
minimizing the SSE \cite{Duda1,Hastie01}:
\begin{equation}\label{eqn_SSE1_ridge}
    \textrm{SSE}_{ridge}(\bm{\alpha}) =
     \frac{1}{2}\|{\bf y}-{\bf P}\bm{\alpha}\|^2_2
         + \frac{\lambda}{2}\|\bm{\alpha}\|_2^2.
\end{equation}
The closed-form solution of \eqref{eqn_SSE1_ridge} is
\begin{equation}\label{eqn_LSEsoln_ridge}
    \bm{\alpha} = ({\bf P}^T{\bf P}+\lambda{\bf I})^{-1}{\bf P}^T{\bf y},
\end{equation}
where the scalar $\lambda$ is called a regularization factor and ${\bf I}$ is {an}
identity matrix {matching the dimension} of ${\bf P}^T{\bf P}$. In the literature, this
is known as \emph{ridge regression} \cite{Hastie01,Tikhonov1,Hoerl1}.

By generalizing the penalty norm term to arbitrary power $p> 0$, as
\begin{equation}\label{eqn_p_norm_1}
    \|\bm{\alpha}\|_p := \left(\sum^{D-1}_{j=0}|\alpha_j|^p
    \right)^{1/p},
\end{equation}
the problem of minimizing
\begin{equation}\label{eqn_SSE1_bridge}
    \textrm{SSE}_{bridge}(\bm{\alpha}) =
         \sum^{M}_{i=1} \left(y_i - \bm{p}^T_i\bm{\alpha}\right)^2
         + \lambda \|\bm{\alpha}\|_p^p
\end{equation}
is then called a \emph{bridge regression} \cite{Frank1}, where $0< p < 2$ is of
particular interest due to its {compression} capability. When $p=1$, the {\emph{Least
Absolute Shrinkage and Selection Operator} (LASSO, \cite{Tibshirani2,Tibshirani1})}
solves the $\ell^1$-norm penalized regression problem. Several extensions to such {a}
regularization solution search can be found treating the penalty norm term in various
forms (see e.g., \cite{Hastie2,Zou1,Yuan1}). A popular example is the \emph{Elastic-Net}
\cite{Zou1}, which stretches between LASSO and ridge regression, adopting
$\lambda_1\|\bm{\alpha}\|_1 + \lambda_2\|\bm{\alpha}\|^2_2$ as the penalty term in its
naive form.

Apart from the above regularization based methods, an alternative approach to deal with
the singularity of covariance matrix ${\bf P}^T{\bf P}$ (inner product) for
under-determined systems is to express it in its dual form as ${\bf P}{\bf P}^T$ (outer
product) where a constrained optimization problem can be posed \cite{Madych1} to find
$\bm{\alpha}$ which
\begin{eqnarray}
  {\rm minimizes} && \|\bm{\alpha}\|^2_2 \label{eqn_min_norm_2}\\
  {\rm subject\ to\ } && {\bf y} = {\bf P}\bm{\alpha}. \nonumber
\end{eqnarray}
Here, an analytical solution can be obtained based on the non-singularity of ${\bf P}{\bf
P}^T$ in the following form
\begin{equation}\label{eqn_LSEsoln_dual}
    \bm{\alpha} = {\bf P}^T({\bf P}{\bf P}^T)^{-1}{\bf y},
\end{equation}
where the outer product term replaces the inner product term of the regularization
approach. This solution is known as the \emph{minimum-norm} or \emph{least-norm} solution
\cite{Boyd1}, which applies particularly well to an under-determined system when $M<d+1$
or $M<D$. The constrained optimization problem in \eqref{eqn_min_norm_2} can, again, be
generalized to
\begin{eqnarray}
  {\rm minimize} && \|\bm{\alpha}\|^p_p \label{eqn_min_norm_p} \\
  {\rm subject\ to\ } && {\bf y} = {\bf P}\bm{\alpha}. \nonumber
\end{eqnarray}
This leads to solving the corresponding unconstrained minimization problem given by
\begin{equation}\label{eqn_q_norm_Lagrangian}
    \min_{\bm{\alpha}} \|\bm{\alpha}\|^p_{p}
    + \bm{\beta}^T({\bf y}-{\bf P}\bm{\alpha} ),
\end{equation}
where different choices of $p$ result in different parametric stretching.
Here, the elements in vector $\bm{\beta}$ %$\succeq \textbf{0}$
are known as the Lagrange multipliers, where each element corresponds to a given data
sample.

This framework is well adopted in compressed sensing research
\cite{Donoho2,Baraniuk1,Candes1,Ahsen1}, where minimization of the $\ell^0$-norm related
objective is adopted for parameter \emph{subsets selection} \cite{Miller1,Dyer1} and
minimization of the $\ell^1$-norm related objective is adopted for an optimally sparse
solution \cite{Donoho1}. Our work here involves solving an approximated norm metric under
this approach.

\subsection{Motivation and Contributions}

Except for the case when $p=2$, existing solutions to minimize an objective function
containing {an} $\ell^p$-norm term ($\|\bm{\alpha}\|_p$) or its powered form
($\|\bm{\alpha}\|^p_p$) require an iterative search to locate the minimizer. For example,
LASSO minimizes an objective {function} with {an} $\ell^1$-norm penalty {term} which
renders the formulation nonlinear. {This results in} a quadratic programming {problem,}
where no analytical or closed-form expression is known (see \cite{Hastie01}, page 68-69).
The same issue {applies} to Elastic-Net because it is transformed into a LASSO problem
for solution search \cite{Zou1}. For bridge regression \cite{Frank1}, the formulation is
non-convex when $p<1$ and hence major focus was paid to the case when $p\geq 1$ (see
e.g., \cite{FuW1}). Bearing in mind its application to compressed estimation, the range
$0\leq p <2$ is of particular interest. The solution of the bridge regression problem
cannot be expressed in closed from because of the absolute operator in the norm term.
Hence the solutions are obtained using iterative numerical algorithms.

Different from existing approaches in the literature
\cite{Donoho3,Friedman2,CaiTT1,Figueiredo1,Nair1,Pokar1}, we make an attempt to seek a
closed-form solution without the need of any iterative search. Following \cite{Toh88},
where the absolute operator in \eqref{eqn_p_norm_1} was omitted, we propose to
approximate the absolute operator by a smooth function. This gives rise to an
approximated $p$-norm which is differentiable. Optimization of the approximated $p$-norm
subject to data fitting turns out to suggest a possible solution in closed-form. Based on
the relationship between the primal parameter and the dual parameter of a restricted
feasible solution set, an extended solution for the least-squares regression is proposed.

Although the idea of using a differentiable approximation to the $\ell^1$-norm penalty
has been explored earlier in the literature, an iterative search on the nonlinear
formulation remains inevitable. These differentiable approximations include the
\emph{epsL1} function, the \emph{Huber} function and the \emph{log-barrier} function (see
e.g., \cite{LeeSI1,Schmidt1}). The \emph{epsL1} function is defined as
$epsL1(\alpha_j,\epsilon) = \sqrt{\alpha_j^2+\epsilon}$ and the \emph{Huber} function is
defined as $Huber(\alpha_j,\epsilon)=\left\{\begin{array}{cc}
  \alpha_j^2/2\epsilon  & \textrm{if\ }|\alpha_j|<\epsilon \\
  |\alpha_j|-\epsilon/2 & \textrm{otherwise} \\
\end{array} \right.$, with gradient\\
$\nabla_{\alpha_j} Huber(\alpha_j,\epsilon)=\left\{\begin{array}{cc}
  \alpha_j/\epsilon  & \textrm{if\ }|\alpha_j|<\epsilon \\
  \sgn(\alpha_j) & \textrm{otherwise} \\
\end{array} \right.$. The \emph{log-barrier} function is defined as $\mu\log c(\balpha)$ where a popular candidate for
$c(\balpha)$ is $\|\balpha\|^2_2$.

In this work, the \emph{epsL1} function, which has recently been shown to be the most
computationally efficient smooth approximation to the absolute function \cite{Ramirez1},
is adopted to approximate the absolute operator in the $\ell^p$-norm function
\eqref{eqn_p_norm_1} where an analytical solution is attempted. Hence, our contributions
include: (i) a novel closed-form solution for an extended version of the least squares
ridge regression that considers the possibility of stretchy parameters, (ii) an input
transformation to facilitate the computation of power roots, (iii) a variance analysis
regarding the proposed estimator, and (iv) the provision of extensive empirical
{evidence} to validate the feasibility and usefulness of the proposed method. In
particular, our experiments illustrate that the stretching works well for problems with
high-dimensional inputs where matrix ill-conditioning could be an issue in
under-determined systems.

The remainder of the paper is organized as follows. In the next section, an approximation
to the $\ell^p$-norm metric and several related lemmas are introduced. These lemmas are
useful to derive the solution of the optimization problem. The proposed solutions are
next presented within the same section in primal and dual forms. This is followed by
introducing a standardization and transformation process to put the data in perspective.
In Section~\ref{sec_synthetic}, two representative synthetic data sets are evaluated to
illustrate the stretching ability of the proposed method. In Section~\ref{sec_expts}, the
numerical experiments are extended to real-world data sets to observe {the} feasibility
of {the} proposed solution on large dimensional problems. Concluding remarks are given in
Section~\ref{sec_conclusion}.

\section{Proposed Stretchy Regression}

\subsection{$\ell^p$-norm Approximation}

\begin{figure}[hhh]
  \begin{center}
  \epsfxsize=6.8cm
  \epsffile[76   199   546   606]{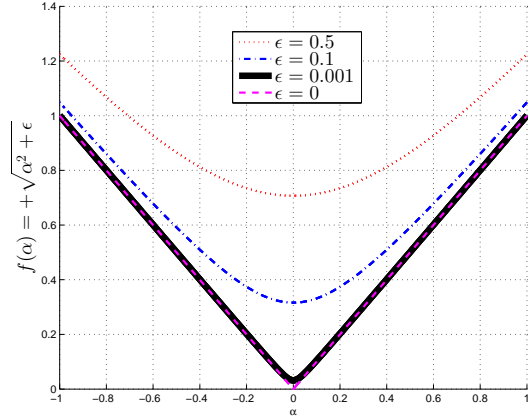}
  \caption{Plot of $f(\alpha)=\sqrt{\alpha^2+\epsilon}$ at several $\epsilon$ values.}
  \label{fig_abs_approx}
  \end{center}
\end{figure}

Consider a positive valued penalty term that is an approximation of the $\ell^p$-norm, in
which the absolute value in \eqref{eqn_p_norm_1} is replaced by a differentiable function
$f$:
\begin{equation}\label{eqn_k_norm}
    \kmeas{\bm{\alpha}} := \left(\sum^{D-1}_{j=0}f(\alpha_j)^{k}
    \right)^{1/k} \negthinspace\negthinspace .
\end{equation}
Here, the power term $k$ replaces $p$ in the $\ell^p$-norm to avoid confusion with the
generalized linear model term $p(\cdot)$ in \eqref{eqn_linear_model_p}. A convenient
choice for such an approximation, which can be efficiently computed, is
$f(\alpha_j)=\sqrt{\alpha_j^2+\epsilon}$, $\epsilon>0$ (see \cite{Ramirez1}) and
Fig.~\ref{fig_abs_approx}). Note that $\lim_{\epsilon\rightarrow 0} f(\cdot) = |\cdot|$
for arbitrary $\epsilon>0$. For finite $\epsilon$ the function $\kmeas{\bm{\alpha}}$ is
not a norm because it does not have the absolute homogeneity property. We shall call
$\kmeas{\cdot}$ \eqref{eqn_k_norm} a \emph{$k$-measure} operator for convenience hereon.

In the following, the raised power form of $k$-measure is shown to be convex when the
approximation function $f(\cdot)$ is convex.\\
\begin{lemma} \label{lemma_convex}
$\kmeask{\balpha}$ is convex on $\balpha$ when $f$ is convex for all $k\geq 1$.
\end{lemma}
Proof: See Appendix A.\\

For our particular case of $f(\alpha_i)=\sqrt{\alpha_i^2+\epsilon}$, it is easy to verify
its convexity because the second derivative of $f(\alpha)$ for each
$\alpha\in\{\alpha_0,\alpha_1,...,\alpha_{D-1}\}$ is
\begin{equation}\label{eqn_2nd_df}
    \frac{d^2f}{d\alpha^2} = \left(\alpha^2+\epsilon \right)^{-\frac{1}{2}}
            - \alpha^2\left(\alpha^2+\epsilon \right)^{-\frac{3}{2}},
\end{equation}
where we see that
\begin{equation}\label{eqn_d2f_ratio}
    \frac{\left(\alpha^2+\epsilon \right)^{-\frac{1}{2}}}{\alpha^2\left(\alpha^2+\epsilon \right)^{-\frac{3}{2}}}
    = \frac{\alpha^2+\epsilon}{\alpha^2} > 1,
\end{equation}
for $\epsilon>0$, which implies convexity of $f$ on $\alpha$ based on $d^2f/d\alpha^2>0$.

\begin{figure}[hhh]
  \begin{center}
  \epsfxsize=9.8cm
  \epsfysize=10.8cm
  \epsffile[58   195   546   602]{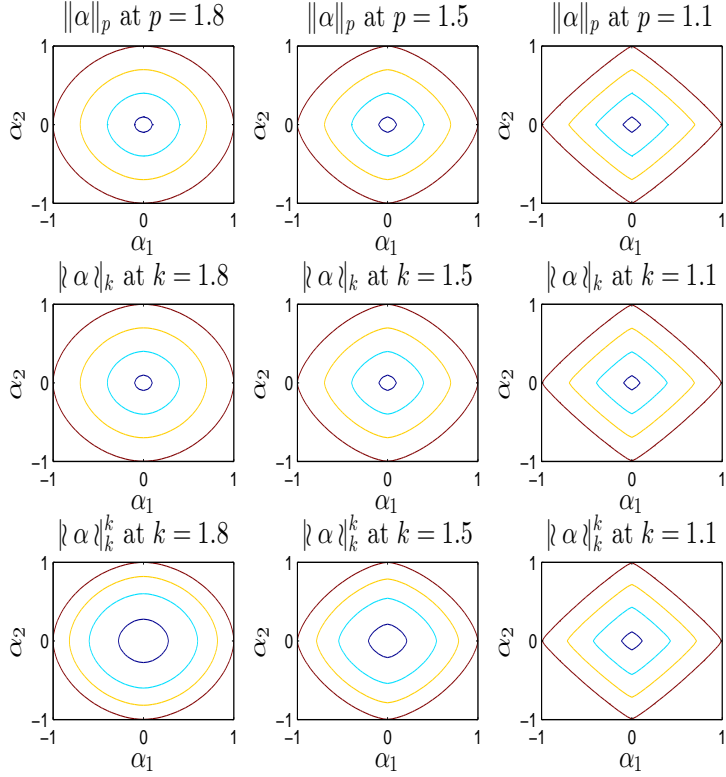}
  \caption{Contour plots at levels [0.1, 0.4, 0.7, 1]. Top panels: $\|\bm{\alpha}\|_p$ for $p\in\{1.8,1.5,1.1\}$.
  Middle Panels: $\kmeas{\bm{\alpha}}$ for $k\in\{1.8,1.5,1.1\}$ at $\epsilon=0.0001$.
  Bottom Panels: $\kmeask{\bm{\alpha}}$ for $k\in\{1.8,1.5,1.1\}$ at $\epsilon=0.0001$.}
  \label{fig_q_norm}
  \end{center}
\end{figure}

Fig.~\ref{fig_q_norm} shows the contours of the $\ell^{p}$-norm metric
\eqref{eqn_p_norm_1} for $1<p<2$ and the corresponding ${k}$-measure ($\kmeas{\balpha}$,
\eqref{eqn_k_norm}) together with its $k$-powered form ($\kmeask{\balpha}$) within the
same interval. From the bottom panels of Fig.~\ref{fig_q_norm}, except for the difference
in curvature, we see that the entire ${k}$-measure and its $k$-powered form approximate
well to the solution $p$-norm for the plotted range of $1<\{p,k\}<2$. Particularly, for
$p,k\rightarrow 1$, we shall show in Lemma~\ref{lemma2} that the two solutions converge.
This suggests vertices of the vector space being feasible solutions for the desired
constrained solution search. Such observation shall be exploited in the following
development for possible sparse solution when $1<k<2$.

\begin{lemma}\label{lemma2}
$\lim_{q\rightarrow 1}\min_{\bm{\alpha}} \kmeasq{\bm{\alpha}}  \ \,{\rm subject\ to\ \,}
{\bf y} = {\bf P}\bm{\alpha}$ is equivalent to $\min_{\bm{\alpha}} \kmeas{\bm{\alpha}}  \
\,{\rm subject\ to\ \,} {\bf y} = {\bf P}\bm{\alpha}$,
$\forall k\in\Real$.\\
\end{lemma}

\begin{proof}
From \eqref{eqn_k_norm}, it is straightforward to check that
$\kmeas{\bm{\alpha}}=\kmeas{\bm{\alpha}}^{q=1} = (\sum^{D-1}_{i=0}f(\alpha_i)^k)^{1/k}$
and $\lim_{q\rightarrow 1}\kmeasq{\bm{\alpha}} = \lim_{q\rightarrow
1}\left(\sum^{D-1}_{i=0}f(\alpha_i)^k
\right)^{q/k}=(\sum^{D-1}_{i=0}f(\alpha_i)^k)^{1/k}=\kmeas{\bm{\alpha}}$. Hence the
proof.
\end{proof}

When $q=k$, we have
\begin{coro}\label{corollary1}
$\lim_{k\rightarrow 1}\min_{\bm{\alpha}} \kmeask{\bm{\alpha}} \ \,{\rm subject\ to\ \,}
{\bf y} = {\bf P}\bm{\alpha}$ is equivalent to $\min_{\bm{\alpha}} \kmea{\bm{\alpha}}_{1}
\ \,{\rm subject\ to\ \,} {\bf
y} = {\bf P}\bm{\alpha}$.\\
\end{coro}

\subsection{Stretchy Regression}

With the above observations, we seek to minimize $\kmeask{\bm{\alpha}}$ subject to ${\bf
y}={\bf P}\bm{\alpha}$ as well as its regularized form by minimizing
$\kmeask{\bm{\alpha}}+\|\bm{e}\|^2_2$ subject to $\bm{e}={\bf y}-{\bf P}\bm{\alpha}$. Our
goal is to have a sparse estimate in the limit of $k\rightarrow 1$. We will need the
following Lemma~\ref{lemma3} for our development of a deterministic solution. Denote the
Hadamard product between vector $\bm{a}\in\Real^d$ and vector $\bm{b}\in\Real^d$ as
$\bm{a}\circ\bm{b}$. By denoting the elementwise operation using $\elementwisen{}$, we
write the elementwise power of matrix/vector as ${\bf A}^{\elementwise{k}}$/${\bf
b}^{\elementwise{k}}$ and the elementwise partial derivative of $\bm{a}$ with respect to
$\bm{b}$ as $\frac{\partial\elementwise{\bm{a}}}{\partial\elementwise{\bm{b}}}$. For
instance, ${\bf A}^{\elementwise{k}}:=[a_{ij}^k]$, for all $i,j$ which index the matrix
elements. Then, consider a scaling vector which factors out ${\bf A}({\bf A}^T{\bf
b})^{\elementwise{k}}$ as $({\bf A}({\bf A}^T)^{\elementwise{k}}{\bf
b}^{\elementwise{k}})\circ\bm{s}$ in the following lemma.\\

\begin{lemma}\label{lemma3}
Consider an $m\times d$ matrix ${\bf A}$ and a $m\times 1$ vector ${\bf b}$. Suppose
$\sum^{d}_{j=1}a_{lj}\sum^{m}_{i=1} a_{ji}^k b_{i}^k\neq 0$ for all $l=1,...,m$, then
there exists a scaling vector $\bm{s}$ such that ${\bf A}({\bf A}^T{\bf
b})^{\elementwise{k}}=({\bf A}({\bf A}^T)^{\elementwise{k}}{\bf
b}^{\elementwise{k}})\circ\bm{s}$.
\end{lemma}
Proof: See Appendix B.\\

We are now ready to derive a deterministic solution for an approximated minimization
problem to \eqref{eqn_q_norm_Lagrangian}.

\begin{thm}\label{thm1}
Consider an under-determined system given by ${\bf y}={\bf P}\bm{\alpha}$ where ${\bf
y}\in\Real^M$, ${\bf P}\in\Real^{M\times D}$ and $\bm{\alpha}\in\Real^D$ with $M<D$.
Suppose ${\bf P}({\bf P}^T)^{\elementwise{\frac{1}{k-1}}}$ is of full rank, then for all
$k>1$, $k\neq\infty$ and under the limiting case of ${\bm{\epsilon}\rightarrow{\bf 0}}$,
the minimizer of
\begin{equation}\label{eqn_q_norm_obj111}
     \min_{\balpha} \kmeask{\balpha}
    \ \,{\rm subject\ to\ \,} {\bf y}={\bf P}\balpha,
\end{equation}
satisfies
\begin{equation}\label{eqn_null_space}
    {\bf P}\left( \balpha - ({\bf P}^T)^{\elementwise{\frac{1}{k-1}}}\left({\bf P}
        ({\bf P}^T)^{\elementwise{\frac{1}{k-1}}}\right)^{-1}{\bf y} \right) = {\bf 0}.
\end{equation}
\end{thm}

\begin{proof}
According to the definition of $k$-measure in \eqref{eqn_k_norm}, let
$\bar{\bm{\alpha}}:=[(\alpha_0^2+\epsilon)^{{k}/4},\cdots,(\alpha_{D-1}^2+\epsilon)^{{k}/4}]^T$
where we can write $\kmeas{\bm{\alpha}}=(\bar{\bm{\alpha}}^T\bar{\bm{\alpha}})^{1/k}$ and
$\kmeask{\bm{\alpha}}=(\bar{\bm{\alpha}}^T\bar{\bm{\alpha}})$. Then, taking the first
derivative of the Lagrange function of \eqref{eqn_q_norm_obj111} and setting it to zero
gives:
\begin{eqnarray}
  \frac{\partial}{\partial\balpha}\left(\bar{\balpha}^T\bar{\balpha}
    + \bm{\beta}^T({\bf y}-{\bf P}\balpha)\right) &=& {\bf 0} \nonumber \\
  \frac{k}{4}\cdot2\balpha\circ\left(\balpha^{\elementwise{2}}+\bm{\epsilon}\right)^{\elementwise{\frac{k}{4}-1}}\circ
    2\bar{\balpha} - {\bf P}^T\bbeta  &=& {\bf 0} \nonumber \\
  {k}\,\balpha\circ\left(\balpha^{\elementwise{2}}+\bm{\epsilon}\right)^{\elementwise{\frac{k}{4}-1}}\circ
    \left(\bm{\alpha}^{\elementwise{2}}+\bm{\epsilon}\right)^{\elementwise{\frac{k}{4}}} - {\bf P}^T\bbeta  &=& {\bf 0} \nonumber \\
  \Rightarrow\hspace{5mm}{k}\,\balpha\circ\left(\balpha^{\elementwise{2}}+\bm{\epsilon}\right)^{\elementwise{\frac{k}{2}-1}} &=& {\bf
  P}^T\bbeta.\hspace{6.8mm}
  \label{eqn_qnorm_alpha110}
\end{eqnarray}
For the limiting case of $\bm{\epsilon}$, we have
\begin{eqnarray}
  \lim_{\bm{\epsilon}\rightarrow{\bf 0}}{k}\,\balpha\circ\left(\balpha^{\elementwise{2}}+\bm{\epsilon}\right)^{\elementwise{\frac{k}{2}-1}}
  &=& {k}\,\balpha\circ\left(\balpha^{\elementwise{2}}\right)^{\elementwise{\frac{k}{2}-1}},
\end{eqnarray}
which implies that for all $k>1$ and $k\neq\infty$,
\begin{eqnarray}
  {k}\,\balpha\circ\left(\balpha^{\elementwise{2}}\right)^{\elementwise{\frac{k-2}{2}}} &=& {\bf P}^T\bbeta \nonumber \\
  {k}\,\sgn(\balpha)\circ(\balpha^{\elementwise{2}})^{\frac{1}{2}}\circ\left(\balpha^{\elementwise{2}}\right)^{\elementwise{\frac{k-2}{2}}} &=& {\bf P}^T\bm{\beta} \nonumber \\
  {k}\,\sgn(\balpha)\circ\left(\balpha^{\elementwise{2}}\right)^{\elementwise{\frac{k-1}{2}}} &=& {\bf P}^T\bm{\beta} \nonumber \\
  && \hspace{-4.3cm}\Rightarrow\hspace{5mm}\left(\balpha^{\elementwise{2}}\right)^{\elementwise{\frac{k-1}{2}}}\ \ =\ \ \sgn(\balpha)\circ\left(\frac{1}{k}{\bf P}^T\bm{\beta}\right)
  .\label{eqn_alpha_square}
\end{eqnarray}
Taking square elementwise for both sides of \eqref{eqn_alpha_square}, we have
\begin{eqnarray}
  \left(\bm{\alpha}^{\elementwise{2}}\right)^{\elementwise{k-1}} &=& \left(\frac{1}{k}{\bf P}^T\bm{\beta}\right)^{\elementwise{2}}
  .
\end{eqnarray}
From \eqref{eqn_qnorm_alpha110}, we know that the vector
$\lim_{\bm{\epsilon}\rightarrow{\bf 0}}(\balpha^{\elementwise{2}}+\bm{\epsilon})$ has
positive elements and thus $\lim_{\bm{\epsilon}\rightarrow{\bf
0}}\left(\balpha^{\elementwise{2}}+\bm{\epsilon}\right)^{\elementwise{\frac{k}{2}-1}}$
has positive elements regardless of $k$. Hence, we deduce that $\sgn(\balpha)=\sgn({\bf
P}^T\bm{\beta})$, and
\begin{eqnarray}
  \bm{\alpha} &=&
    \left({\bf P}^T\left\{\frac{1}{k}\bm{\beta}\right\}\right)^{\elementwise{\frac{1}{k-1}}}
   .
\label{eqn_qnorm_alpha111}
\end{eqnarray}

For the matrix $({\bf P}^T)^{\elementwise{\frac{1}{k-1}}}$ with elements
$p_{ji}^{\frac{1}{k-1}}$, $j=0,...,{D-1}$, $i=1,...,M$ and the vector
$\left\{\frac{1}{k}\bm{\beta}\right\}^{\elementwise{\frac{1}{k-1}}}$ with elements
$\beta_i/k$, $i=1,...,M$, supposing $\sum^{D-1}_{i=0} p_{lj}\sum^{M}_{i=1}
p_{ji}^{\frac{1}{k-1}} (\beta_{i}/k)^{\frac{1}{k-1}}\neq 0$ for all $l=1,...,M$, we can
factor out \eqref{eqn_qnorm_alpha111} using Lemma~\ref{lemma3}, giving
\begin{eqnarray}
    {\bf P}\bm{\alpha}
    &=& {\bf P}\left({\bf P}^T\left\{\frac{1}{k}\bm{\beta}\right\}\right)^{\elementwise{\frac{1}{k-1}}}
    \nonumber\\
    &=& \left( {\bf P}
        ({\bf P}^T)^{\elementwise{\frac{1}{k-1}}}\left\{\frac{1}{k}\bm{\beta}\right\}^{\elementwise{\frac{1}{k-1}}}
        \right) \circ\bm{s}.
\label{eqn_alpha_knorm}
\end{eqnarray}

Next, replacing ${\bf P}\bm{\alpha}$ by ${\bf y}$ gives
\begin{eqnarray}
  {\bf y} &=& \left(
    {\bf P} ({\bf P}^T)^{\elementwise{\frac{1}{k-1}}}\left\{\frac{1}{k}\bm{\beta}\right\}^{\elementwise{\frac{1}{k-1}}} \right) \circ\bm{s} \nonumber \\
  \Rightarrow\hspace{3mm}\frac{1}{k}\bm{\beta} &=& \left[\left({\bf P}({\bf P}^T)^{\elementwise{\frac{1}{k-1}}}\right)^{-1}{\bf y}\circ\bm{s}^{\elementwise{-1}}
    \right]^{\elementwise{k-1}} .
\end{eqnarray}
Substituting $\frac{1}{k}\bm{\beta}$ into \eqref{eqn_alpha_knorm} and simplifying, we
have for all $k>1$, $k\neq\infty$ and the limiting case of $\bm{\epsilon}\rightarrow{\bf
0}$,
\begin{eqnarray}
  {\bf P}\bm{\alpha}
   &=&  \left( {\bf P}
        ({\bf P}^T)^{\elementwise{\frac{1}{k-1}}}\left\{\frac{1}{k}\bm{\beta}\right\}^{\elementwise{\frac{1}{k-1}}}
        \right) \circ\bm{s}  \nonumber \\
   &=&  \left( {\bf P}
   ({\bf P}^T)^{\elementwise{\frac{1}{k-1}}}\left({\bf P}({\bf P}^T)^{\elementwise{\frac{1}{k-1}}}\right)^{-1}{\bf y}\circ\bm{s}^{\elementwise{-1}}
        \right) \circ\bm{s}  \nonumber \\
   &=&  {\bf P}
        ({\bf P}^T)^{\elementwise{\frac{1}{k-1}}}\left({\bf P}
        ({\bf P}^T)^{\elementwise{\frac{1}{k-1}}}\right)^{-1}{\bf y} .
        \label{eqn_qnorm_dual_soln}
\end{eqnarray}
As $\bm{s}$ vanishes in the above derivation, its possible singularity (i.e., even when
$\sum^{D-1}_{j=0} p_{lj}$ $\sum^{M}_{i=1} p_{ji}^{\frac{1}{k-1}}
(\beta_{i}/k)^{\frac{1}{k-1}}= 0$, $l=1,...,M$) takes no effect in the resultant
solution.

The above cannot be simplified further since ${\bf P}$ cannot be removed from the left in
(24) due to its rank deficiency for $M<D$. Nevertheless, \eqref{eqn_qnorm_dual_soln}
implies \eqref{eqn_null_space} and gives the solution of $\balpha$ implicitly, that is $(
\balpha - ({\bf P}^T)^{\elementwise{\frac{1}{k-1}}}({\bf P}
        ({\bf P}^T)^{\elementwise{\frac{1}{k-1}}})^{-1}{\bf y} )$
lies in the null space of ${\bf P}$. The optimality of the minimizer comes from the
summation of the convex $\kmeask{\cdot}$ (Lemma~\ref{lemma_convex}) and linear
$\bm{\beta}^T({\bf y}-{\bf P}\bm{\alpha})$ functions. Hence the result.
\end{proof}

A direct consequence of the above result is\\
\begin{coro}\label{coro-two}
Suppose ${\bf P}({\bf P}^T)^{\elementwise{\frac{1}{k-1}}}$ is of full rank. Then, by
constraining the feasible solution space to $\bm{\alpha} = ({\bf
P}^T)^{\elementwise{\frac{1}{k-1}}}\bbeta$, $\forall\bbeta\in\Real^M$ for all $k>1$ and
$k\neq\infty$, problem \eqref{eqn_q_norm_obj111} admits the particular solution
\begin{equation}\label{eqn_k_norm_solution}
    \balpha = \left({\bf P}^T\right)^{\elementwise{\frac{1}{k-1}}}\left[{\bf P}
        ({\bf P}^T)^{\elementwise{\frac{1}{k-1}}}\right]^{-1}{\bf y}.
\end{equation}
\end{coro}
\begin{proof}
The result is obtained by substituting $\bm{\alpha} = ({\bf
P}^T)^{\elementwise{\frac{1}{k-1}}}\bbeta$ into \eqref{eqn_qnorm_dual_soln}.
\end{proof}

Under practical considerations, an explicit suppression of the estimation noise along
with the $k$-measure regularization is useful:
\begin{thm}\label{thm3}
Consider an under-determined system given by ${\bf y}={\bf P}\bm{\alpha}+\bm{e}$ where
${\bf y},\bm{e}\in\Real^M$, ${\bf P}\in\Real^{M\times D}$ and $\bm{\alpha}\in\Real^D$
with $M<D$. Suppose ${\bf P}({\bf P}^T)^{\elementwise{\frac{1}{k-1}}}$ is of full rank.
Then, by restricting the feasible solution space to $\bm{\alpha} = ({\bf
P}^T)^{\elementwise{\frac{1}{k-1}}} \bbeta$ for all $k>1$ and $k\neq\infty$ on the
Lagrange multipliers $\bbeta\in\Real^M$, minimization of
\begin{equation}\label{eqn_LSE_obj555}
     \kmeask{\balpha} + {c}\|\bm{e}\|^2_2
\end{equation}
subject to the constraint
\begin{equation}\label{eqn_constrain_with_noise}
    \bm{e} = {\bf y} - {\bf P}\balpha
\end{equation}
under the limiting case of ${\bm{\epsilon}\rightarrow{\bf 0}}$ admits the solution
\begin{equation}\label{eqn_k_norm_solution_reg}
    \balpha = \left({\bf P}^T\right)^{\elementwise{\frac{1}{k-1}}}
    \left[{\bf P}({\bf P}^T)^{\elementwise{\frac{1}{k-1}}}
     + \frac{1}{{c}k}{\bf I} \right]^{-1}  {\bf y} ,
\end{equation}
where $c>0$ is the regularization parameter.
\end{thm}

\begin{proof}
Taking the first derivative of the Lagrange function and considering the independency of
the random variable $\bm{e}$ under this dual space construction \cite{Saunders1}, we have
\begin{eqnarray}
  \frac{\partial}{\partial\balpha}\left( \kmeask{\balpha}
    + {c}\bm{e}^T\bm{e}
    + \bm{\beta}^T({\bf y}-{\bf P}\balpha-\bm{e})\right) &=& {\bf 0}
    \nonumber \\
  \Rightarrow\hspace{5mm}\frac{\partial}{\partial\balpha}\left( \kmeask{\balpha}
    + \bm{\beta}^T({\bf y}-{\bf P}\balpha)\right) &=& {\bf 0}
    .\label{eqn_qnorm_alpha555}
\end{eqnarray}
Similar to \eqref{eqn_qnorm_alpha110}, solving \eqref{eqn_qnorm_alpha555} results in
\eqref{eqn_alpha_square}.

Substituting the error constraint \eqref{eqn_constrain_with_noise} and the solution
\eqref{eqn_alpha_square} into the cost function \eqref{eqn_LSE_obj555} under the limiting
case of $\bepsilon\rightarrow{\bf 0}$ we have
\begin{eqnarray}
  \lim_{\bm{\epsilon}\rightarrow 0}\kmeask{\balpha} + {c}\bm{e}^T\bm{e}
  &=& ((\balpha^T)^{\elementwise{2}})^{\elementwise{\frac{k-1}{2}}}(\balpha^{\elementwise{2}})^{\elementwise{\frac{1}{2}}}
        + {c}({\bf y}-{\bf P}\balpha)^T({\bf y}-{\bf P}\balpha)
        \nonumber\\
  &=&
  \left(\sgn(\balpha^T)\circ\frac{1}{k}\bbeta^T{\bf P}\right)\left(\balpha\circ\sgn(\balpha)\right)
        + {c}{\bf y}^T{\bf y} - 2{c}{\bf y}^T{\bf P}\balpha
        \nonumber + {c}\balpha^T{\bf P}^T{\bf P}\balpha
        \nonumber\\
  &=&
  \frac{1}{k}\bbeta^T{\bf P}\balpha
        + {c}{\bf y}^T{\bf y} - 2{c}{\bf y}^T{\bf P}\balpha
        + {c}\balpha^T{\bf P}^T{\bf P}\balpha
  .
  \label{eqn_reg_obj_0}
\end{eqnarray}
Since the solution is restricted to the subspace $\balpha = ({\bf
P}^T)^{\elementwise{\frac{1}{k-1}}} \bbeta$, equation~\eqref{eqn_reg_obj_0} can be
written as
\begin{eqnarray}
  \lim_{\bm{\epsilon}\rightarrow 0}\kmeask{\balpha} + c\bm{e}^T\bm{e}
  &=& \frac{1}{k}\bbeta^T{\bf P} ({\bf P}^T)^{\elementwise{\frac{1}{k-1}}}\bbeta
        + {c}{\bf y}^T{\bf y} - 2{c}{\bf y}^T{\bf P}({\bf P}^T)^{\elementwise{\frac{1}{k-1}}}\bbeta
        \nonumber\\ &&
        \hspace{3mm}
        + {c}\bbeta^T{\bf P}^{\elementwise{\frac{1}{k-1}}}{\bf P}^T{\bf P}
        ({\bf P}^T)^{\elementwise{\frac{1}{k-1}}}\bbeta
        \nonumber\\
  &=& \frac{1}{k}\bbeta^T {\bf K} \bbeta
        + {c}{\bf y}^T{\bf y} - 2{c}{\bf y}^T{\bf K} \bbeta
        + {c}\bbeta^T{\bf K}^T{\bf K} \bbeta ,
        \label{eqn_reg_obj_1}
\end{eqnarray}
where ${\bf K}={\bf P}({\bf P}^T)^{\elementwise{\frac{1}{k-1}}}$.

Differentiating \eqref{eqn_reg_obj_1} with respect to $\bbeta$ and put to zero gives
\begin{eqnarray}
    2\frac{1}{k}{\bf K} \bbeta - 2{c}{\bf K}{\bf y} + 2{c}{\bf K}^T{\bf K} \bbeta &=& {\bf 0} .
\label{eqn_qnorm_alpha_derivation4}
\end{eqnarray}
Since ${\bf K}$ is nonsingular, we have
\begin{eqnarray}
    (\frac{1}{k}{\bf K} + {c}{\bf K}^T{\bf K}) \bbeta  &=& c{\bf K}{\bf y}
    \nonumber\\
    \Rightarrow\hspace{1cm}\bbeta &=& {\bf K}^{-1}(\frac{1}{{c}k}{\bf I} + {\bf K})^{-1}{\bf K}{\bf y}.
\label{eqn_qnorm_alpha_derivation5}
\end{eqnarray}
Using the matrix identity $({\bf I} + {\bf A}{\bf B})^{-1}{\bf A} = {\bf A}({\bf I} +
{\bf B}{\bf A})^{-1}$ \cite{Petersen1}, the above can be simplified as
\begin{eqnarray}
    \bbeta &=& {ck}{\bf K}^{-1}({\bf I} + {\bf K}{c}k)^{-1}{\bf K}{\bf y}
            \nonumber \\
    &=& {ck}{\bf K}^{-1}{\bf K}({\bf I} + {c}k{\bf K})^{-1}{\bf y}
            \nonumber \\
    &=& (\frac{1}{{c}k}{\bf I} + {\bf K})^{-1}{\bf y} .
\label{eqn_qnorm_alpha_derivation6}
\end{eqnarray}
The result \eqref{eqn_k_norm_solution_reg} arises from substituting
\eqref{eqn_qnorm_alpha_derivation6} into the restricted solution space $\bm{\alpha} =
({\bf P}^T)^{\elementwise{\frac{1}{k-1}}} \bbeta$. Hence the proof.
\end{proof}

From minimizing the SSE perspective, the following result is a direct consequence.\\
\begin{coro}\label{coro-three}
Suppose ${\bf P}({\bf P}^T)^{\elementwise{\frac{1}{k-1}}}$ is of full rank. Then, by
restricting the feasible solution space to $\bm{\alpha} = ({\bf
P}^T)^{\elementwise{\frac{1}{k-1}}} \bbeta$, $\forall\bbeta\in\Real^M$ and under the
limiting case of $c\rightarrow \infty$, $\bm{\epsilon}\rightarrow {\bf 0}$, the problem
\eqref{eqn_LSE_obj555}-\eqref{eqn_constrain_with_noise} leads to SSE \eqref{eqn_SSE1}
which admits an extended least-squares solution in the form of
\eqref{eqn_k_norm_solution} for all $k>1$ and $k\neq\infty$.
\end{coro}

\begin{proof}  Substituting $\bm{\alpha} =
({\bf P}^T)^{\elementwise{\frac{1}{k-1}}} \bbeta$ into the least-squares error objective
function \eqref{eqn_SSE1} and minimizing it with respect to $\bbeta$, we have
\begin{eqnarray}
    \frac{\partial}{\partial\bbeta}\left({\bf y}-{\bf P}({\bf P}^T)^{\elementwise{\frac{1}{k-1}}} \bbeta\right)^T
        \left({\bf y}-{\bf P}({\bf P}^T)^{\elementwise{\frac{1}{k-1}}} \bbeta\right)
       = {\bf 0} && \nonumber \\ \Rightarrow\hspace{1cm}
    {\bf P}({\bf P}^T)^{\elementwise{\frac{1}{k-1}}} \left({\bf y}-
        {\bf P}({\bf P}^T)^{\elementwise{\frac{1}{k-1}}} \bbeta\right)
       = {\bf 0} && \nonumber \\ \Rightarrow\hspace{2.3cm}
    \bbeta = \left[{\bf P}({\bf P}^T)^{\elementwise{\frac{1}{k-1}}}\right]^{-1}{\bf  y}, &&
    \label{eqn_min_theta}
\end{eqnarray}
because ${\bf P}({\bf P}^T)^{\elementwise{\frac{1}{k-1}}}$ is nonsingular. The solution
\eqref{eqn_k_norm_solution} follows from substitution of \eqref{eqn_min_theta} into
$\bm{\alpha} = ({\bf P}^T)^{\elementwise{\frac{1}{k-1}}} \bbeta$. Hence the result.
\end{proof}

{\bf Remark 1: } When $c=0$, problem
\eqref{eqn_LSE_obj555}-\eqref{eqn_constrain_with_noise} becomes degenerate. This is
because minimizing $\kmeask{\balpha}$ alone subject to the constraint $\bm{e}={\bf y} -
{\bf P}\balpha$ without specifying a bound is not meaningful.

When $c\rightarrow \infty$, problem
\eqref{eqn_LSE_obj555}-\eqref{eqn_constrain_with_noise} approaches problem
\eqref{eqn_SSE1}, and the solution \eqref{eqn_k_norm_solution_reg} approaches
\eqref{eqn_k_norm_solution}. Here, although the SSE objective function \eqref{eqn_SSE1}
does not involve any $k$ term, the particular solution \eqref{eqn_k_norm_solution}, which
contains the power term $1/(k-1)$, is one that arises from the restricted solution search
space.  In other words, given $M<D$, the restricted feasible space mapping $\bm{\alpha} =
({\bf P}^T)^{\elementwise{\frac{1}{k-1}}} \bbeta$ projects the parameter vector
$\balpha\in\Real^D$ onto a lower dimensional vector $\bbeta\in\Real^M$ for deterministic
estimation of the under-determined system. When $k=2$, the solution
\eqref{eqn_k_norm_solution} leads to \eqref{eqn_LSEsoln_dual}. For $k<2$, we shall
observe whether such stretchy estimation can be realized without impairing
the estimation accuracy in the experiments. \hspace*{\fill}$\square$ \\

Apart from the elementwise powered matrix, we further note that the solution forms
\eqref{eqn_k_norm_solution} and \eqref{eqn_k_norm_solution_reg} are analogous to that of
the least-norm solution in \eqref{eqn_LSEsoln_dual} and its regularized form. Besides the
parameter vector of interest ($\balpha$), derivation of the above solution involves two
other parameters $k$ and $c$. The parameter $k$ can be considered as a hyper-parameter to
be pre-fixed for any desired compression level. This adds a degree of freedom for
compressing the parameter vector of interest. The other hyper-parameter $c$ controls the
amount of regularization. The impact of the values of $k$ and $c$ shall be studied in the
analysis of
variance and numerical experiments using both synthetic and real-world data. \\

\subsection{Analysis of Variance}

To analyze the performance of the estimator $\hat{\bm{ \alpha}}$, we assume that ${\bf y}
= {\bf P}\bm{\alpha} + \bm{e}$ where $\bm{e}$ is an independent and identically
distributed noise of zero mean with covariance matrix ${\bf C}$.

Let the size of $\bm{\alpha}$ be $D\times 1$ and that of ${\bf P}$ be $M \times D$.
For under-determined systems \eqref{eqn_k_norm_solution_reg}, %\eqref{eqn_k_norm_solution},
we have $M < D$. In such a case, the expectation of $\hat{\bm{ \alpha}}$ is given by (see
Appendix \ref{app_AoV}):
 \begin{eqnarray}
   E [\hat{\bm{ \alpha}}]
    &=&
         \left({\bf P}^T\right)^{\elementwise{\frac{1}{k-1}}}\left[{\bf P}
        ({\bf P}^T)^{\elementwise{\frac{1}{k-1}}} + \frac{1}{{c}k}{\bf I}\right]^{-1}
    {\bf P}\bm{\alpha} .
    \label{eqn_expect1}
\end{eqnarray}
In order for $ E [\hat{\bm{ \alpha}}] = \bm{\alpha}$, it is required that the matrix
$\left({\bf P}^T\right)^{\elementwise{\frac{1}{k-1}}}\left[{\bf P}
        ({\bf P}^T)^{\elementwise{\frac{1}{k-1}}} + \frac{1}{{c}k}{\bf I}\right]^{-1}
    {\bf P}$ must be an $D \times D$ identity matrix, which is not possible
    since for $M < D$,\\
    $\left({\bf P}^T\right)^{\elementwise{\frac{1}{k-1}}}\left[{\bf P}
        ({\bf P}^T)^{\elementwise{\frac{1}{k-1}}} + \frac{1}{{c}k}{\bf I}\right]^{-1}
    {\bf P}$ is at most of rank $M < D$. Hence,
$\hat{\bm{ \alpha}}$ is a biased estimator. This is similar to the LS estimator (with
$k=2$). This result is congruence to the proposed compressed estimate where some
parameters can vanish, causing $E [\hat{\bm{ \alpha}}] \neq \bm{\alpha}$.

The corresponding covariance matrix of $\hat{\bm{ \alpha}}$ is
\begin{eqnarray}
E [ (\hat{\bm{ \alpha}}-E [\hat{\bm{ \alpha}}]) (\hat{\bm{ \alpha}}-E [\hat{\bm{
\alpha}}])^T ]
   &=&
    \left \{
    \left({\bf P}^T\right)^{\elementwise{\frac{1}{k-1}}}\left[{\bf P}
        ({\bf P}^T)^{\elementwise{\frac{1}{k-1}}} + \frac{1}{{c}k}{\bf I}\right]^{-1}%\right \}
    \right \} {\bf C}  \nonumber\\
    && \qquad \times
      \left \{
    \left({\bf P}^T\right)^{\elementwise{\frac{1}{k-1}}}\left[{\bf P}
        ({\bf P}^T)^{\elementwise{\frac{1}{k-1}}} + \frac{1}{{c}k}{\bf I}\right]^{-1}
   \right \}^T  .
   \label{eqn_covar1}
\end{eqnarray}
When $k=2$, it becomes the ordinary LS estimator and standard LS result can be applied to
simplify the above expression.

Next, we proceed to convert the above solution in the dual space form (outer product
form) to the primal form (inner product form). Based on the matrix identity ${\bf A}({\bf
I}+{\bf B}{\bf A})^{-1}=({\bf I}+{\bf A}{\bf B})^{-1}{\bf A}$ \cite{Petersen1}, the
solution \eqref{eqn_k_norm_solution_reg} under the \emph{dual} space can be re-written in
the \emph{primal} space after simple manipulations as
\begin{equation}\label{eqn_qnorm_primal_soln}
  \bm{\alpha} = \left[({\bf P}^T)^{\elementwise{\frac{1}{k-1}}}{\bf P}
        + \frac{1}{ck}{\bf I} \right]^{-1}
        \left({\bf P}^T\right)^{\elementwise{\frac{1}{k-1}}}{\bf y}.
\end{equation}

 For variance analysis, consider \eqref{eqn_qnorm_primal_soln} with over-determined systems where
 $({\bf P}^T)^{\elementwise{\frac{1}{k-1}}}{\bf P}$ is nonsingular. Let the size of
 $\bm{\alpha}$ be $D\times 1$ and that of ${\bf P}$ be $M \times D$. For even-determined and over-determined systems,
 we have $M \geq D$. In such a case, the expectation of $\hat{\bm{ \alpha}}$ is given
by
 \begin{eqnarray}
   E [\hat{\bm{ \alpha}}]
   =
     E \left [
    \left[({\bf P}^T)^{\elementwise{\frac{1}{k-1}}}{\bf P}
         + \frac{1}{{c}k}{\bf I}\right]^{-1}
        \left({\bf P}^T\right)^{\elementwise{\frac{1}{k-1}}}
    ({\bf P}\bm{\alpha}) \right ], &&
    \label{eqn_expect2}
\end{eqnarray}
where $
  \lim_{c\rightarrow\infty}  E [\hat{\bm{ \alpha}}] = \bm{\alpha} .
$ Hence, $\hat{\bm{ \alpha}}$ is an un-biased estimator regardless of the value of $k$
and for a general ${\bf C}$ under the least-squares minimization case in \eqref{eqn_SSE1}
when $c\rightarrow\infty$. This may be considered as a generalization of the LS estimator
(with $k=2$). Here, $\lim_{c\rightarrow\infty} E [\hat{\bm{ \alpha}}] = \bm{\alpha}$
indicates that no parameter vanishes in the expectation. In other words, the analysis
shows that estimation under the over-determined setting cannot be sparse for the proposed
solution.

For this case with over-determined systems, the covariance matrix of $\hat{\bm{ \alpha}}$
is
\begin{eqnarray}
E [ (\hat{\bm{ \alpha}}-E [\hat{\bm{ \alpha}}]) (\hat{\bm{ \alpha}}-E [\hat{\bm{
\alpha}}])^T ]
   &=&
  \left \{
  \left[({\bf P}^T)^{\elementwise{\frac{1}{k-1}}}{\bf P}
         + \frac{1}{{c}k}{\bf I}\right]^{-1}
         \left({\bf P}^T\right)^{\elementwise{\frac{1}{k-1}}}
         \right \} {\bf C} \nonumber \\
    &&   \qquad \times
     \left \{
     \left[({\bf P}^T)^{\elementwise{\frac{1}{k-1}}}{\bf P}
         + \frac{1}{{c}k}{\bf I}\right]^{-1}
         \left({\bf P}^T\right)^{\elementwise{\frac{1}{k-1}}}
       \right \}^T .
       \label{eqn_covar2}
\end{eqnarray}

Consider now some special cases. When $k=2$, it becomes the ordinary LS estimator and
standard LS result can be applied to simplify the above (even for the general ${\bf C}$
matrix).

When $k \neq 2$, the situation becomes more complicated. Assume first that ${\bf C} =
\sigma^2 {\bf I}_D$. We have
\begin{eqnarray}
E [ (\hat{\bm{ \alpha}}-E [\hat{\bm{ \alpha}}]) (\hat{\bm{ \alpha}}-E [\hat{\bm{
\alpha}}])^T ]
  &=&
  \sigma^2 \left \{
  \left[({\bf P}^T)^{\elementwise{\frac{1}{k-1}}}{\bf P}
         + \frac{1}{{c}k}{\bf I}\right]^{-1}
         \left({\bf P}^T\right)^{\elementwise{\frac{1}{k-1}}}
         \left({\bf P}\right)^{\elementwise{\frac{1}{k-1}}}
         \right \} \nonumber\\
    && \qquad \times
     \left \{
     \left[({\bf P}^T)^{\elementwise{\frac{1}{k-1}}}{\bf P}
         + \frac{1}{{c}k}{\bf I}\right]^{-1}
         \right \}^T .
\end{eqnarray}
Since $k \neq 2$, $({\bf P}^T)^{\elementwise{\frac{1}{k-1}}}{\bf P} \neq \left({\bf
P}^T\right)^{\elementwise{\frac{1}{k-1}}}
        \left({\bf P}\right)^{\elementwise{\frac{1}{k-1}}}$. Hence,\\
        $\left[({\bf P}^T)^{\elementwise{\frac{1}{k-1}}}{\bf P}
         + \frac{1}{ck}{\bf I}\right]^{-1}
        \left({\bf P}^T\right)^{\elementwise{\frac{1}{k-1}}}
        \left({\bf P}\right)^{\elementwise{\frac{1}{k-1}}} \neq {\bf I}_D$. Therefore, unlike the case with
        $k=2$, the above expression cannot be simplified further in general.

Another issue is the singularity of the matrix $({\bf
P}^T)^{\elementwise{\frac{1}{k-1}}}{\bf P}$ or ${\bf P}({\bf
P}^T)^{\elementwise{\frac{1}{k-1}}}$. So far, we assume that it is invertible. For the
case of $k=2$, the non-singularity of $({\bf P}^T)^{\elementwise{\frac{1}{k-1}}}{\bf P}$
or ${\bf P}({\bf P}^T)^{\elementwise{\frac{1}{k-1}}}$ is guaranteed by making ${\bf P}$
non-singular. However, this may no longer be true for $k \neq 2$. We limit our discussion
to $1<k<\infty$ where $k$ can be a non-integer.

It is easy to show that when $k\rightarrow 1$, $\frac{1}{k-1} \rightarrow \infty$, and
consequently,
 $({\bf P}^T)^{\elementwise{\frac{1}{k-1}}}$ is approaching a rank 1 matrix even if ${\bf P}$ is non-singular. Similarly,
 it can be shown that when $k\rightarrow \infty$, $\frac{1}{k-1} \rightarrow 0$, and consequently,
 $({\bf P}^T)^{\elementwise{\frac{1}{k-1}}}$ is approaching a rank 1 matrix even if ${\bf P}$ is non-singular. In both cases,
$({\bf P}^T)^{\elementwise{\frac{1}{k-1}}}{\bf P}$ and ${\bf P}({\bf
P}^T)^{\elementwise{\frac{1}{k-1}}}$ become rank 1 matrices also and it is hence
non-invertible. We conjecture that $k=2$ is the most stable case (in terms of matrix
regularity) and it tends to singular when $k$ is moving away from 2, in both directions.

In the experiments, we shall adopt \eqref{eqn_k_norm_solution_reg} when dealing with
under-determined systems and adopt \eqref{eqn_qnorm_primal_soln} when dealing with
over-determined systems.

\subsection{First Quadrant Transformation} \label{sec_firstQ}

Consider a stacked set of raw training input data given by
\begin{equation}\label{eqn_X}
    {\bf X}_{raw} = \stackrel{
\left[
\begin{array}{cccc}
  {\bf x}_1 & {\bf x}_2 & \cdots & {\bf x}_d \\
\end{array}
\right]
    }{\overbrace{\left[
\begin{array}{cccc}
  x_{11} & x_{12} & \cdots & x_{1d} \\
  x_{21} & x_{22} & \cdots & x_{2d} \\
  \vdots & \vdots & \ddots & \vdots \\
  x_{M1} & x_{M2} & \cdots & x_{Md} \\
\end{array}
\right] }} = \left[
\begin{array}{c}
  \bm{x}_1 \\
  \bm{x}_2 \\
  \vdots \\
  \bm{x}_M \\
\end{array}
\right].
\end{equation}
A standardization is first performed for each data column by a $z$-score normalization
based on the statistics of training set ($\mu_{{\bf x}_j}$ and $\sigma_{{\bf x}_j}$):
\begin{equation}\label{eqn_zscore}
    \underline{{\bf x}}_j=({\bf x}_j - \mu_{{\bf x}_j})/\sigma_{{\bf x}_j},
    \ j=1,...,d.
\end{equation}
Then, an exponential function is adopted to map the standardized data into the first
quadrant:
\begin{equation}\label{eqn_exponential_transform}
    \breve{{\bf x}}=\exp(a\underline{{\bf x}}_j + \bm{b}_j).
\end{equation}

\begin{figure}[hhhh]
  \begin{center}
  \epsfxsize=10cm
  \epsffile[0    17   580   489]{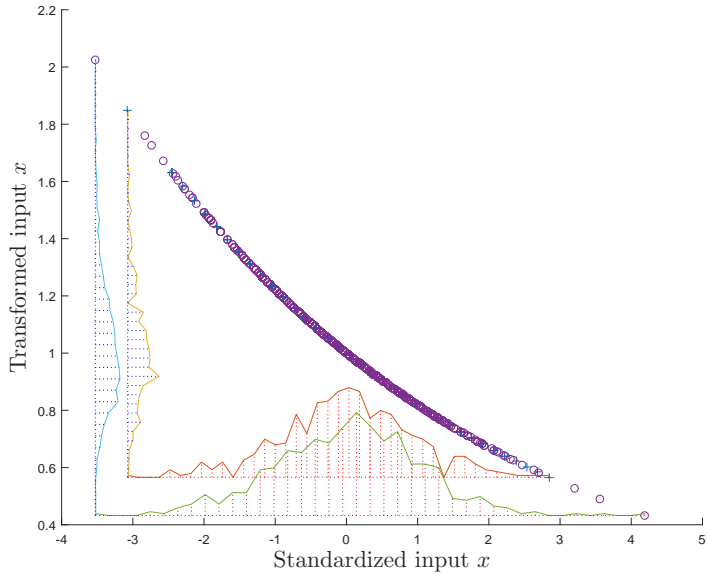}
  \caption{Transformation of the first two dimensions of the Madelon data. The shaded
  regions are the corresponding distributions of the input features.}
  \label{fig_transformation_study}
  \end{center}
\end{figure}

Fig.~\ref{fig_transformation_study} shows the plots of transforming the first two
dimensions of the Madelon data by using $a=-0.2$ and $\bm{b}=\bm{0}$ of
\eqref{eqn_exponential_transform}. Since the test data is assumed to be unseen, the
$z$-score standardization for the test set has been performed based on the parameters
(mean and variance) obtained from the training set. This results in a slight variation of
data spread for each dimension. The proposed transformation mapping is seen to be nearly
linear for the observed data range. The merit of such mapping is that it can be easily
integrated as part of the feature transformation matrix ${\bf P}$ in
\eqref{eqn_linear_model_BigP}.

Here, we note that the exponential transformation serves two purposes: first quadrant
transformation and data warping. The main reason for the first quadrant transformation is
to ensure that all input data contains only positive real values. This is to avoid
occurrence of complex numbers when taking the fractional power of $1/(k-1)$ on negative
elements of ${\bf P}$. The data warping mechanism \emph{twists} the original data such
that large values are far more distinguishable than those small values. This twisting
further stretches (or compresses) the relative difference among the input variables on
top of the stretchy regression.

\section{Synthetic Data} \label{sec_synthetic}

In order to understand the stretching capability as well as the underlying decision
boundary of the proposed method, two synthetic data examples are studied in this section.
The first example demonstrates the scenario of an under-determined system for regression
application. The second example represents an over-determined system for binary
classification applications. We believe these two examples demonstrate several essential
components of physical data distribution to help our study. The components of interest
include nonlinearity, overlapping category distributions, under- and over-determined
system scenarios.

\begin{figure}[hhhh]
  \begin{center}
\begin{tabular}{cc}
  \epsfxsize=6.8cm
  \hspace{-2mm} \epsffile[32     9   565   429]{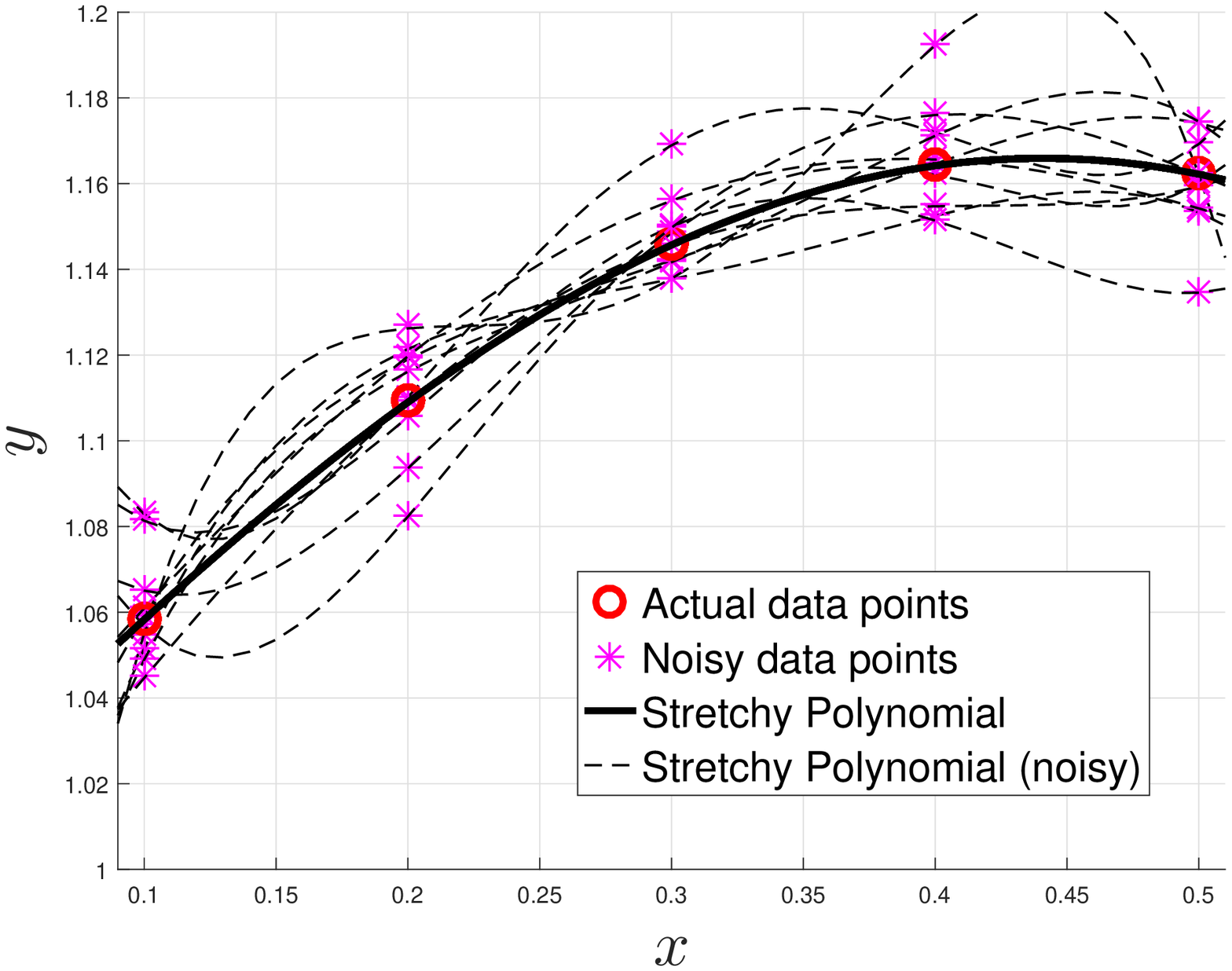}
  &
  \epsfxsize=6.8cm
  \hspace{6mm} \epsffile[25     9   517   401]{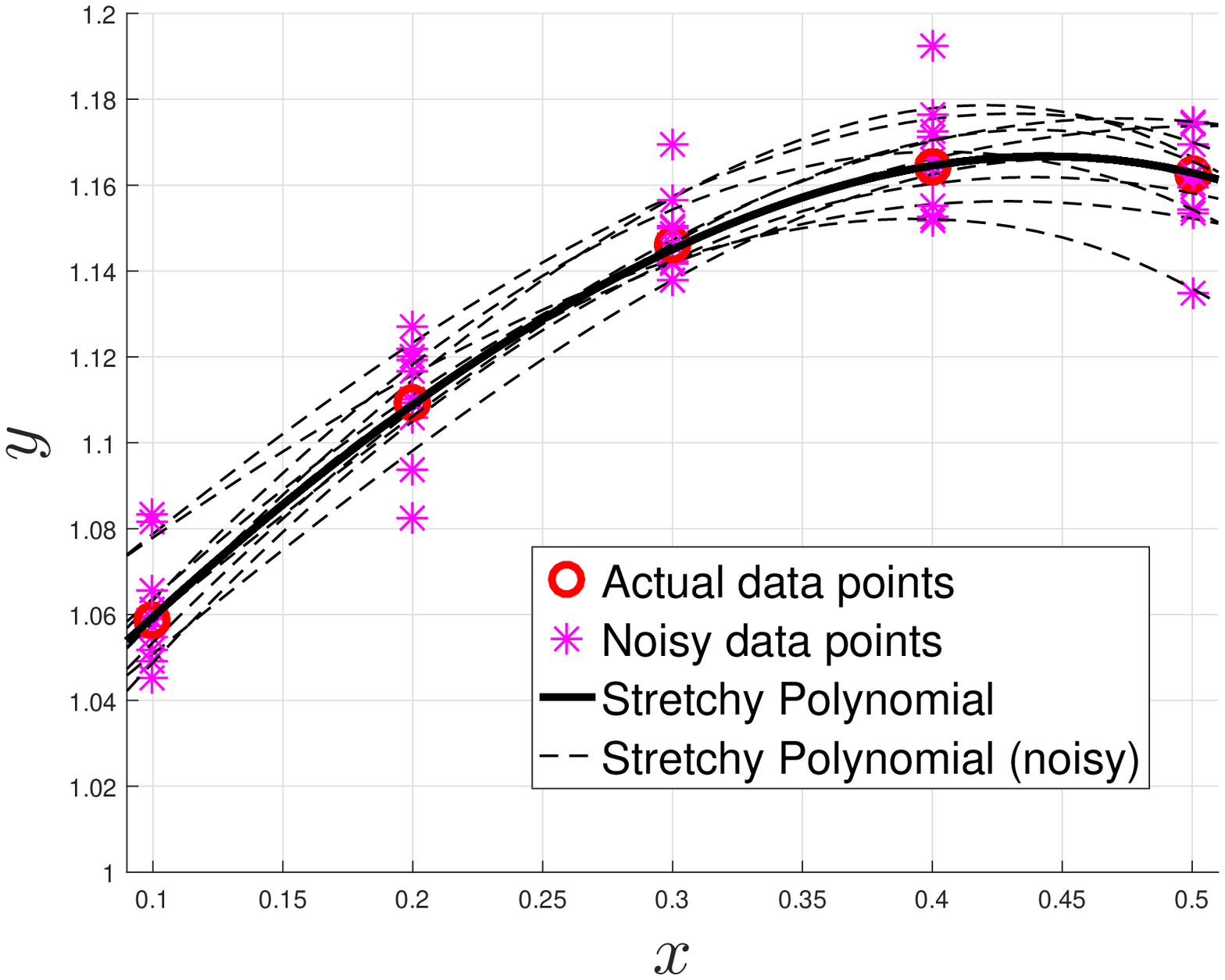}

  \\*[2mm]  (a) ${k}=1.2$, ${c}=10^{15}$ & (b) ${k}=1.8$, ${c}=10^4$
\end{tabular}
  \caption{Learned outputs from noisy inputs at different ${k}$ and ${c}$ values.}
  \label{fig_contours1}
  \end{center}
\end{figure}

Consider the first example, which contains five single-dimensional training samples (see
the red circles in Fig.~\ref{fig_contours1}). These data points are generated using $y=1
+ 0.6x - 1.5x^3 + 0.8x^4$ based on $x\in\{0.1,0.2,0.3,0.4,0.5\}$. A polynomial model of
$10^{th}$-order including the intercept term is adopted to fit these data points using
the proposed stretchy regression. The solid lines in Fig.~\ref{fig_contours1}(a)-(b)
shows the learned outputs at different $k$ and ${c}$ values. These results show good
fitting when no noise is considered, i.e., the solid line passes through all data points
marked by the red circles. Apart from the noiseless case, the learned outputs (marked by
dashed lines) from ten `noisy' measurements (marked by asterisks) are included in each
figure. Here, due to the excessive number of parameters over the number of data samples,
stretchy regression attempts to fit all the `noisy' data points when ${c}$ is large
(e.g., at negligible regularization such as ${c}=10^{15}$ in
Fig.~\ref{fig_contours1}(a)). At $c=10^4$, the effect of regularization is observed to
have a smoother fit where the noisy data points are not fully attempted in
Fig.~\ref{fig_contours1}(b).

Table~\ref{table_coefficients1} shows the variation of learned polynomial coefficients
($\hat{\balpha}=[\hat{\alpha}_0,...,\hat{\alpha}_{10}]^T$) at different ${k}$ and ${c}$
values. The results for both noiseless and noisy cases show convergence to a sparse
solution when ${k}$ approaches $1$. Particularly, at $k=1.2$, the reconstructed
polynomial coefficients for the noiseless case with weak regularization (at
${c}=10^{15}$, marked in boldface) compare relatively well with the ground truth
($\balpha=[1, 0.6, 0, -1.5, 0.8, 0, \cdots,0]^T$). For stronger regularization at
${c}=10^{4}$, The estimation becomes smoother as evident from the low standard
deviations. However, this comes at the price of having a lower fidelity to coefficient
reconstruction. This is due to the heavier weight in regularization than error
minimization.

\begin{table}[hhh]
\caption{Average polynomial coefficient values estimated over 10 noiselsess/noisy samples
at different ${k}$ and ${c}$ values}\label{table_coefficients1} \centering {\scriptsize
\begin{tabular}{|c|c|c||c|c|}
\hline
                 & ${k}:1.8$, ${c}:10^{4}$ & ${k}:1.2$, ${c}:10^{4}$ & ${k}:1.8$, ${c}:10^{15}$ &  ${k}:1.2$, ${c}:10^{15}$ \\
                 &  noiseless/noisy(std) & noiseless/noisy(std)  & noiseless/noisy(std)  &  noiseless/noisy(std) \\
\hline
$\hat{\alpha}_0$ &  1.000/ 1.001( 0.019) &  1.063/ 1.067( 0.014) &  0.999/ 1.006( 0.186) & {\bf 1.000}/ 1.008( 0.237) \\
$\hat{\alpha}_1$ &  0.641/ 0.640( 0.142) &  0.234/ 0.226( 0.045) &  0.626/ 0.544( 3.033) & {\bf 0.602}/ 0.517( 4.181) \\
$\hat{\alpha}_2$ & -0.402/-0.373( 0.380) & -0.046/-0.052( 0.018) & -0.198/ 0.118(14.957) & {\bf -0.014}/ 0.331(24.060) \\
$\hat{\alpha}_3$ & -0.340/-0.382( 0.171) & -0.002/-0.002( 0.001) & -0.826/-1.134(24.053) & {\bf -1.457}/-1.854(55.477) \\
$\hat{\alpha}_4$ & -0.179/-0.227( 0.250) & -0.000/-0.000( 0.000) & -0.140/-0.304( 1.298) & {\bf 0.738}/ 0.680(43.384) \\
$\hat{\alpha}_5$ & -0.083/-0.114( 0.174) & -0.000/-0.000( 0.000) &  0.219/ 0.210(12.138) & {\bf 0.033}/ 0.031( 1.944) \\
$\hat{\alpha}_6$ & -0.036/-0.053( 0.097) & -0.000/-0.000( 0.000) &  0.244/ 0.283(12.018) & {\bf 0.001}/ 0.001( 0.067) \\
$\hat{\alpha}_7$ & -0.015/-0.024( 0.049) & -0.000/-0.000( 0.000) &  0.168/ 0.204( 8.022) & {\bf 0.000}/ 0.000( 0.002) \\
$\hat{\alpha}_8$ & -0.007/-0.011( 0.023) & -0.000/-0.000( 0.000) &  0.095/ 0.118( 4.504) & {\bf 0.000}/ 0.000( 0.000) \\
$\hat{\alpha}_9$ & -0.003/-0.005( 0.011) & -0.000/-0.000( 0.000) &  0.049/ 0.061( 2.297) & {\bf 0.000}/ 0.000( 0.000) \\
$\hat{\alpha}_{10}$ & -0.001/-0.002( 0.005) & -0.000/-0.000( 0.000) &  0.024/ 0.030( 1.104) & {\bf 0.000}/ 0.000( 0.000) \\
\hline
\end{tabular} }
\end{table}

\begin{figure}[hhhh]
  \begin{center}
  \begin{tabular}{cc}
  \epsfxsize=7.3cm
  \hspace{-5mm}\epsffile[0     0   628   445]{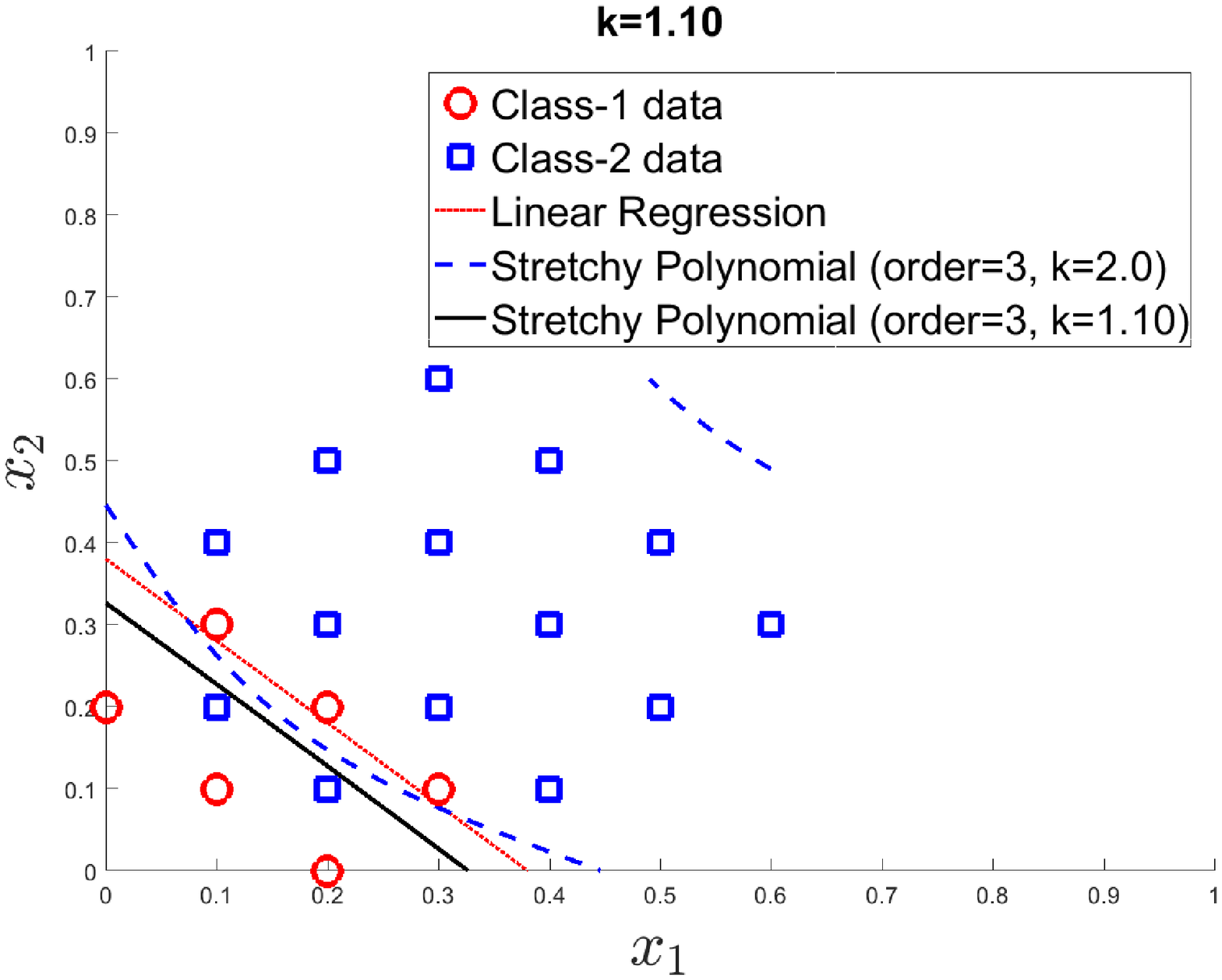} &
  \epsfxsize=5.8cm
  \hspace{-3mm}\epsffile[28     9   514   414]{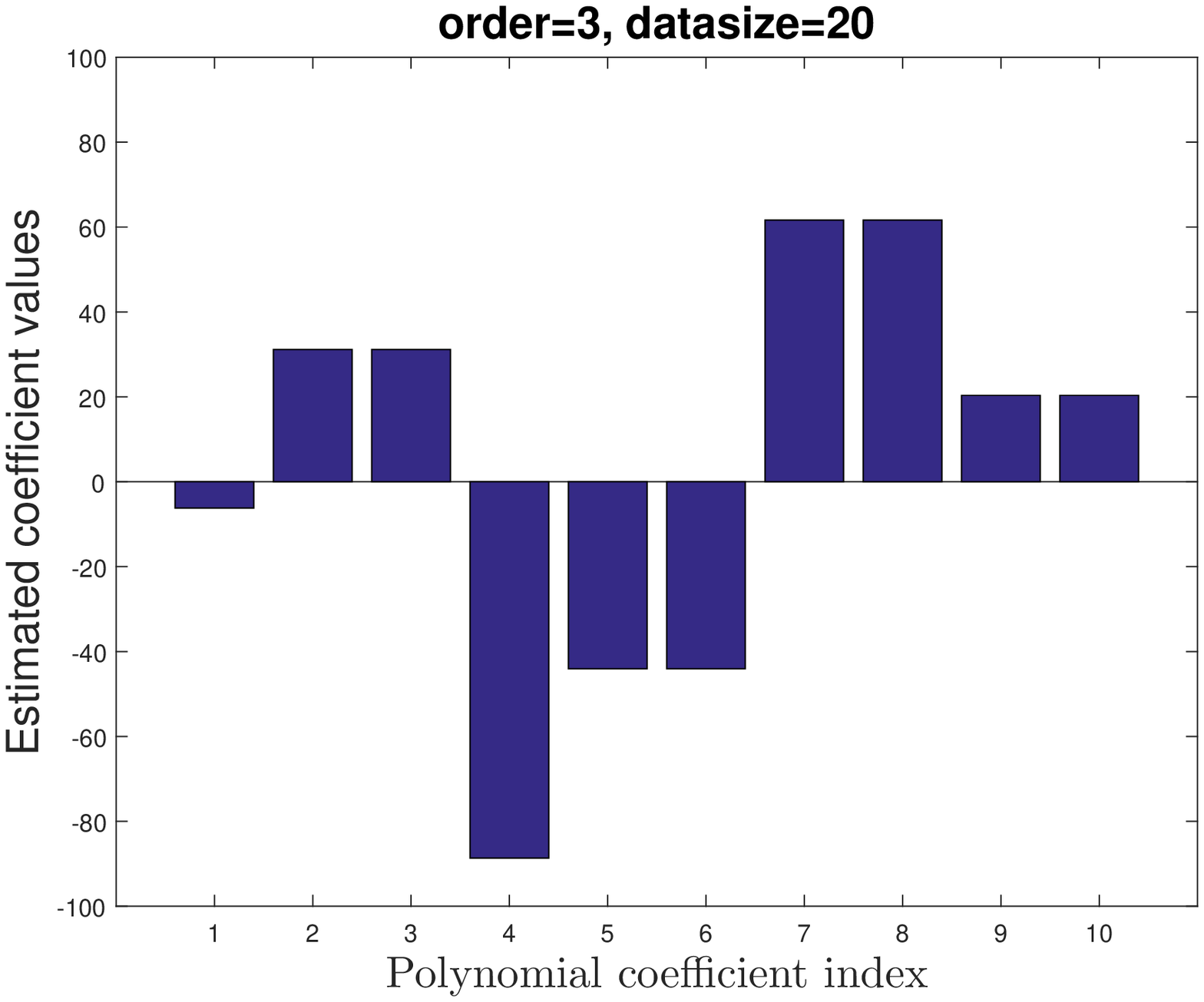}
  \\*[1mm]  (a)   & (b)
  \end{tabular}
  \caption{(a) Decision boundaries (at zero threshold level) of a linear model
  and a 3rd-order polynomial model learned from 20 data points at different ${k}$ values
  (all at ${c}=10^{100}$);
  (b) Estimated polynomial coefficient values at ${k}=1.1$.}
  \label{fig_contours3}
  \end{center}
\end{figure}

In the second example, we consider the case of 20 two-dimensional data points with a few
of them overlapping in class distribution. This constitutes an over-determined system
when a 3rd-order bivariate polynomial model with 10 parameters is adopted.
Fig.~\ref{fig_contours3}(a) shows the decision boundary plots of the polynomial model
learned at $k=1.1$ and ${k}=2$ when ${c}=10^{100}$ (i.e., without regularization).
Fig.~\ref{fig_contours3}(b) shows the estimated coefficients
({$\hat{\balpha}=[\hat{\alpha}_0,...,\hat{\alpha}_9]$}) at ${k}=1.1$. This result,
however, does not show convergence to sparse solution when $k=1.1$. This is in congruence
to our observation of the analysis of variance for over-determined systems.

\begin{figure}[hhhh]
  \begin{center}
  \begin{tabular}{cc}
  \epsfxsize=7.3cm
  \hspace{-5mm}\epsffile[0     0   642   446]{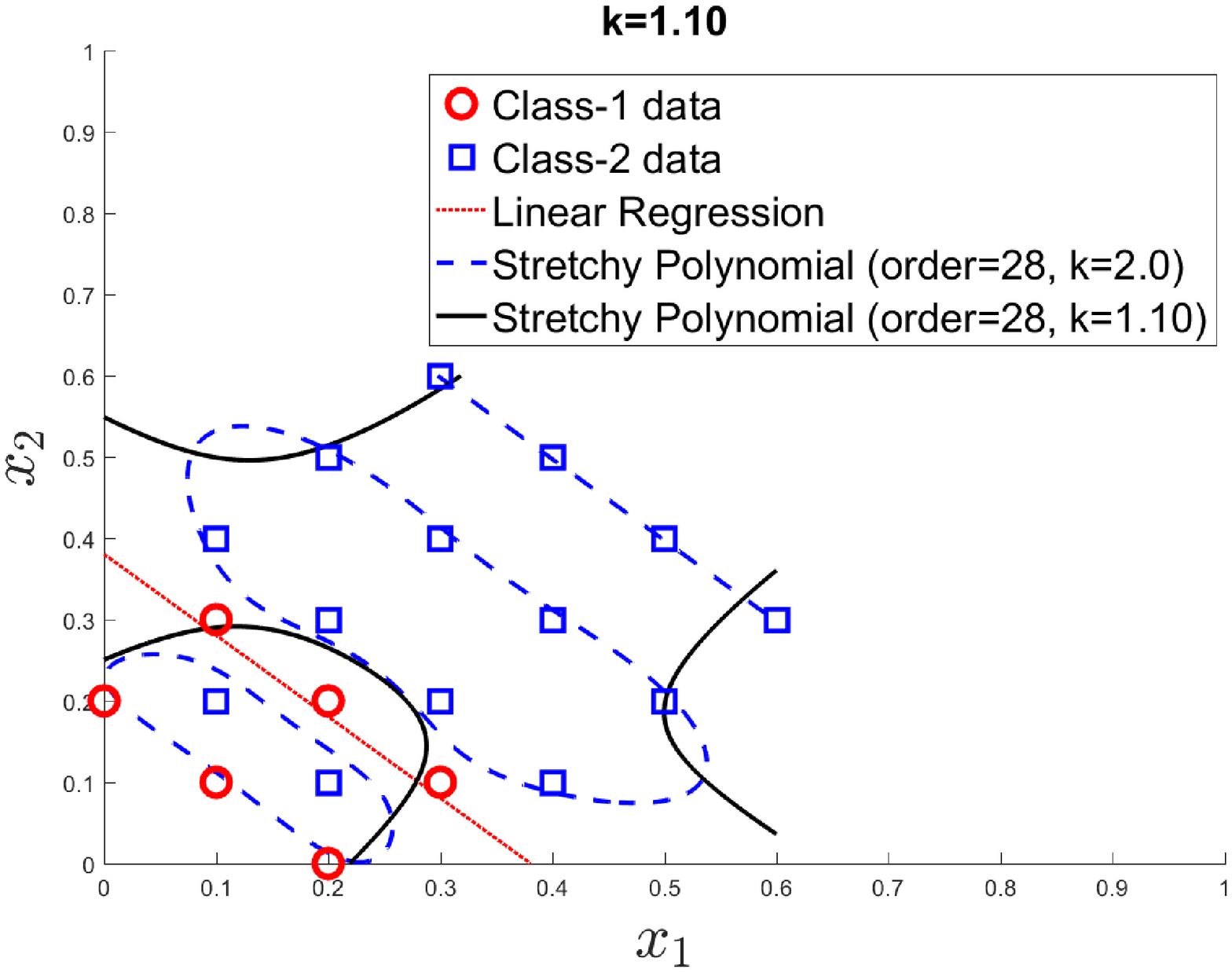} &
  \epsfxsize=7.1cm
  \hspace{-3mm}\epsffile[27     3   545   355]{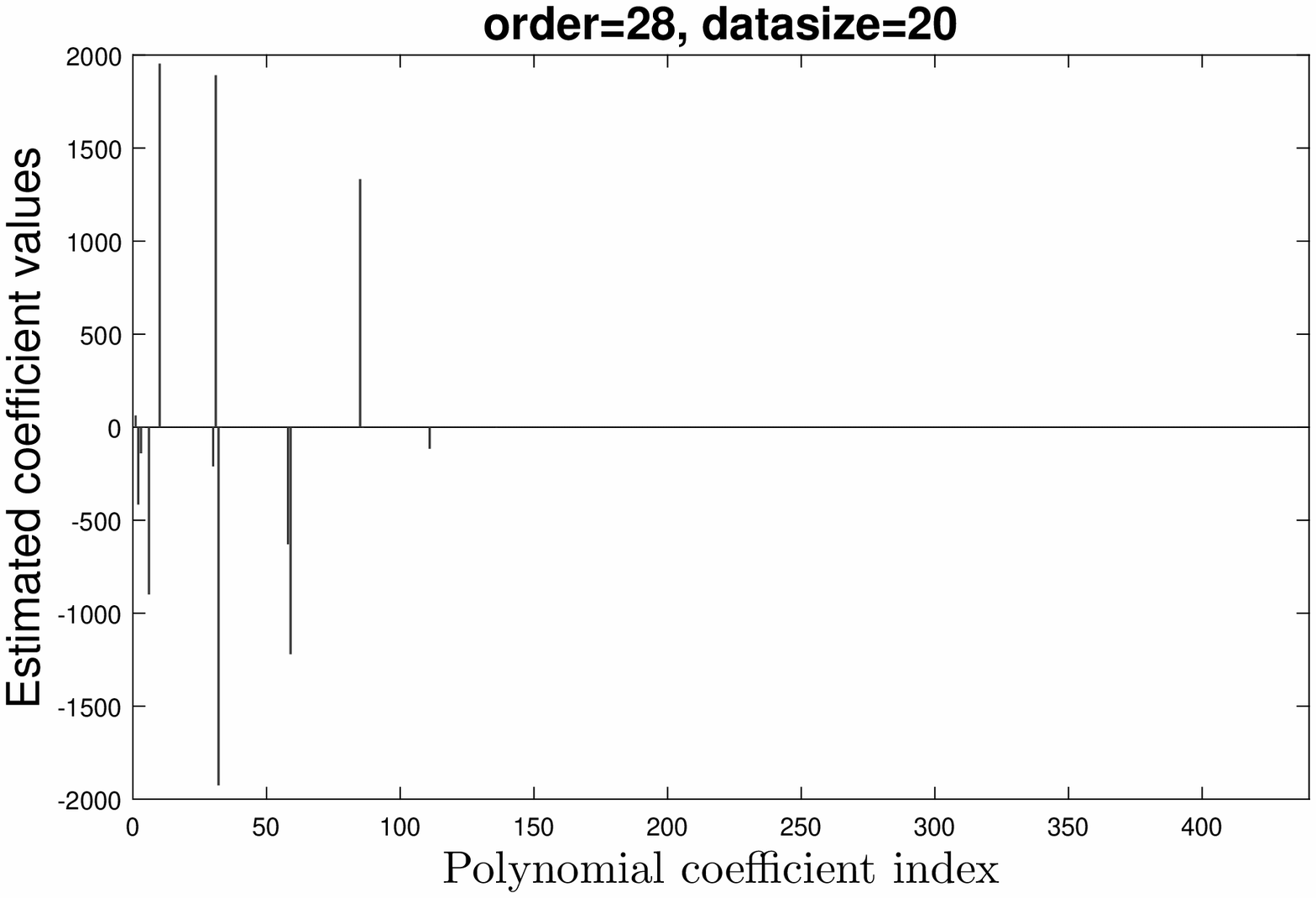}
  \\*[1mm]  (a)   & (b)
  \end{tabular}
  \caption{(a) Decision boundaries (at zero threshold level) of a linear model
  and a 28th-order polynomial model learned from 20 data points at different ${k}$ values
  (all at ${c}=10^{100}$);
  (b) Estimated polynomial coefficient values at ${k}=1.1$.}
  \label{fig_contours_20data_28th_order}
  \end{center}
\end{figure}

Finally, the bivariate polynomial model is raised to 28th-order, where the total number
of parameters to be estimated is 435. This turns the system an under-determined one since
there are more parameters than data samples. Fig.~\ref{fig_contours_20data_28th_order}(a)
shows the decision contours for the polynomial model learned at $k=1.1$ and $k=2$ which
are plotted together with the contour of the linear model. Here, the SR at $k=1.1$ gives
4 error count while the SR at $k=2$ gives 0 error count. Although the SR at $k=2$
produces no error, the decision contour becomes overly complex. This leads to an
over-fitting scenario if the underlying distributions are not complex.
Fig.~\ref{fig_contours_20data_28th_order}(b) shows the sparseness of the estimated
parameters at ${k}=1.1$. This example shows the applicability of SR in problems
characterized by high-dimensional features with small number of samples.

To summarize, the above synthetic examples illustrate stretching of relatively irrelevant
weight coefficients towards zero for under-determined systems when $k\rightarrow 1$ while
maintaining the necessary mapping. For over-determined systems, the stretching does not
show sparseness when $k\rightarrow 1$. Apart from the much desired compressed estimation,
the stretching capability offers beyond the fixed norm minimization which is particularly
useful for numerical adaptation in practice. For example, when encountering numerical
stability issue at $k=1.1$ for the given data set, a slightly higher value of $k$ (such
as $k=1.2$) is yet available for a less compressed solution which comes with better
numerical stability. In a nutshell, the stretching capability offers an adjustable
mechanism for practicality. The effects of stretching on the estimation accuracy at
various $k$ value settings are also observed in the following experiments on physical
data.

\section{Experiments on Physical Data} \label{sec_expts}

Apart from the regression problems, this study is extended to several binary
classification problems to illustrate the applicability of the proposed method on
physical data. Our first goal here is to observe whether the stretching can be applicable
to physical data of high input dimensionality where matrix ill-conditioning is a commonly
encountered problem. After verifying the applicability of parametric stretching, our
subsequent goals are to observe whether the prediction accuracy can be impaired by the
stretchy estimation and whether the training processing time is largely affected.\\

\subsection{Data Sets}

\subsection*{(i) Regression Data Sets}

Five real-world regression problems from the public domain are utilized for this study.
While the first two data sets represent over-determined systems, the remaining three data
sets represent under-determined systems. The prostate cancer data set was adopted in
\cite{Hastie1}, which came from a study in \cite{Stamey1}. In this data set, the
correlation between the level of prostate-specific antigen and several clinical measures
was studied. The diabetes data set was adopted in \cite{Efron2}, where ten baseline
variables were obtained for each of 442 samples with a quantitative measure of disease
progression one year after baseline. The corn data set consists of 80 samples of corn
measured on three different NIR spectrometers with 700 channels \cite{Corn1,FuGH1}. As a
representative case in this study, we use m5spec to regress upon the moisture response.

E2006-TFIDF \cite{Kogan1} is a text regression problem for predicting a real-world
continuous quantity associated with the text's meaning. In this study, the classic TFIDF
(term-frequency/inverted-document-frequency) metric related to the frequency of
occurrence of certain words within a document is adopted. The data set, which is of
150,360 dimensions, contains 3,308 samples for test and 16,087 samples for training. Due
to our constraint in computing facilities, only the test set is utilized under a 2-fold
cross-validation setting. Finally, we use the Columbia horizontal gaze data set for
regression evaluation \cite{Smith1}. Each image has a resolution of 5,184 $\times$ 3,456
pixels. Again, due to the limitation of our computing facilities, only one-third (1960
samples) of the entire data (5880 samples) are utilized in this 2-fold evaluation. To
further reduce the computational load, all the images are cropped to 1,200 $\times$ 1,000
pixels by removing much of the background. Each image is reshaped to a row vector of 1.2
million dimensions and stacked as a regressor matrix. Because the images contain only
positive values normalized within [0,1], no further transformation is employed. A summary
of these regression data sets is provided in Table~\ref{Regression_data_sets}.

\begin{table}[hhh] \caption{{Regression data sets}}\label{Regression_data_sets}
\begin{center} {\small
\begin{tabular}{|c|c|r|r|}
  \hline
  Dataset & Domain & \# Feat. & \# Samples  \\
  \hline
  Prostate  & Medical           &     8     &   97  \\
  Diabetes  & Medical           &    10     &  442  \\
  Corn      & Spectrometry      &   700     &   80  \\
  TFIDF     & Text              & 150,360   & 3,308 \\
  Gaze      & Image             & 1,200,000 & 1,960 \\
  \hline
\end{tabular}}
\end{center}
\end{table}

\subsection*{(ii) Binary Classification Data Sets}

The classification problems of the NIPS feature selection challenge
\cite{NIPS2003,NIPS2004} are utilized for this extended study. This challenge was
introduced in 2003 with five high dimensional data sets, aiming to find feature selection
algorithms that significantly outperform methods using all features. Each data set {was}
partitioned into training, validation and test sets where the users {were} provided only
the labels of training and validation sets together with their respective input features
at the initial stage of challenge. Although the challenge ended in December 2003, the
data sets are nevertheless available for evaluation and benchmarking of algorithms with
only the test set labels withheld by the organizer. Table~\ref{NIPS_data_sets} provides
an overview of the data sets in terms of the sizes of data and their feature dimension.
Because the test data set is not available, we perform 10 trials of 2-fold
cross-validation tests on all the available data, formed by combining the training set
and the validation set.

\begin{table}[hhh]
\caption{NIPS feature selection challenge data sets}\label{NIPS_data_sets}
\begin{center} {\small
\begin{tabular}{|c|c|c|r|r|r|r|}
  \hline
  Dataset & Domain & Type & \# Feat. & \# Trn & \# Valid. & \# Test \\
  \hline
  Arcene    & Mass spec.        & Dense     &  10000 &  100 &  100 &  700 \\
  Dexter    & Text categ.       & Sparse    &  20000 &  300 &  300 & 2000 \\
  Dorothea  & Drug discov.      & Sp. bin.  & 100000 &  800 &  350 &  800 \\
  Gisette   & Digit recog.      & Dense     &   5000 & 6000 & 1000 & 6500 \\
  Madelon   & Artifical data    & Dense     &    500 & 2000 &  600 & 1800 \\
  \hline
\end{tabular}}
\end{center}
\end{table}

\subsection{Evaluation Settings}

To deal with inevitable statistical variations in experiments, we perform 10 trials of
2-fold tests on each data set. In terms of the accuracy measurement, the average mean
squared error (MSE) of fitting is adopted for the regression data sets. For the
classification data sets, the classification Balanced Error Rate (BER) is recorded in
terms of its average values taken from these 10 trials of two-fold tests. For both the
regression and the classification data sets, the average training CPU time in seconds
(CPU) are recorded from the two-fold training process. The balanced error rate is defined
as the average of the error rates of the positive class and the negative class. According
to the provider of these data sets, this metric was used because some data sets are
imbalanced. All experiments were conducted on a 64-bit Intel Core i7-3.6GHz processor
with 16GB RAM, except for the TFIDF and Gaze data sets, which were evaluated on a 64-bit
Intel Xeon E5-v2-3.5GHz processor with 64GB RAM.

Based on our empirical observation, a transformation using $a\in[-1,+1]$, $a\neq 0$ for
most regression applications and at $a\in[-10,0)$ for most classification applications,
all without inclusion of an offset ($\bm{b}={\bf 0}$, see
\eqref{eqn_exponential_transform}) produces reasonable results. Without diverting our
focus towards data normalization issues, except for the Gaze data which used a normalized
image with feature values falling within interval [0,1], all data sets are standardized
for the proposed stretchy regression using a nominal setting of $a=-0.2$ and $\bm{b}={\bf
0}$ for positive quadrant transformation. For benchmarking purpose, the well-known LASSO
\cite{Tibshirani1} and Elastic-net \cite{Zou1} are included in the experiments alongside
with the proposed stretchy regression. For convenience, we abbreviate the proposed
stretchy regression as SR, and for LASSO-Elastic-net, the LASSO function \cite{Matlab} is
adopted where it will be abbreviated as LASSO. This LASSO function traverses between
LASSO ($\texttt{Alpha}=1$) and Elastic-net ($0<\texttt{Alpha}<1$) at various
\emph{elastic net values} (called \texttt{Alpha}, which controls the relative balance
between $\ell^1$ and $\ell^2$ penalty) settings.

In order to gather a good picture regarding the sensitivity and performance over
different algorithmic settings, the compared algorithms are evaluated under various
parameter configurations. For the proposed algorithm, several stretching power values at
${k}\in\{1.1,1.25,1.5,1.75,2\}$ are experimented at regularization values
${c}\in\{10,10^2,10^3,10^4,10^8,10^{100}\}$ according to \eqref{eqn_k_norm_solution_reg}.
We note that \eqref{eqn_k_norm_solution} is adopted for the case when ${c}=10^{100}$. For
LASSO, the following elastic net mixing values are adopted:
\texttt{Alpha}$\in\{1,0.75,0.5,0.25,0.0001\}$. Here we note that $\texttt{Alpha}=0$ is
not allowed in the algorithm and it is equivalent to the ordinary $\ell^2$ regression and
$k=2$ in our SR implementation. The elastic net mixing value \texttt{Alpha} in LASSO will
be plotted alongside with $k$ values in SR in the experiments since they are related to
elastic parametric solution even though not in exact sense due to their different
optimization objectives.

Apart from \texttt{Alpha}, another setting for for LASSO is the \emph{sequence of lambda
penalties} which could affect the convergence of the solution. Here, the \emph{number of
lambda values} \texttt{NumLambda} is configured at 1, 2, 100 {and a CV (cross-validated)
search within the set $\{1,2,5,10,20,30,40,$ $60,80,100\}$} to observe its impact on the
training CPU time and testing accuracy. Particularly, \texttt{NumLambda}=1 and
\texttt{NumLambda}=CV explore the impact of having the lowest number of lambda penalties
versus that selected from a large number of lambda penalties. We abbreviate these
settings as LASSO1, LASSO2, LASSO100 {and LASSOCV}. For \texttt{NumLambda}$>1$, only the
best accuracy in terms of lowest mean squared error is recorded.

Based on the above settings, both SR and LASSO are evaluated. Since the estimation by SR
gives a smooth range of parametric values instead of a crisp set of sparse parameters,
the proposed SR adopts a two-pass training process wherein the top parameter values from
first-pass training are selected for a second-pass training according to user-specified
feature density. For instance, for SR at 0.1 feature density, only the highest 10\% of
parameters, obtained from the first-pass training that utilizes all features, are
selected for second-pass training. To observe the impact of sparse features for high
dimensional data, we perform experiments on several feature density values, namely SR0.5,
SR0.1 and SR0.01 together with the original SR1.0 with full features.

\subsection{Results: Estimated Parameters}

\begin{figure}[hhhh]
  \begin{center}
\begin{tabular}{cc}
  \epsfxsize=6.8cm
  \epsfysize=3.9cm
  \hspace{-2mm}\epsffile[52   199   545   590]{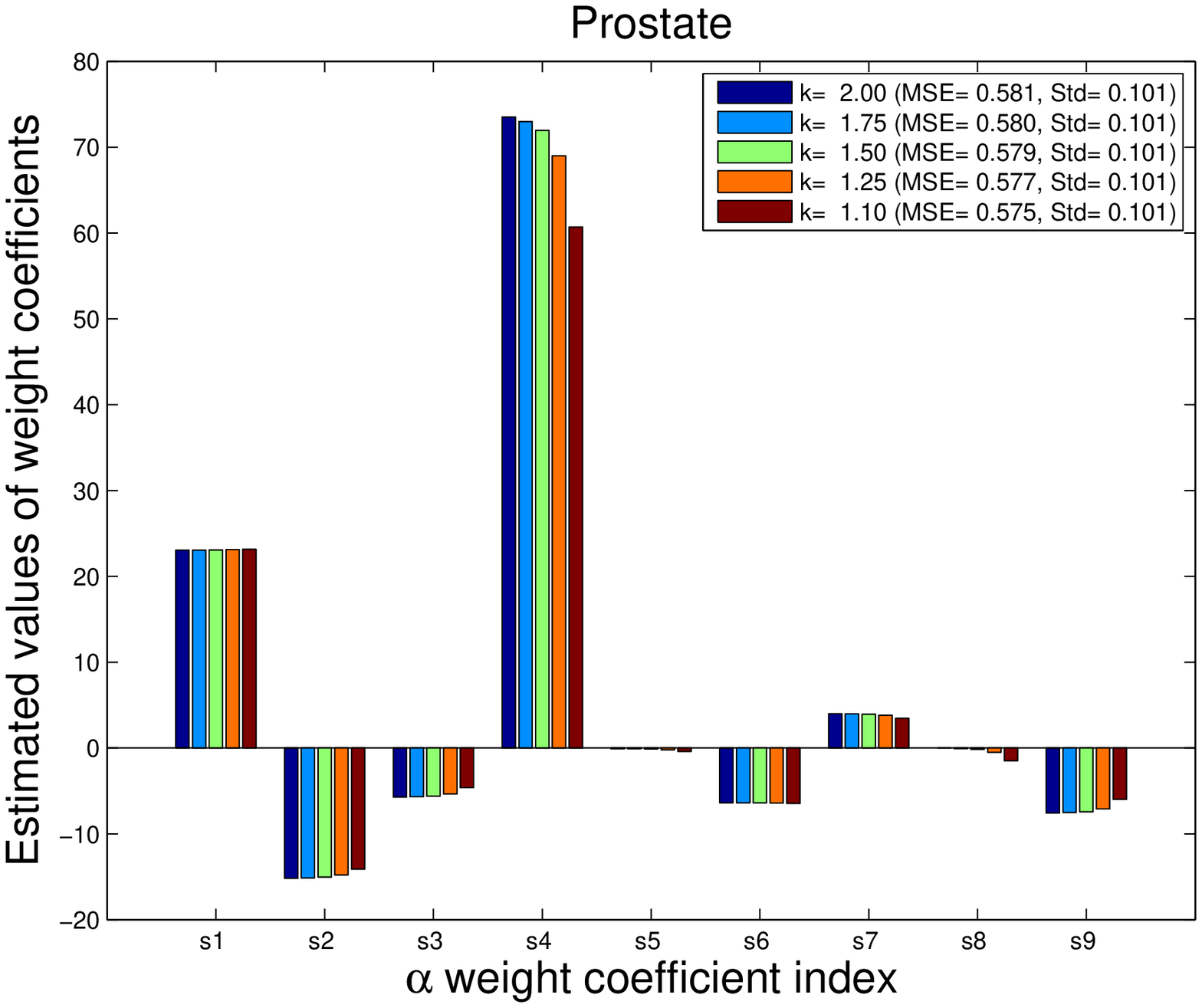} &
  \epsfxsize=6.8cm
  \epsfysize=3.9cm
  \hspace{0mm}\epsffile[52   199   545   590]{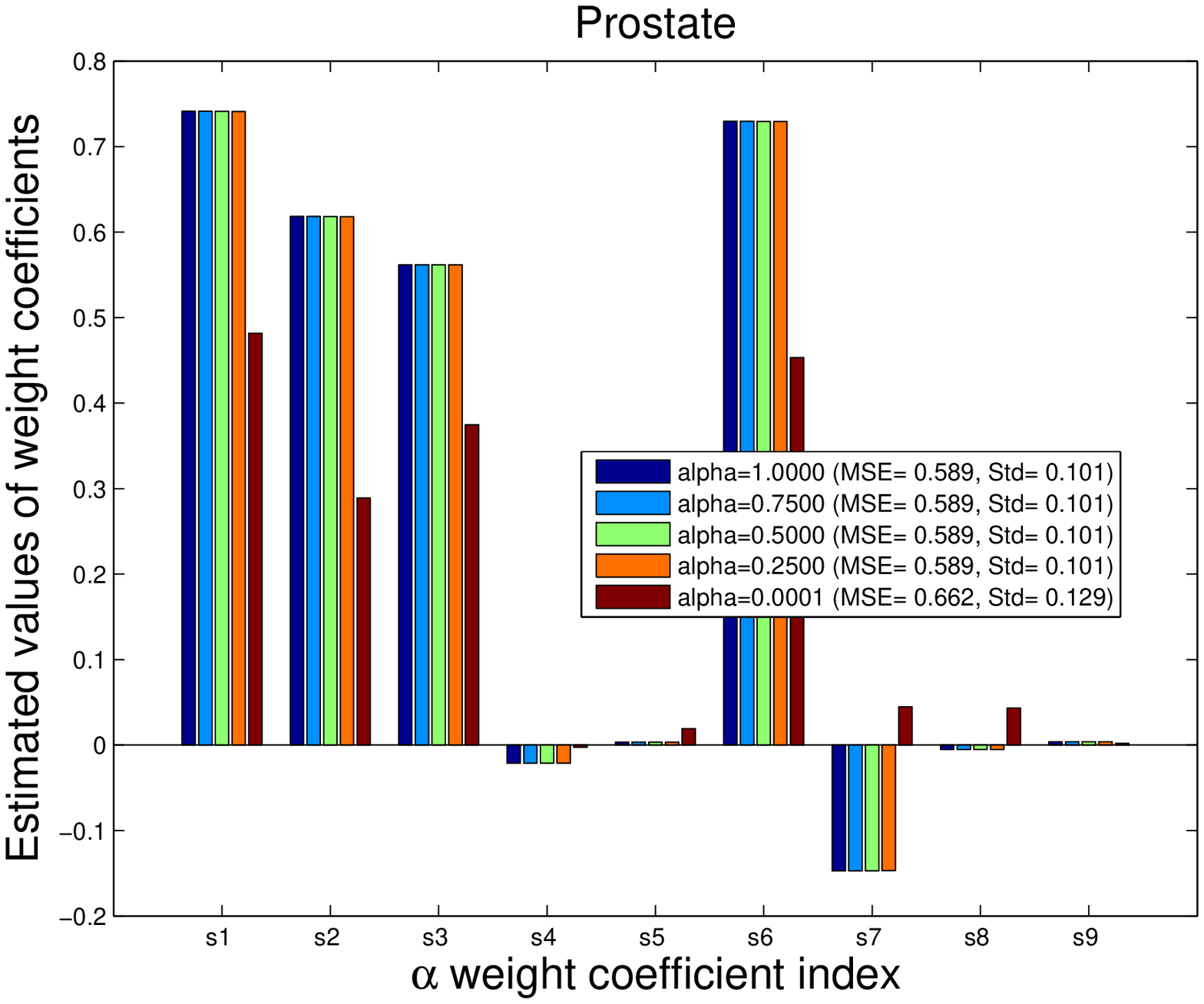}
  \\  \small{Prostate:  (a) SR}  & \small{(b) LASSO}
  \\  \small{$k\in\{1.1,1.25,1.5,1.75,2\}$}  & \small{\texttt{Alpha}$\in\{1,0.75,0.5,0.25,0.0001\}$} \\*[2mm]
  \epsfxsize=6.8cm
  \epsfysize=3.9cm
  \hspace{-2mm}\epsffile[52   199   545   590]{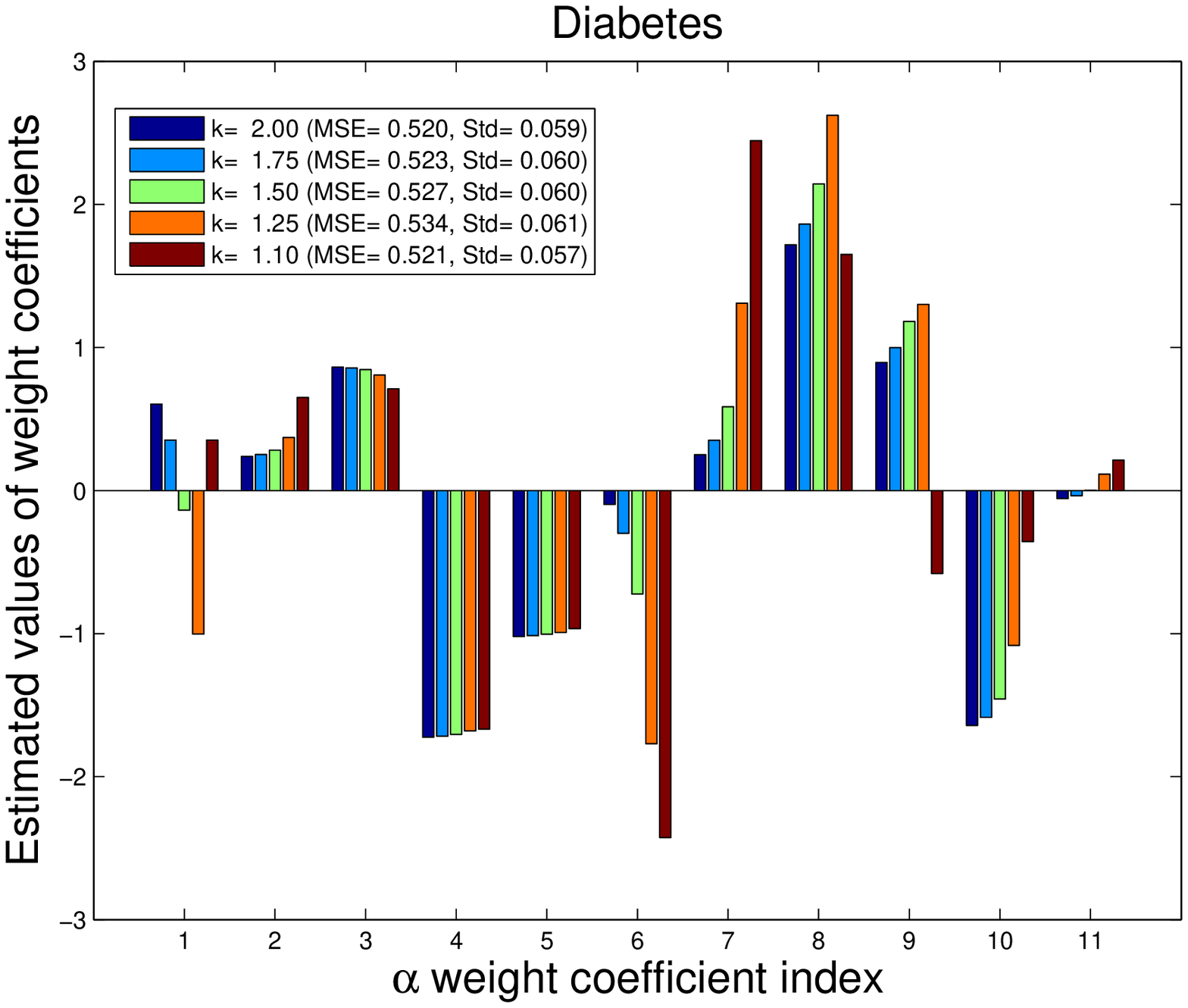} &
  \epsfxsize=6.8cm
  \epsfysize=3.9cm
  \hspace{0mm}\epsffile[52   199   545   590]{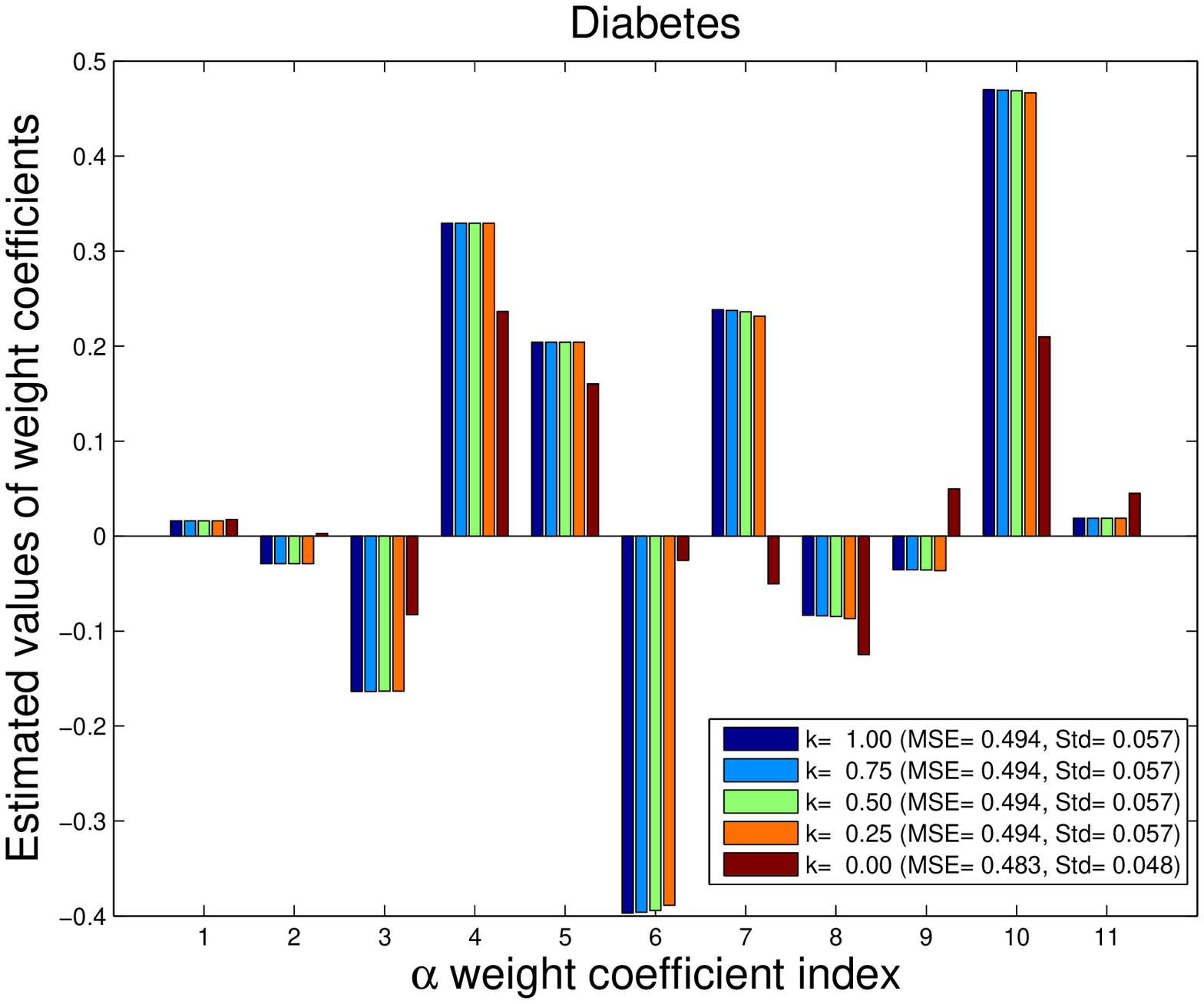}
  \\  \small{Diabetes: (c) SR}  & \small{(d) LASSO}
  \\  \small{$k\in\{1.1,1.25,1.5,1.75,2\}$}  & \small{\texttt{Alpha}$\in\{1,0.75,0.5,0.25,0.0001\}$} \\*[2mm]
  \epsfxsize=6.8cm
  \epsfysize=3.9cm
  \hspace{-2mm}\epsffile[52   199   545   590]{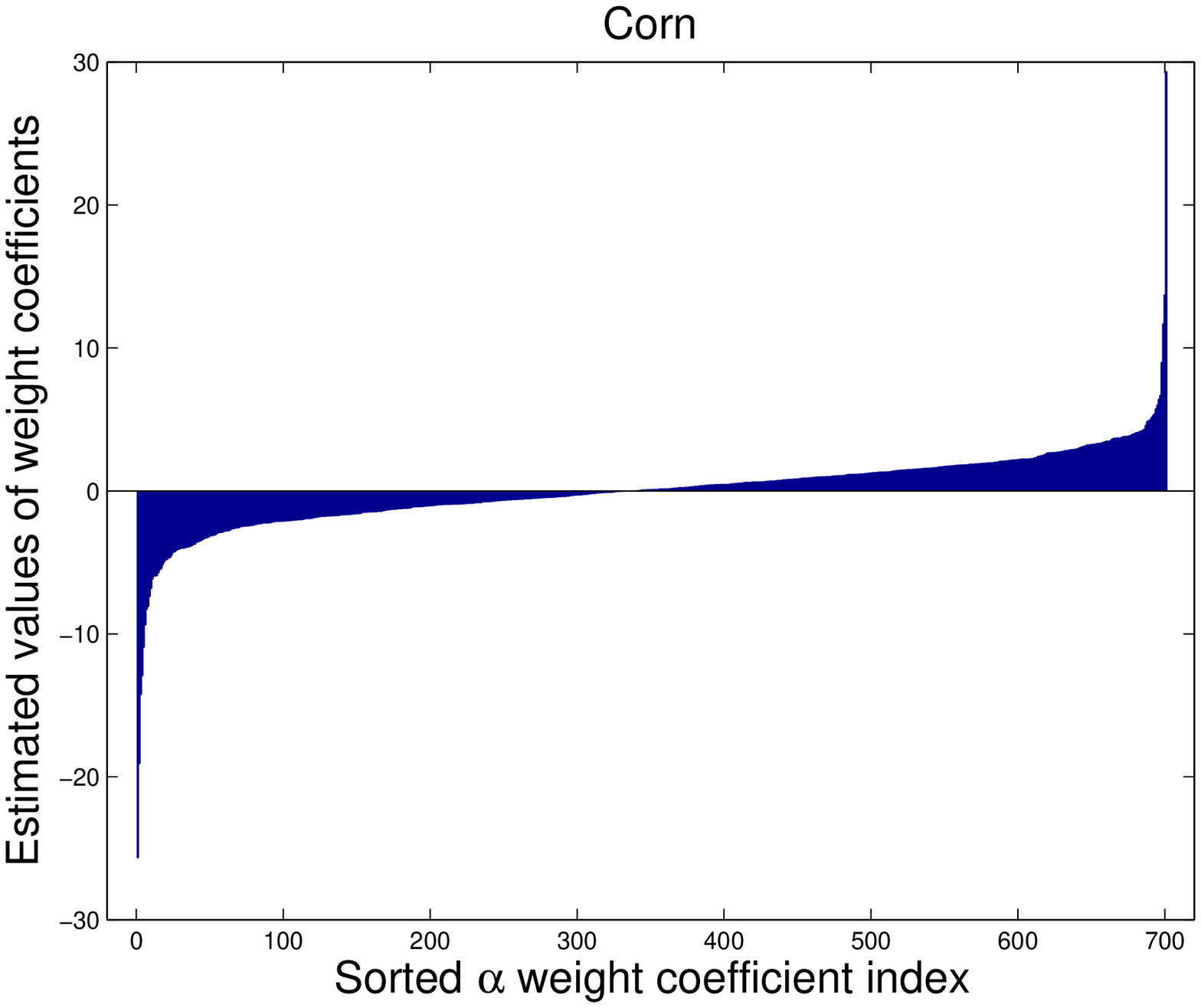} &
  \epsfxsize=6.8cm
  \epsfysize=3.9cm
  \hspace{0mm}\epsffile[52   199   545   590]{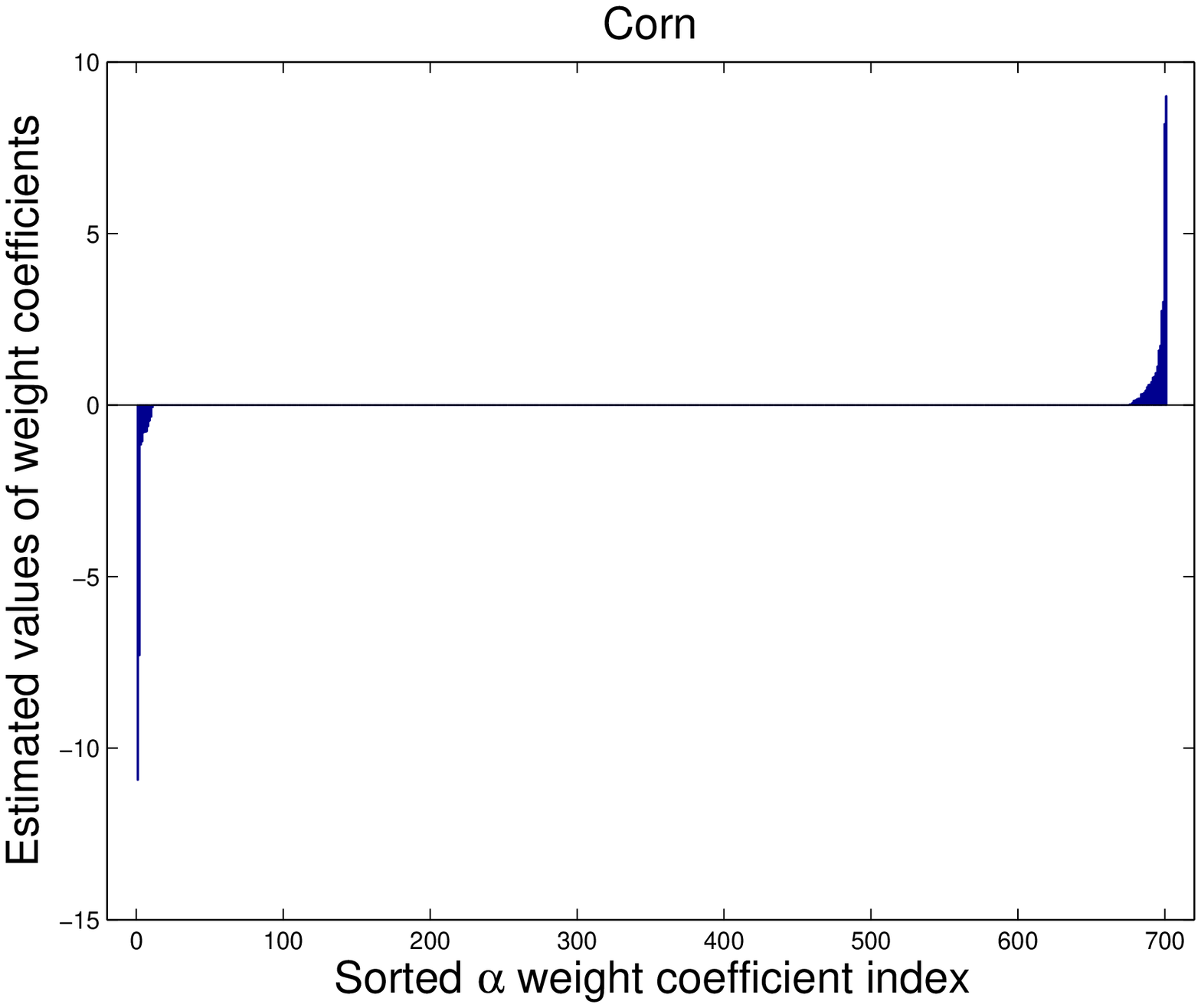}
  \\  \small{Corn: (e) SR at $k=1.1$}  & \small{(f) LASSO at \texttt{Alpha}=1} \\*[2mm]
  \epsfxsize=8.0cm
  \epsfysize=3.9cm
  \hspace{-2mm}\epsffile[67    49   470   352]{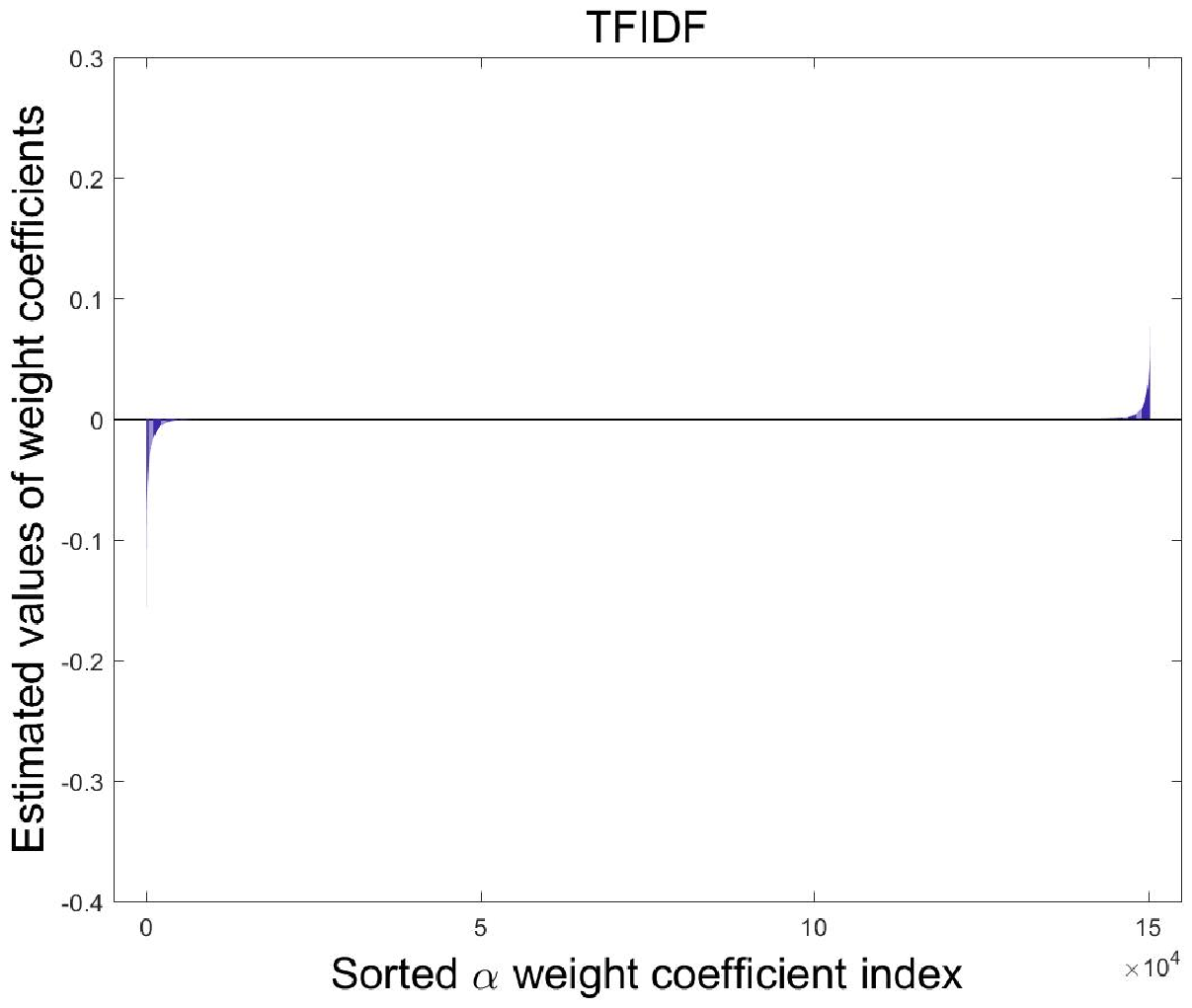} &
  \epsfxsize=8.0cm
  \epsfysize=3.9cm
  \hspace{0mm} \epsffile[62    64   515   404]{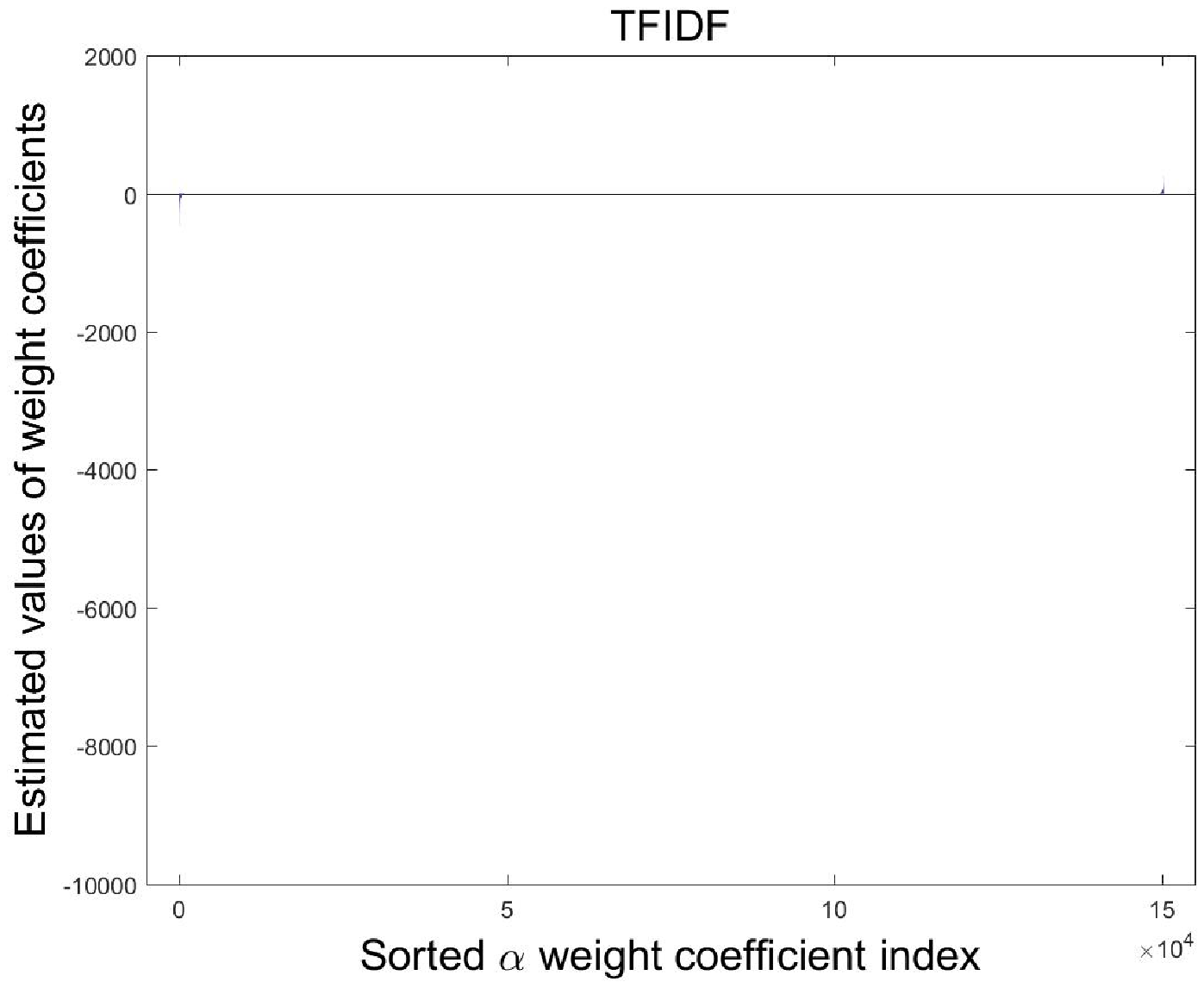}
  \\  \small{TFIDF: (g) SR at $k=1.25$}  & \small{(h) LASSO at \texttt{Alpha}=1} \\*[2mm]
  \epsfxsize=8.0cm
  \epsfysize=3.9cm
  \hspace{-2mm}\epsffile[62    64   515   404]{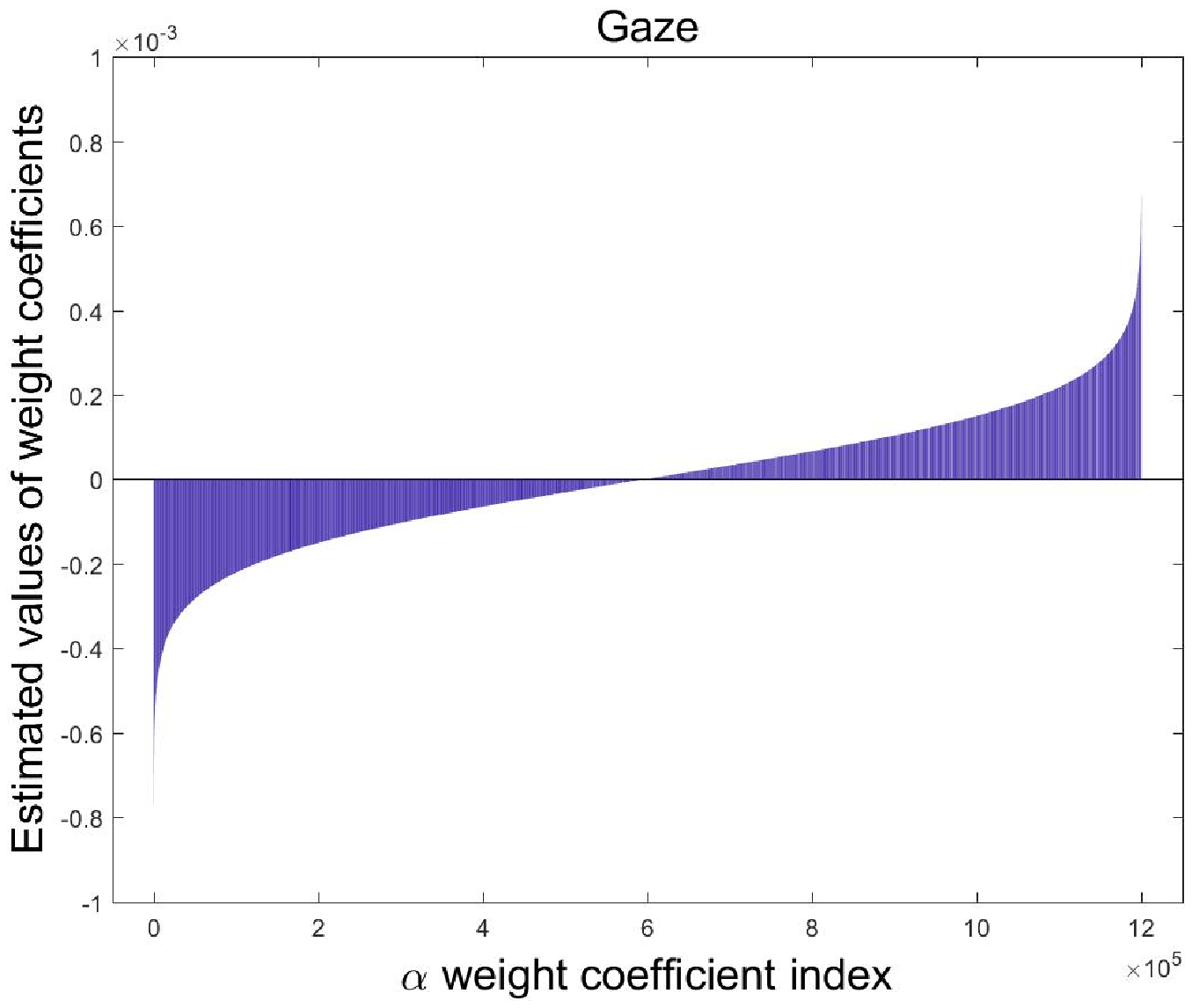} &
  \epsfxsize=8.0cm
  \epsfysize=3.9cm
  \hspace{0mm}\epsffile[62    64   515   404]{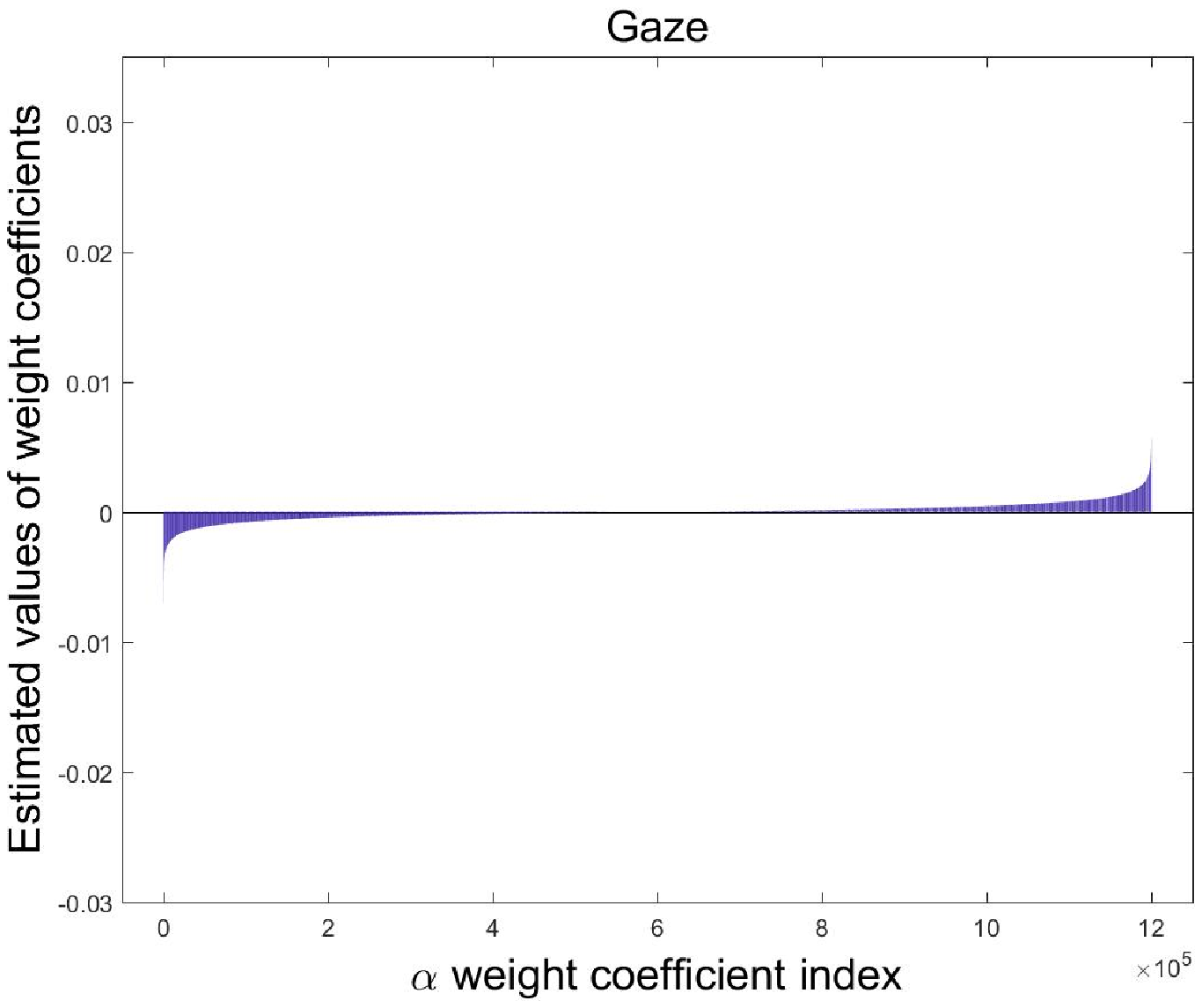}
  \\*[1mm]  \small{Gaze: (i) SR at $k=1.25$}  & \small{(j) LASSO at \texttt{Alpha}=1}
\end{tabular}
  \caption{Regression: sorted parameter $\balpha$ estimation values for the proposed SR
  and the LASSO.}
  \label{fig_results_para_regression}
  \end{center}
\end{figure}

\textbf{(i) Regression Problems: } Fig.~\ref{fig_results_para_regression} shows some
representative examples of the learned parameters for the proposed SR and the LASSO.
These reported results had used \eqref{eqn_k_norm_solution} for learning in order to
observe the extreme case without regularization. The parameters are sorted in ascending
order for visualization purpose. For the Prostate
(Fig.~\ref{fig_results_para_regression}(a)-(b)) and the Diabetes
(Fig.~\ref{fig_results_para_regression}(c)-(d)) data sets, a sparse estimation for SR is
not observed due to the over-determined problem nature. For the Corn
(Fig.~\ref{fig_results_para_regression}(e)-(f)), TFIDF
(Fig.~\ref{fig_results_para_regression}(g)-(h)) and Gaze
(Fig.~\ref{fig_results_para_regression}(i)-(j)) data sets, the estimated parameters are
seen to have values with relatively large difference. Such difference in the estimated
parameter values has been utilized for sparse parameter selection for the proposed SR. It
is worth noting that for the TFIDF data set, a large proportion of the estimated
parameter values are either zero or almost zero.

\begin{figure}[hhhh]
  \begin{center}
\begin{tabular}{cc}
  \epsfxsize=8cm
  \epsfysize=3.9cm
  \hspace{-2mm}\epsffile[55    48   691   443]{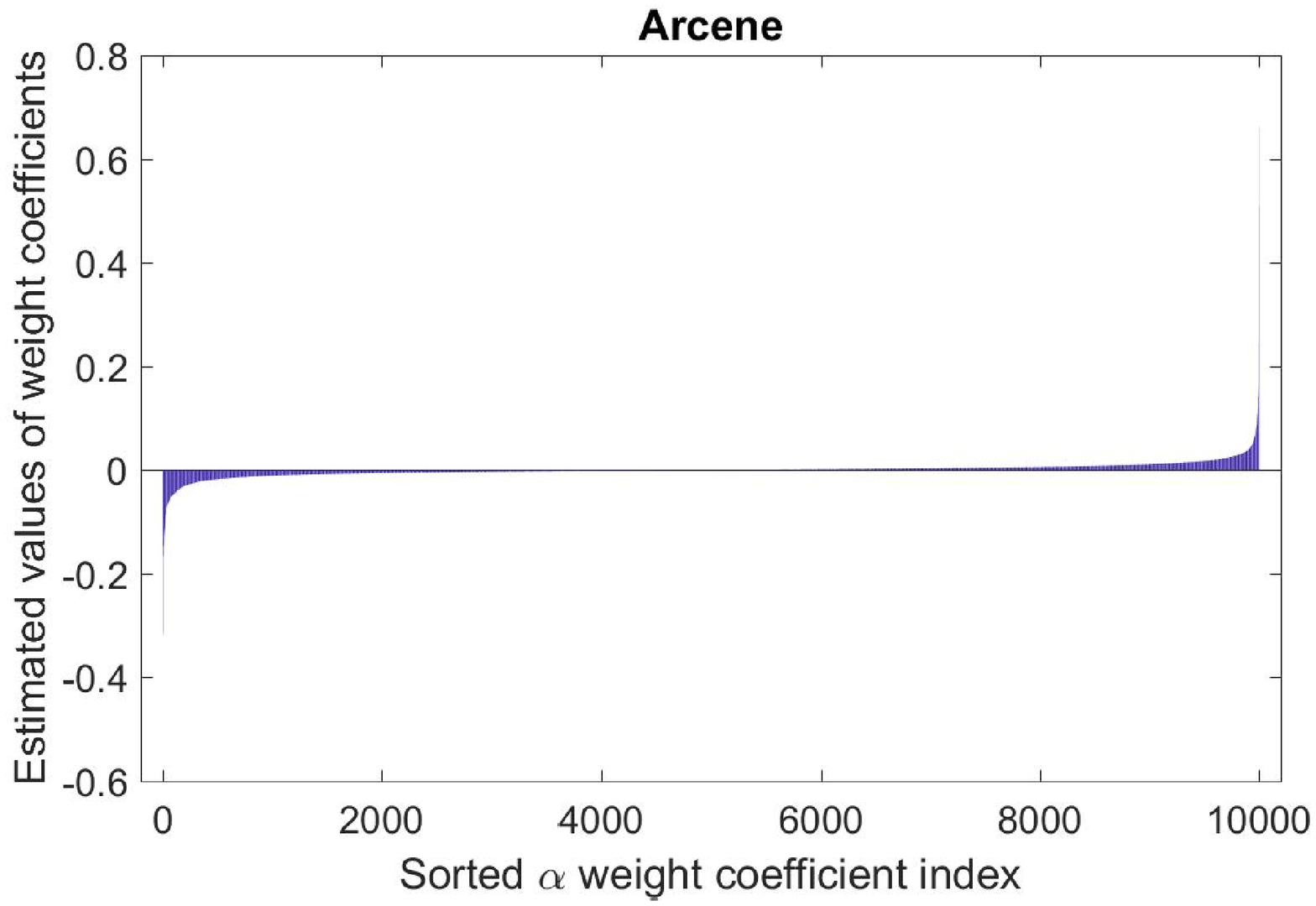} &
  \epsfxsize=8cm
  \epsfysize=3.9cm
  \hspace{0mm}\epsffile[75    53   519   354]{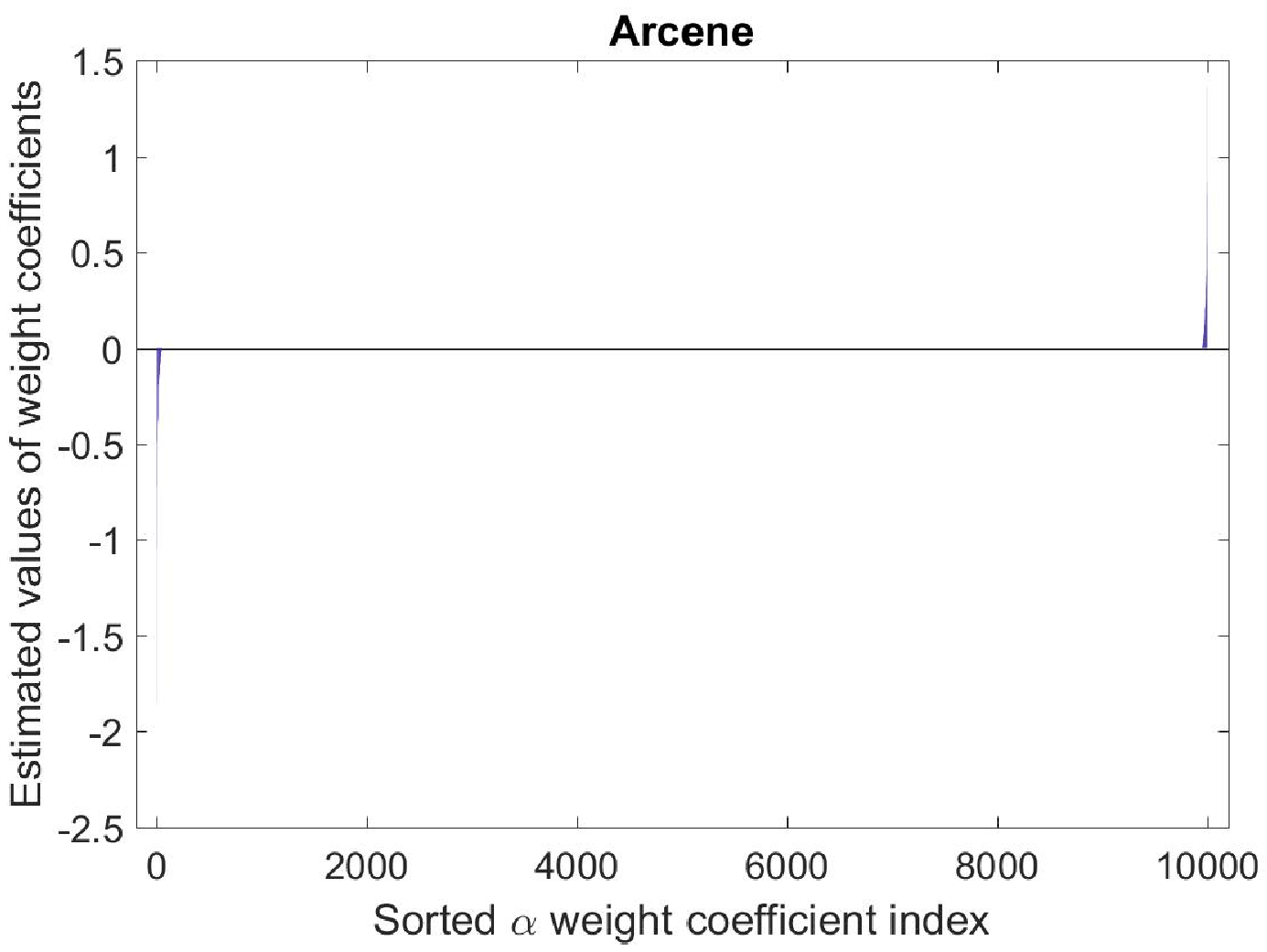}
  \\  \small{Arcene: (a) SR}  & \small{(b) LASSO} \\*[2mm]
  \epsfxsize=8cm
  \epsfysize=3.9cm
  \hspace{-2mm}\epsffile[58    48   683   457]{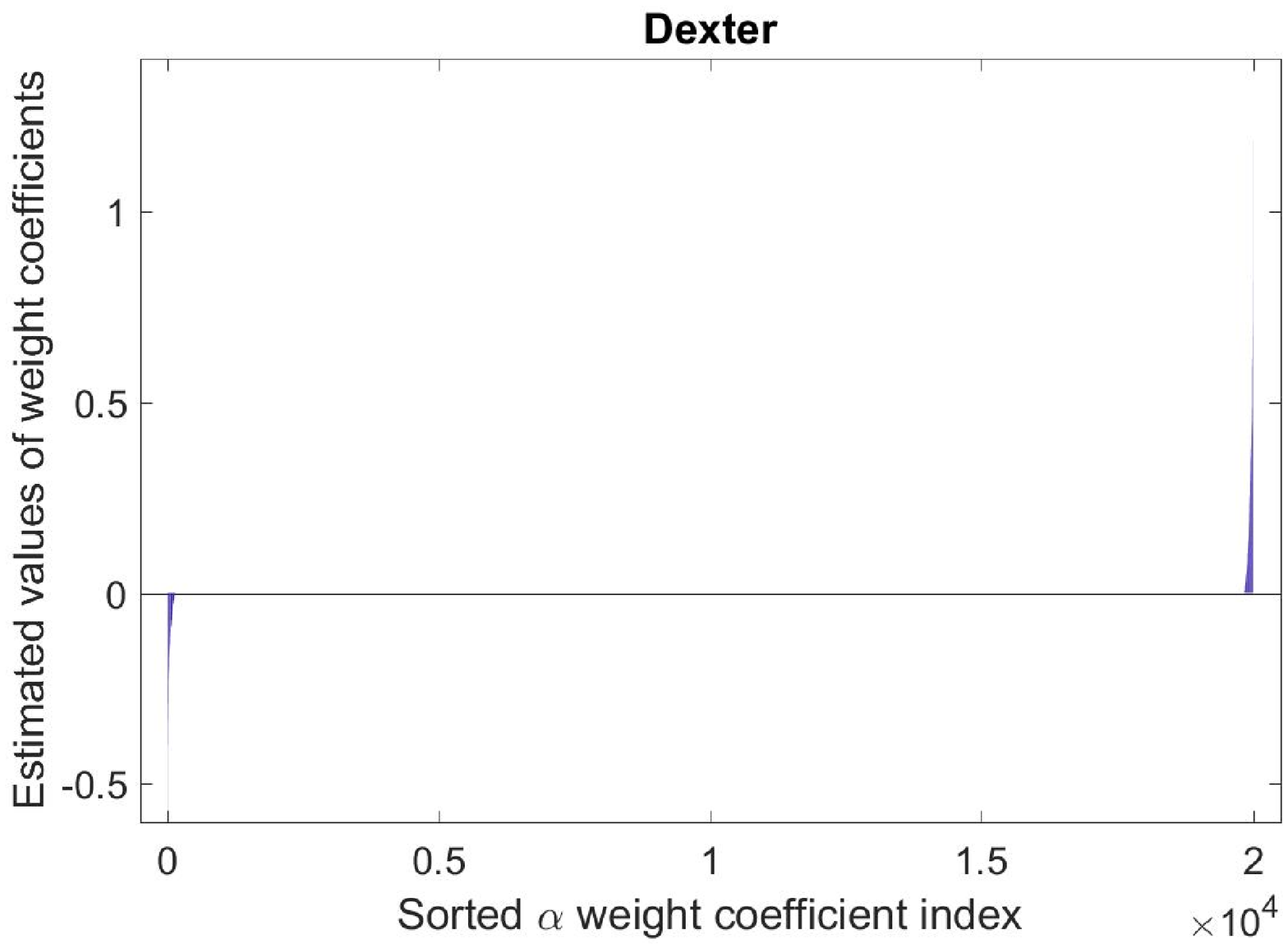} &
  \epsfxsize=8cm
  \epsfysize=3.9cm
  \hspace{0mm}\epsffile[60    61   576   383]{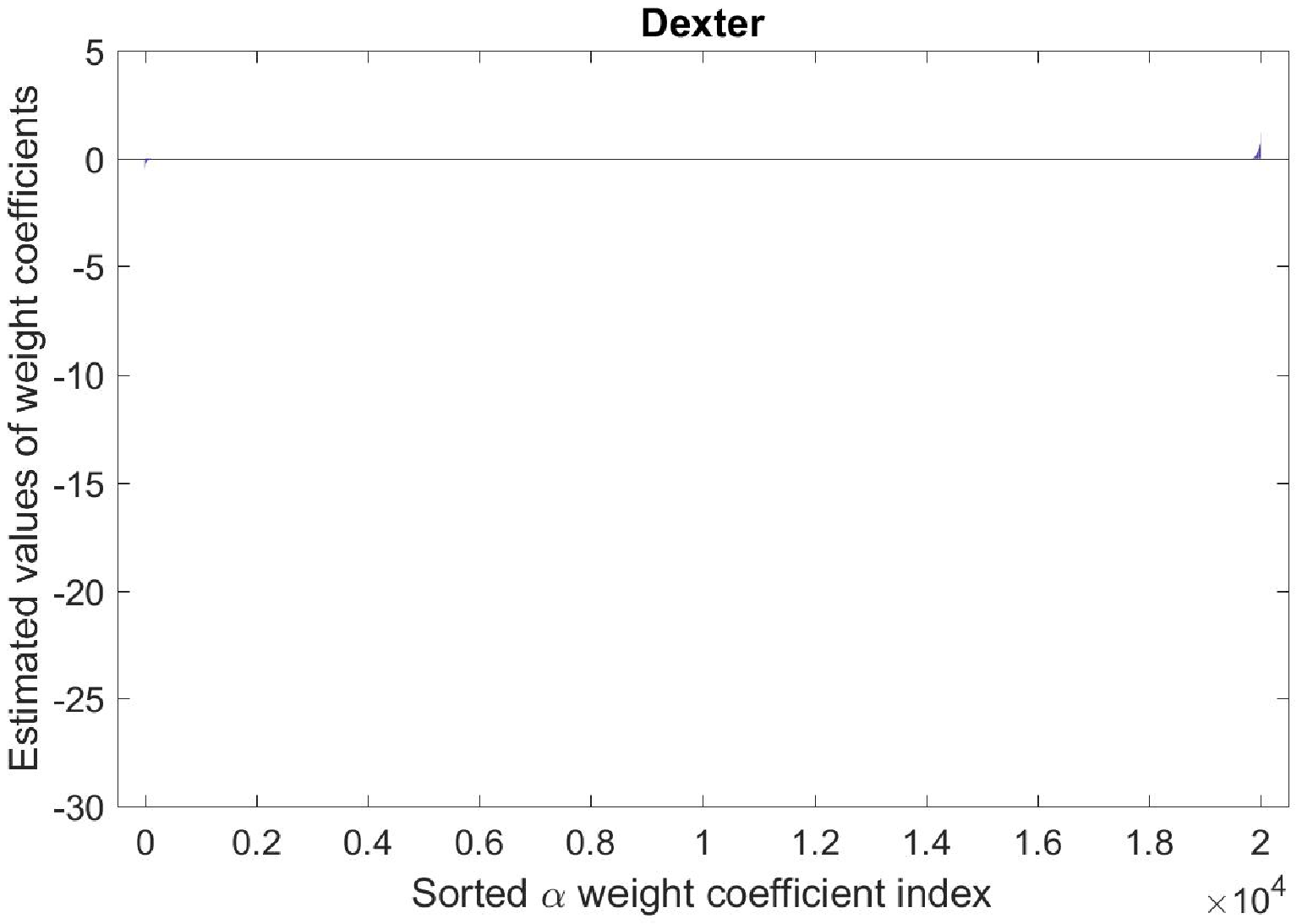}
  \\  \small{Dexter: (c) SR}  & \small{(d) LASSO} \\*[2mm]
  \epsfxsize=8cm
  \epsfysize=3.9cm
  \hspace{-2mm}\epsffile[55    49   681   464]{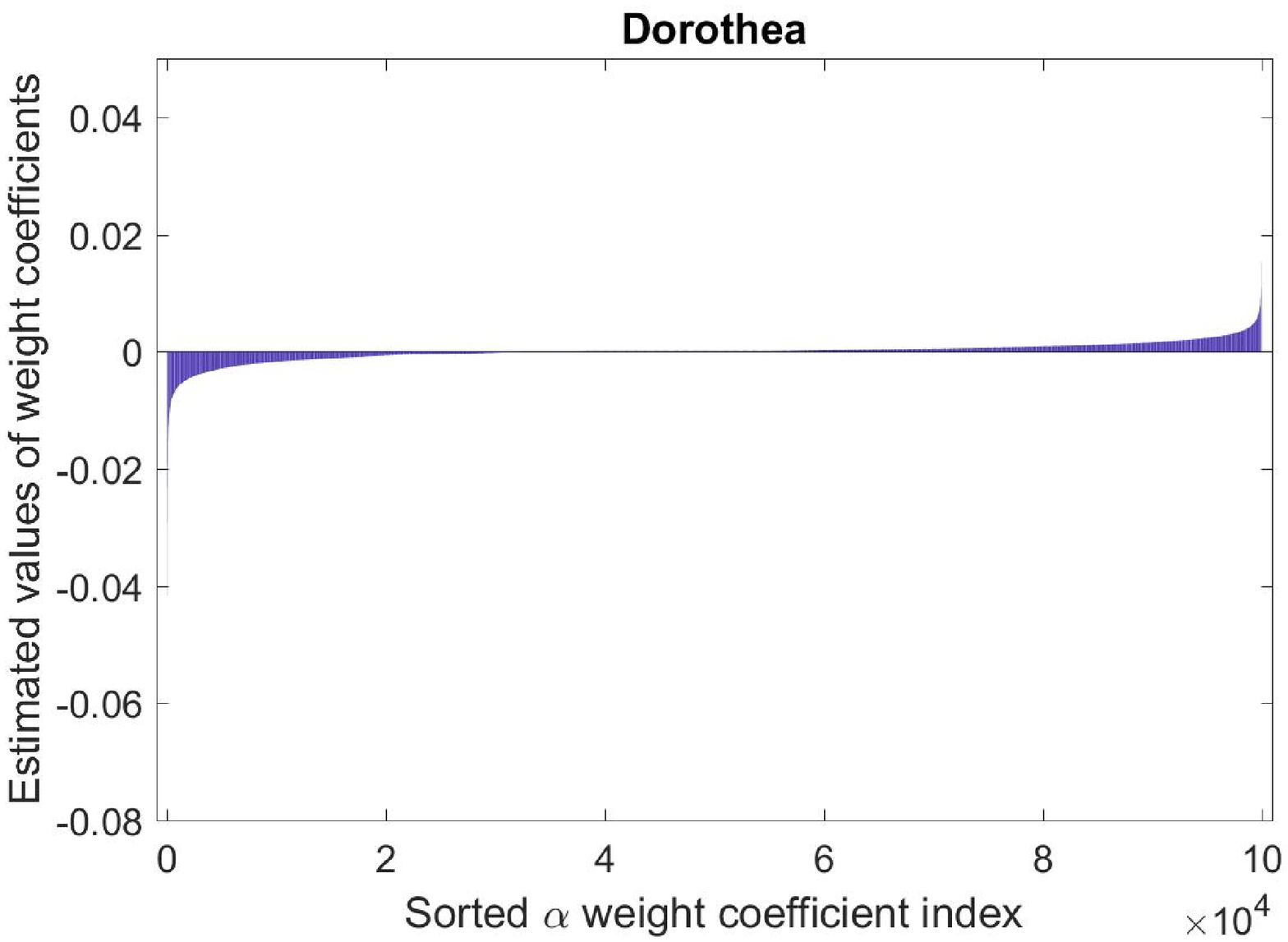} &
  \epsfxsize=8cm
  \epsfysize=3.9cm
  \hspace{0mm}\epsffile[90    55   623   381]{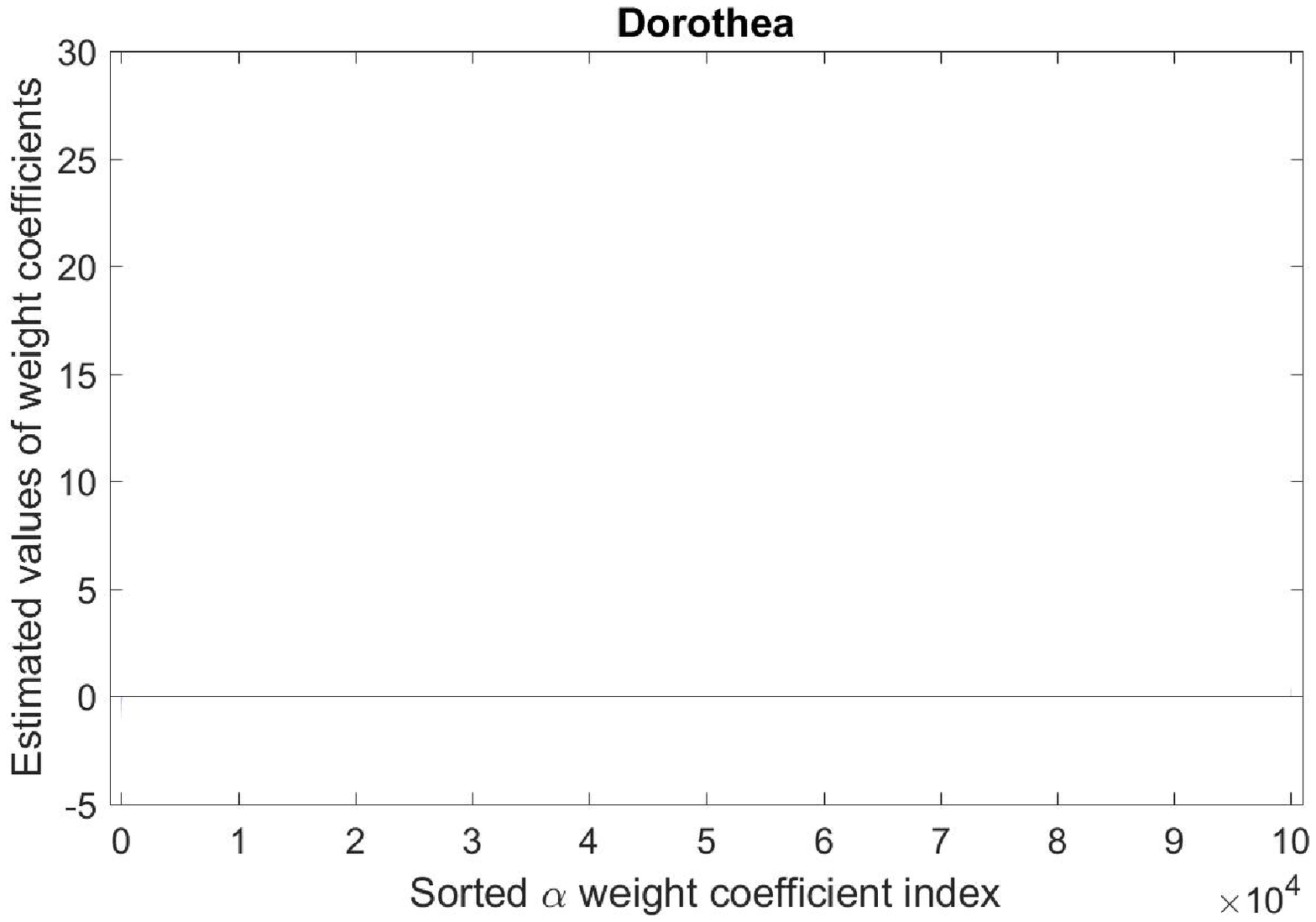}
  \\  \small{Dorothea: (e) SR}  & \small{(f) LASSO} \\*[2mm]
  \epsfxsize=8cm
  \epsfysize=3.9cm
  \hspace{-2mm}\epsffile[55    48   603   428]{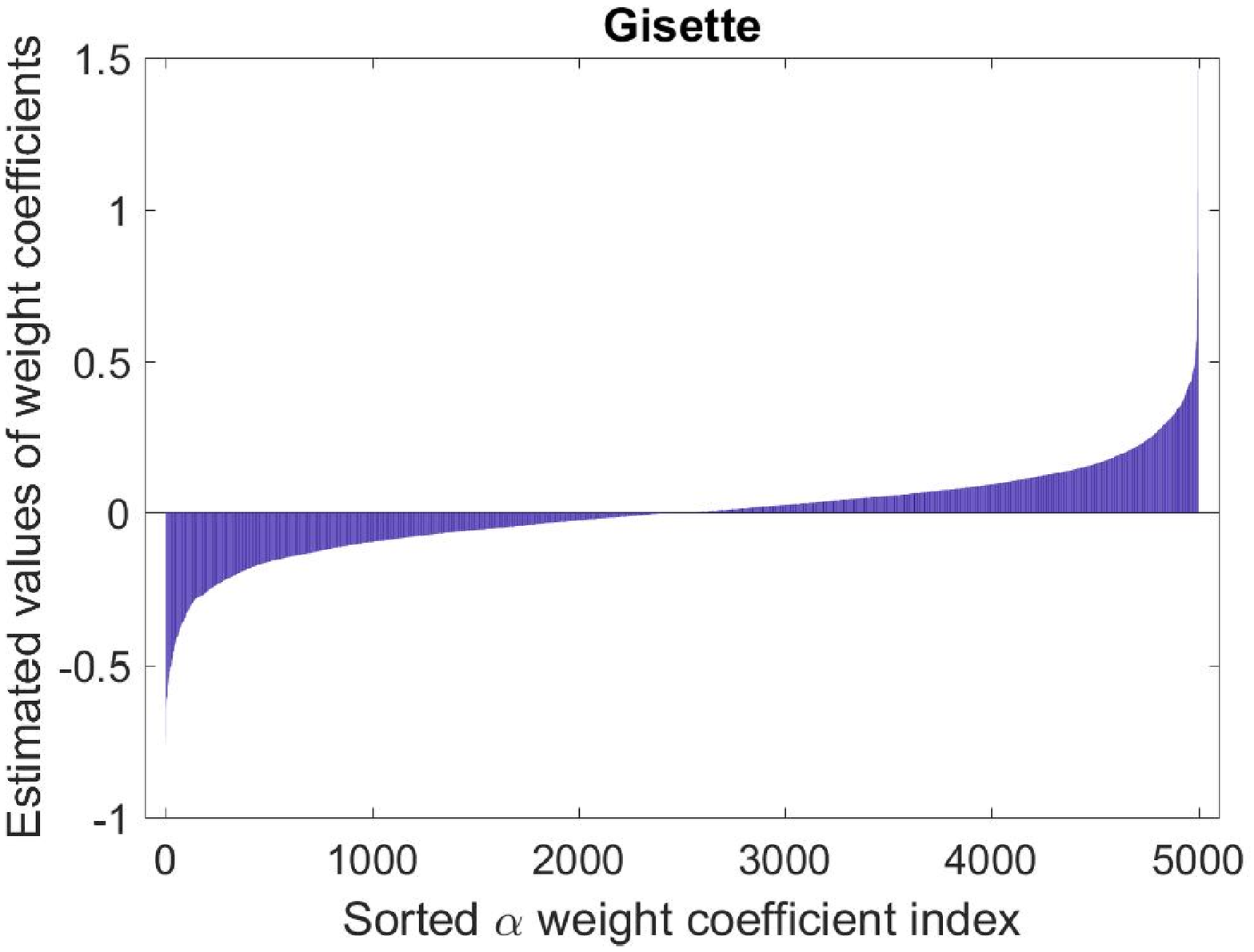} &
  \epsfxsize=8cm
  \epsfysize=3.9cm
  \hspace{0mm}\epsffile[55    48   662   449]{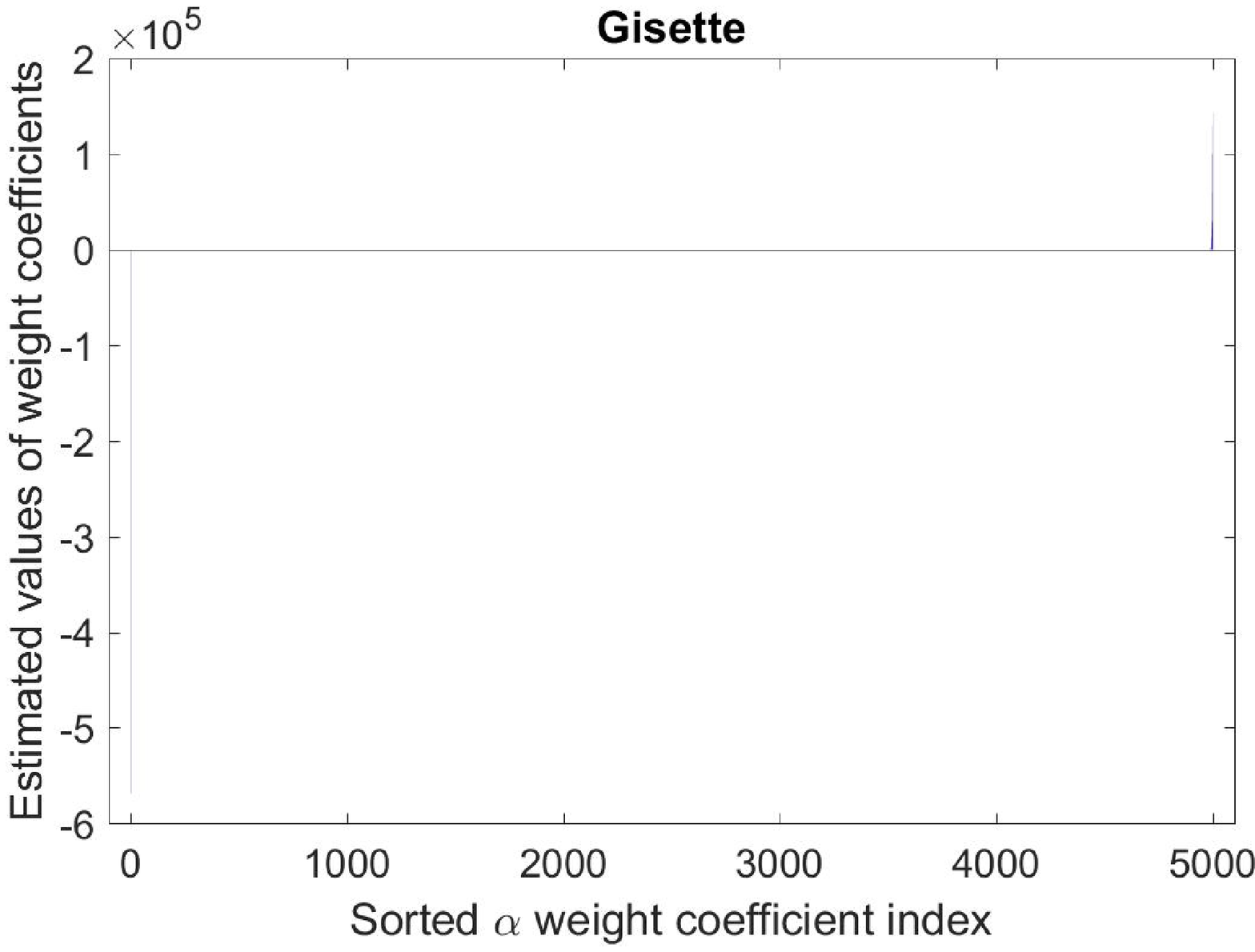}
  \\  \small{Gisette: (g) SR}  & (h) \small{LASSO} \\*[2mm]
  \epsfxsize=8cm
  \epsfysize=3.9cm
  \hspace{-2mm}\epsffile[55    49   639   452]{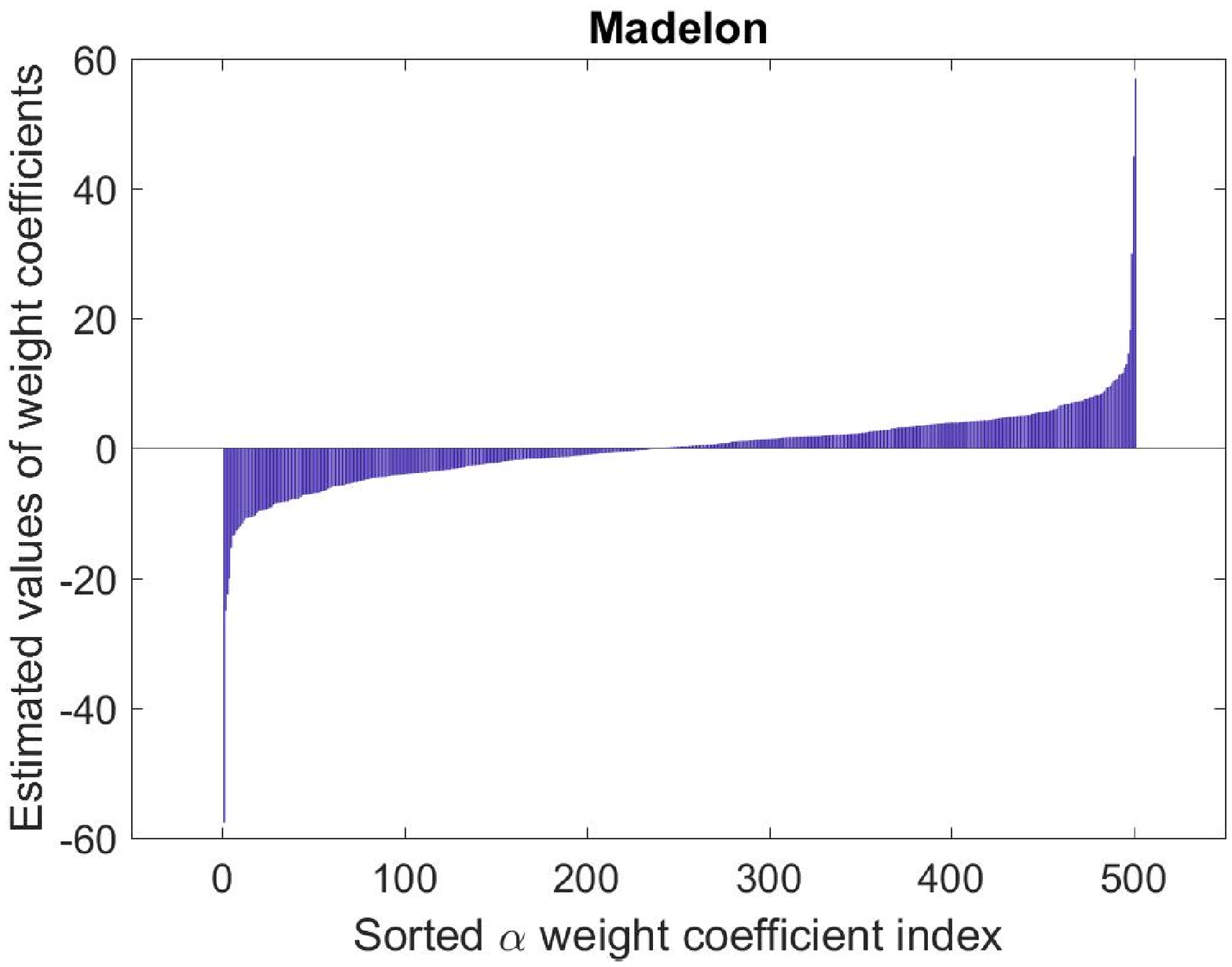} &
  \epsfxsize=8cm
  \epsfysize=3.9cm
  \hspace{0mm}\epsffile[55    49   639   457]{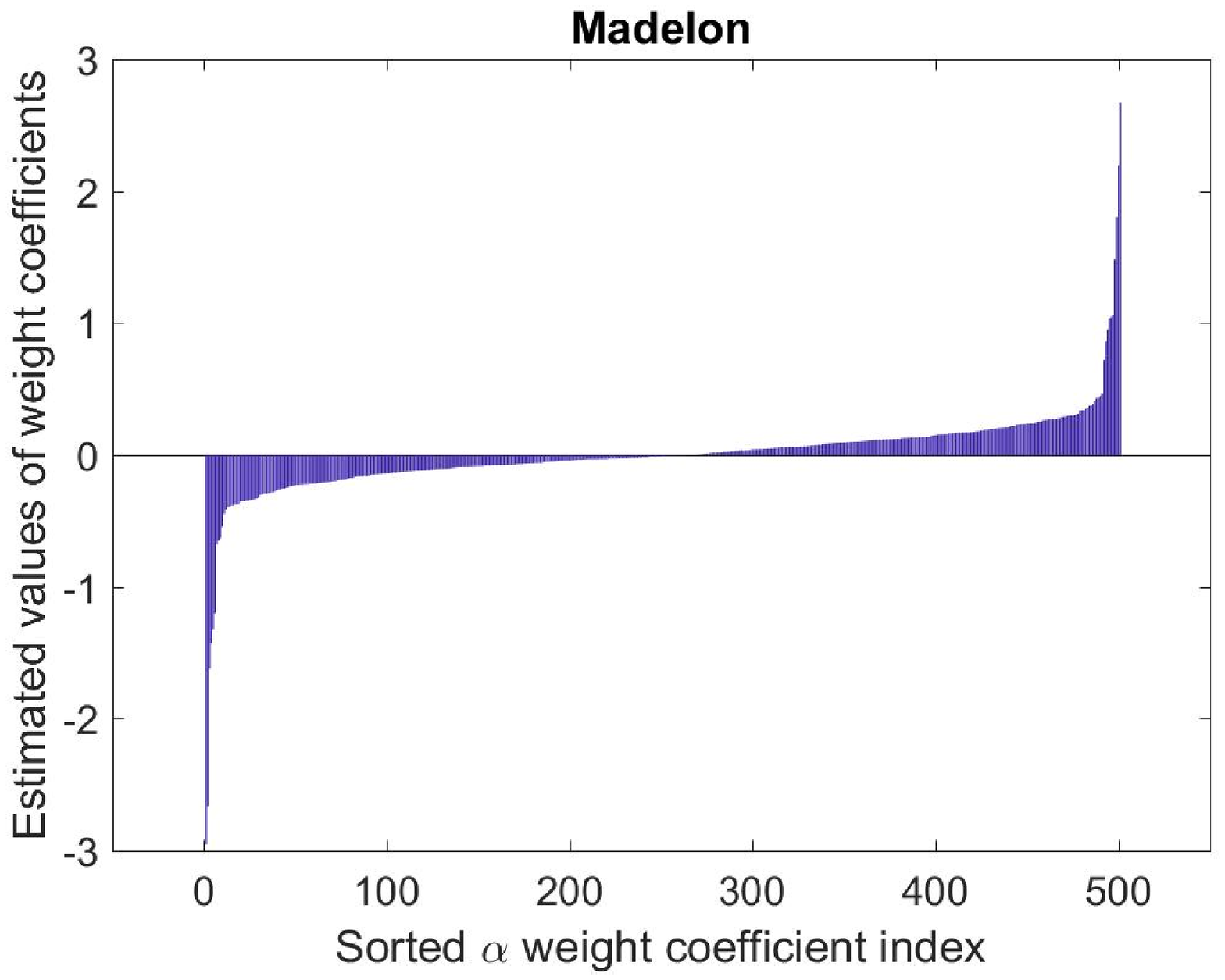}
  \\  \small{Madelon: (i) SR}  & \small{(j) LASSO}
\end{tabular}
  \caption{Classification: sorted parameter $\bm{\alpha}$ estimation values for proposed
  SR at $k=1.1$ and LASSO at $\texttt{Alpha}=1$.}
  \label{fig_results_para_classification}
  \end{center}
\end{figure}

\textbf{(ii) Classification Problems: } Fig.~\ref{fig_results_para_classification} shows
the exemplar estimated parameters which are sorted in ascending order for each of the
classification data sets using \eqref{eqn_k_norm_solution}. In most cases, the parameters
are observed to have relatively large difference in terms of their estimation values. The
sparse distribution of the parameters is particularly discernible for the Arcene, Dexter
and Dorothea data sets (Fig.~\ref{fig_results_para_classification}(a), (c), (e)). For the
Gisette data set (Fig.~\ref{fig_results_para_classification}(g) and (h)), the proposed SR
show a relatively large difference in the magnitudes of parameters between $\alpha_i<0.5$
and $\alpha_i>0.5$, $i\in\{0,1,...,5000\}$. For the Madelon data set,
Fig.~\ref{fig_results_para_classification}(i) and (j) show a similar parameter
distribution trend for both LASSO and SR except for their estimation value ranges. Due to
the over-determined setting, the parameters do not show sparse distribution here.

\subsection{Results: Prediction Accuracy}

\textbf{(A) Effect of ${k}$ values}: In this part of the experiment, the impact of
different ${k}$ values on the estimation accuracy is observed.

\begin{figure}[hhhh]
  \begin{center}
\begin{tabular}{cc}
  \epsfxsize=6.8cm
  \epsfysize=3.9cm
  \hspace{-2mm}\epsffile[15     0   383   295]{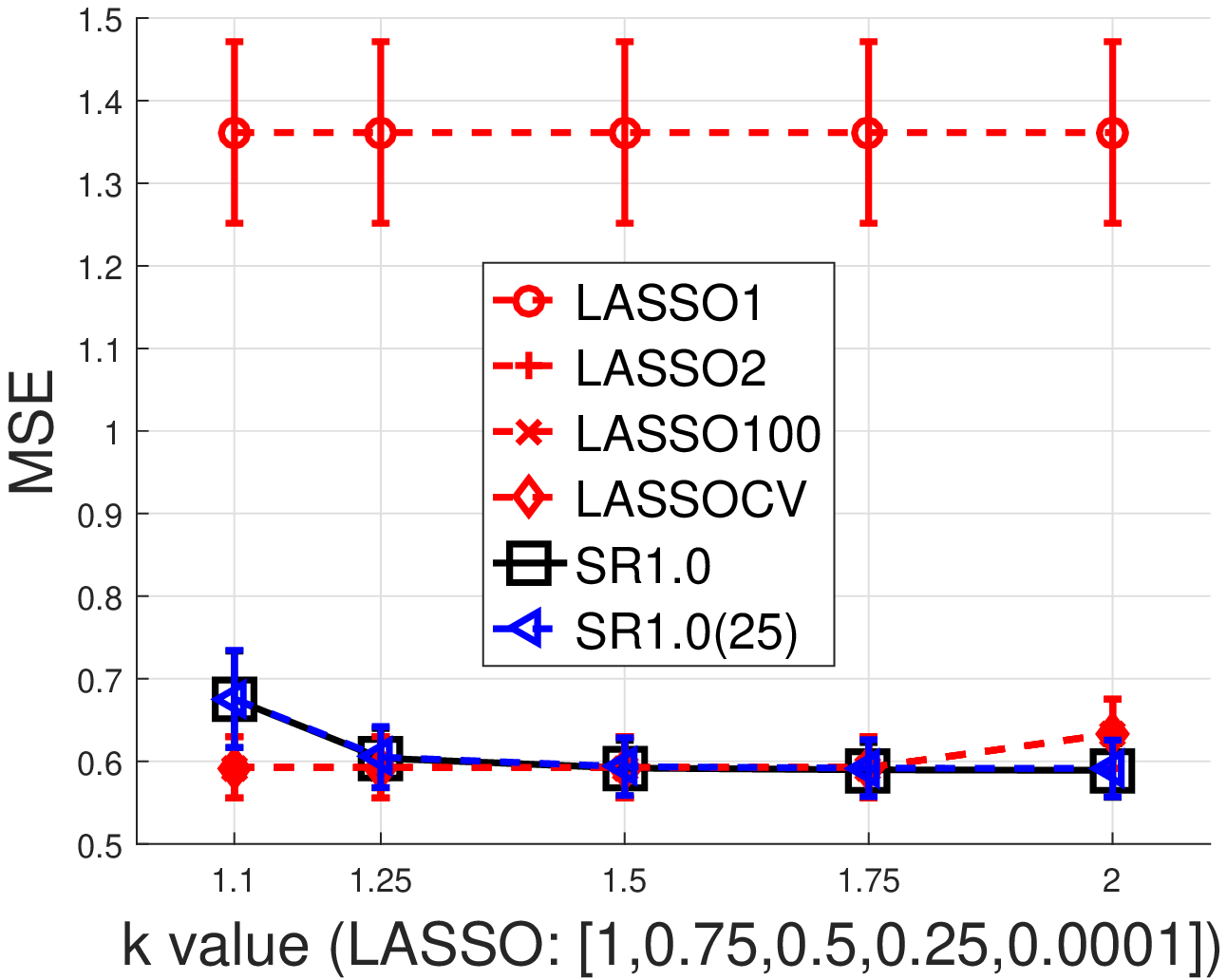} &
  \epsfxsize=6.8cm
  \epsfysize=3.9cm
  \hspace{-2mm}\epsffile[11     0   383   297]{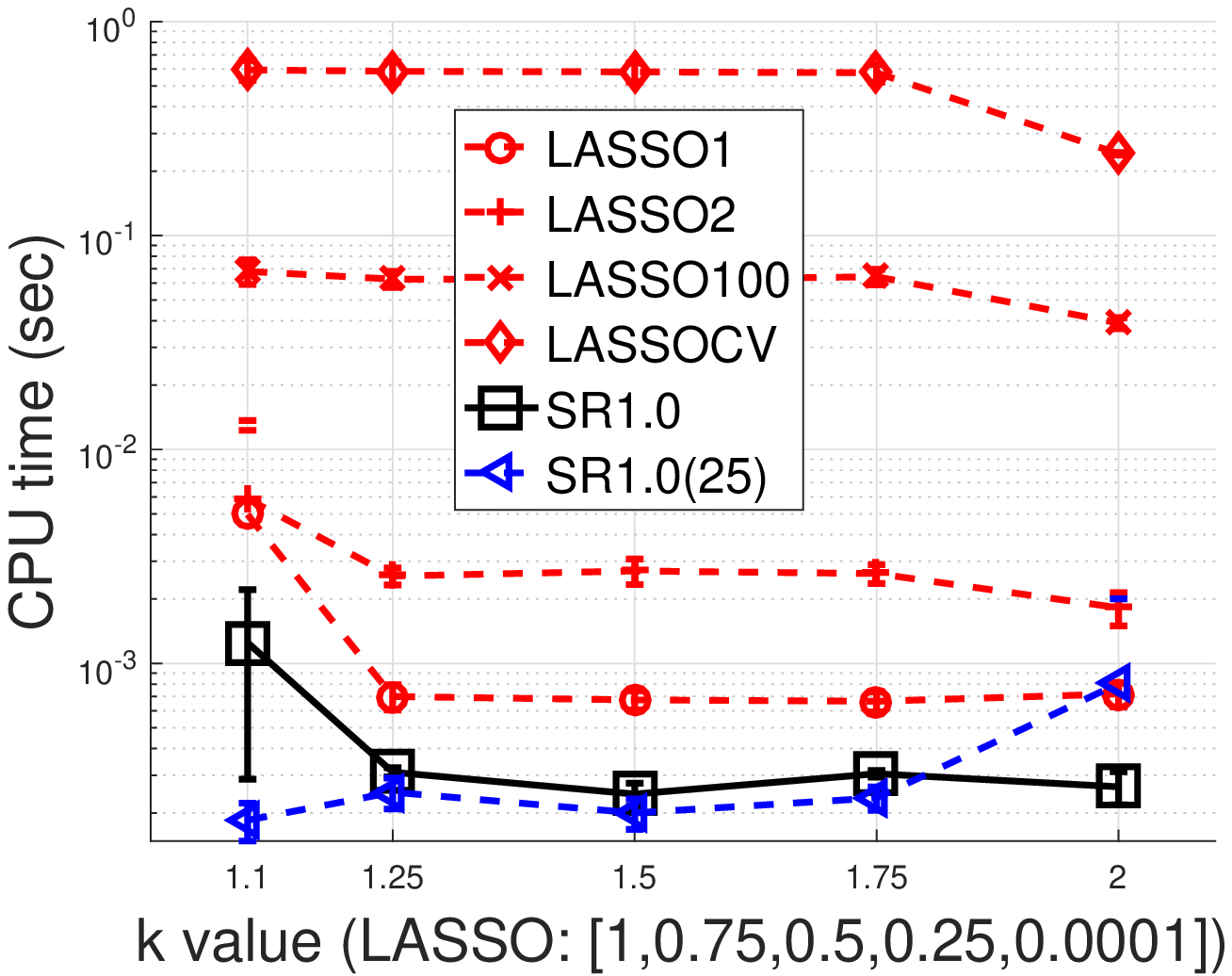}
  \\  \small{Prostate:  (a) MSE}  & \small{(b) CPU} \\*[2mm]
  \epsfxsize=6.8cm
  \epsfysize=3.9cm
  \hspace{-2mm}\epsffile[15     0   383   295]{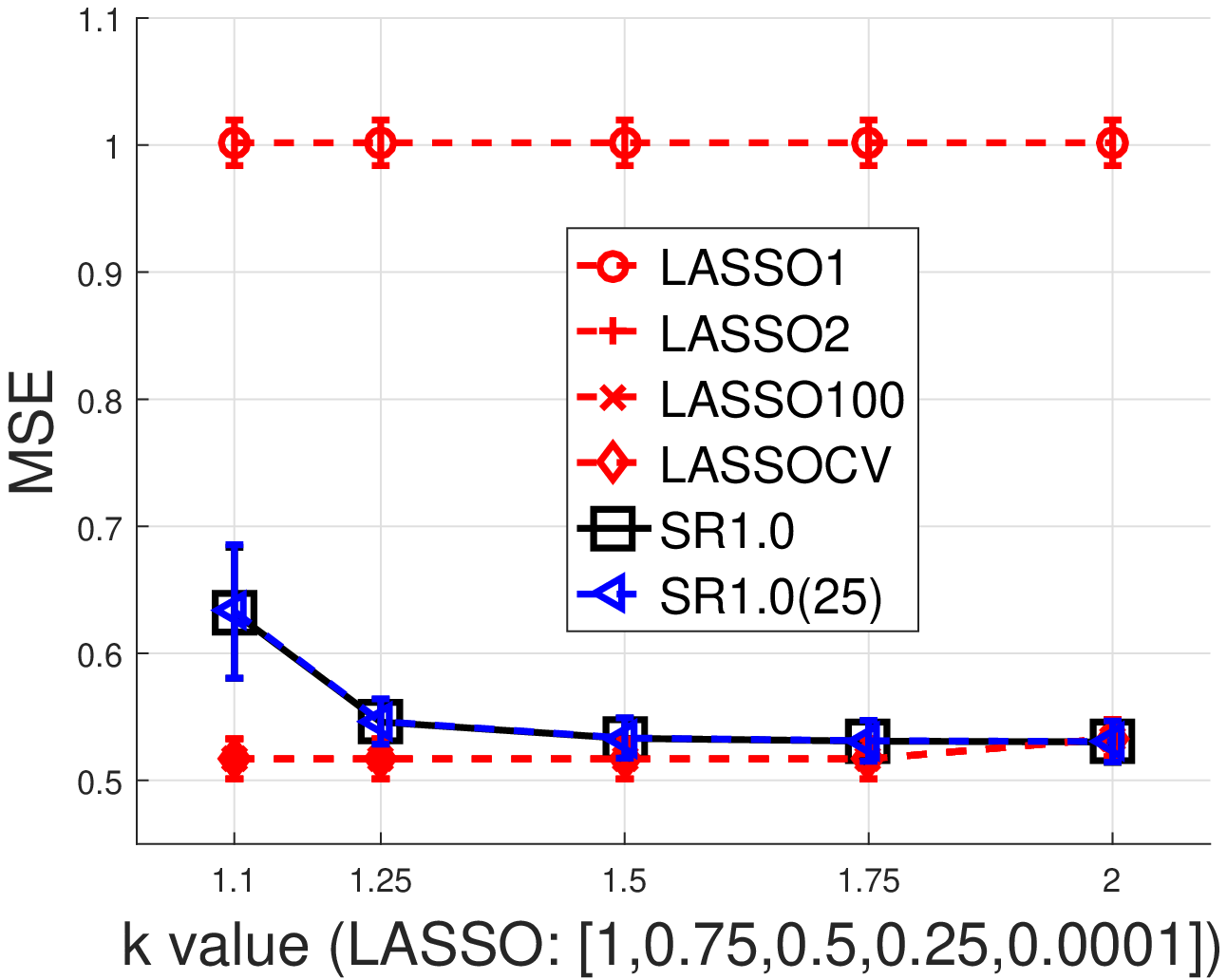} &
  \epsfxsize=6.8cm
  \epsfysize=3.9cm
  \hspace{-2mm}\epsffile[11     0   383   297]{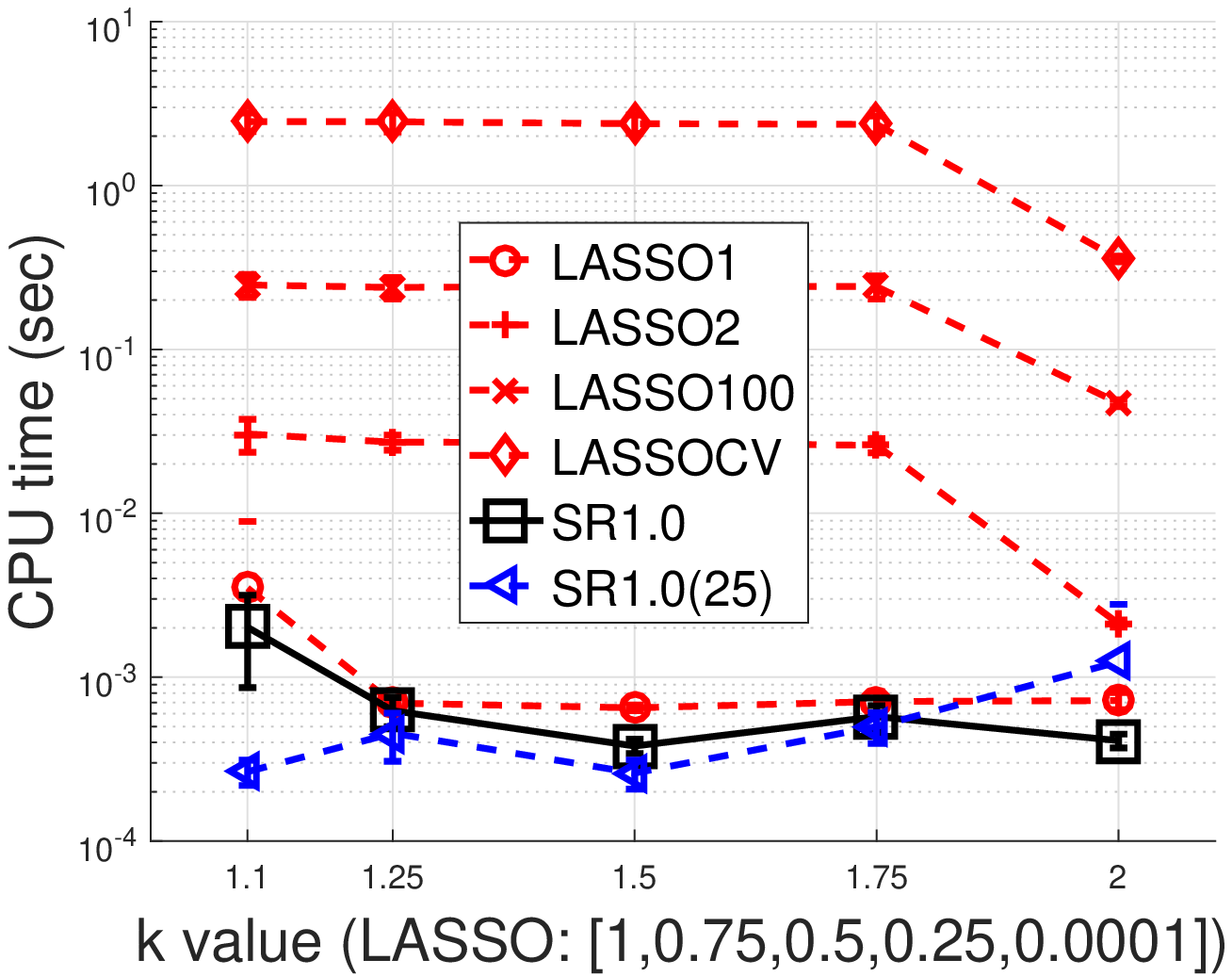}
  \\  \small{Diabetes: (c) MSE}  & \small{(d) CPU} \\*[2mm]
  \epsfxsize=6.8cm
  \epsfysize=3.9cm
  \hspace{-2mm}\epsffile[11     0   383   296]{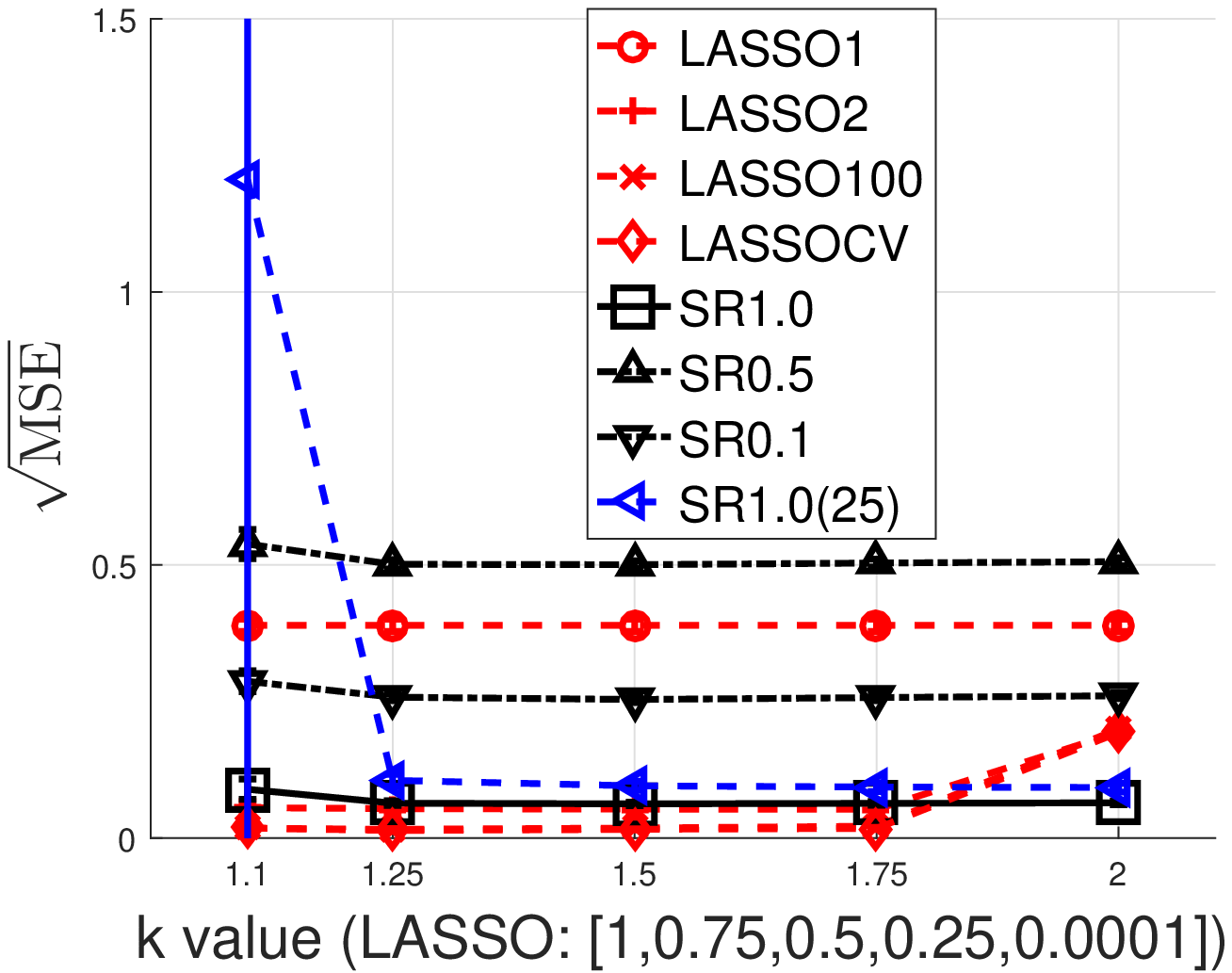} &
  \epsfxsize=6.8cm
  \epsfysize=3.9cm
  \hspace{-2mm}\epsffile[11     0   383   297]{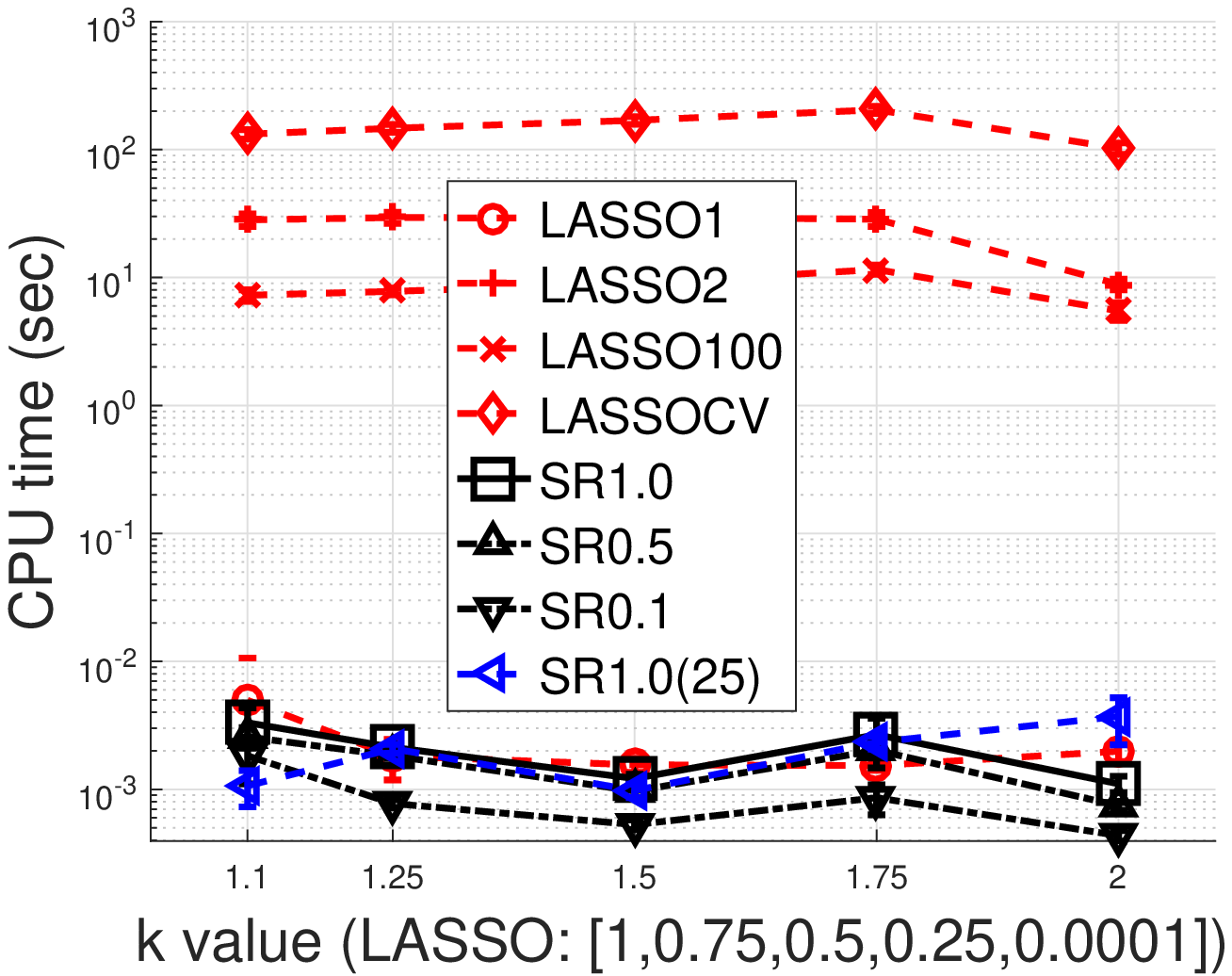}
  \\  \small{Corn: (e) $\sqrt{\rm MSE}$}  & \small{(f) CPU} \\*[2mm]
  \epsfxsize=6.8cm
  \epsfysize=3.9cm
  \hspace{-2mm}\epsffile[33     0   444   349]{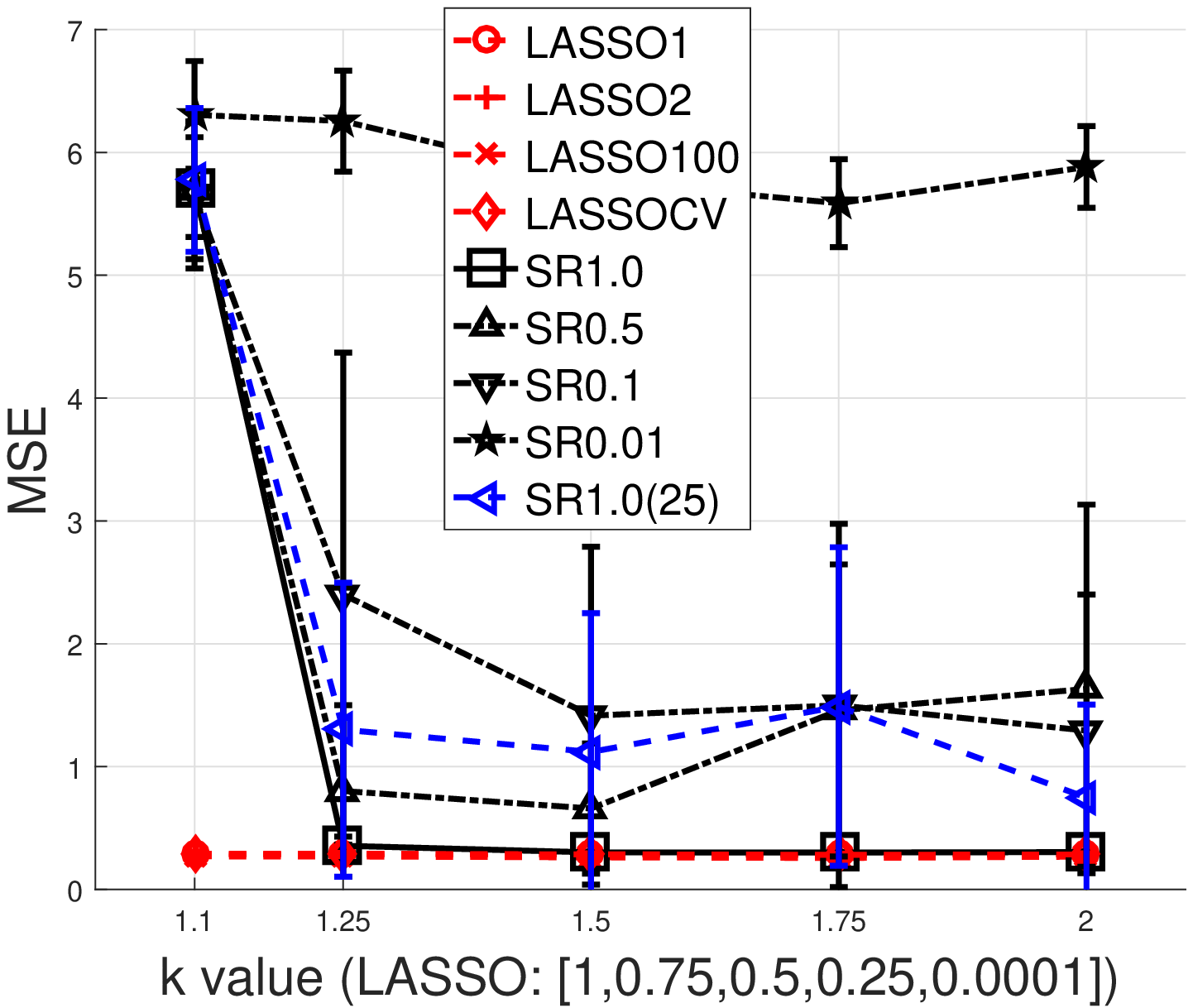} &
  \epsfxsize=6.8cm
  \epsfysize=3.9cm
  \hspace{-2mm}\epsffile[20     0   445   343]{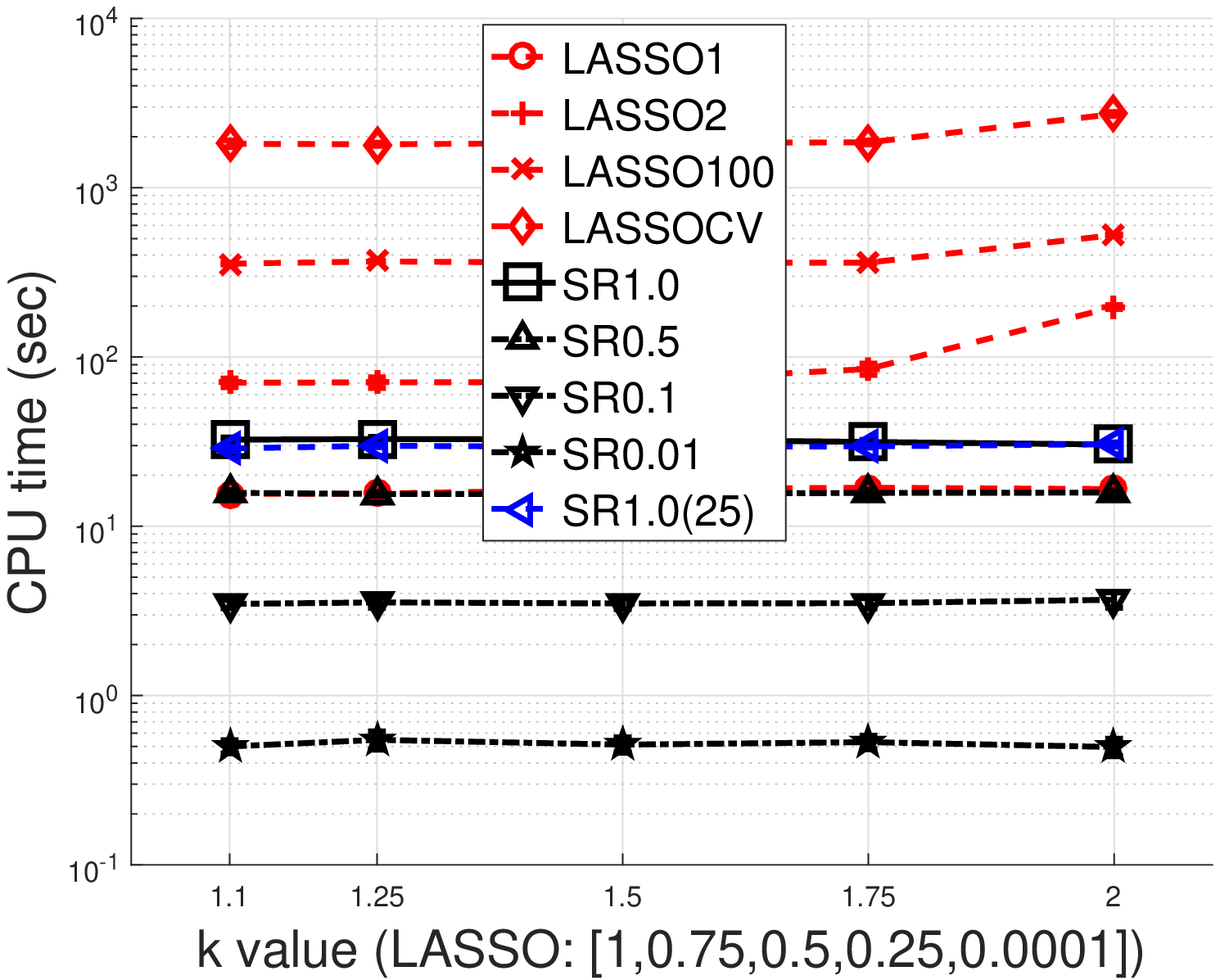}
  \\  \small{TFIDF: (g) MSE}  & \small{(h) CPU} \\*[2mm]
  \epsfxsize=6.8cm
  \epsfysize=3.9cm
  \hspace{-2mm}\epsffile[19     0   445   341]{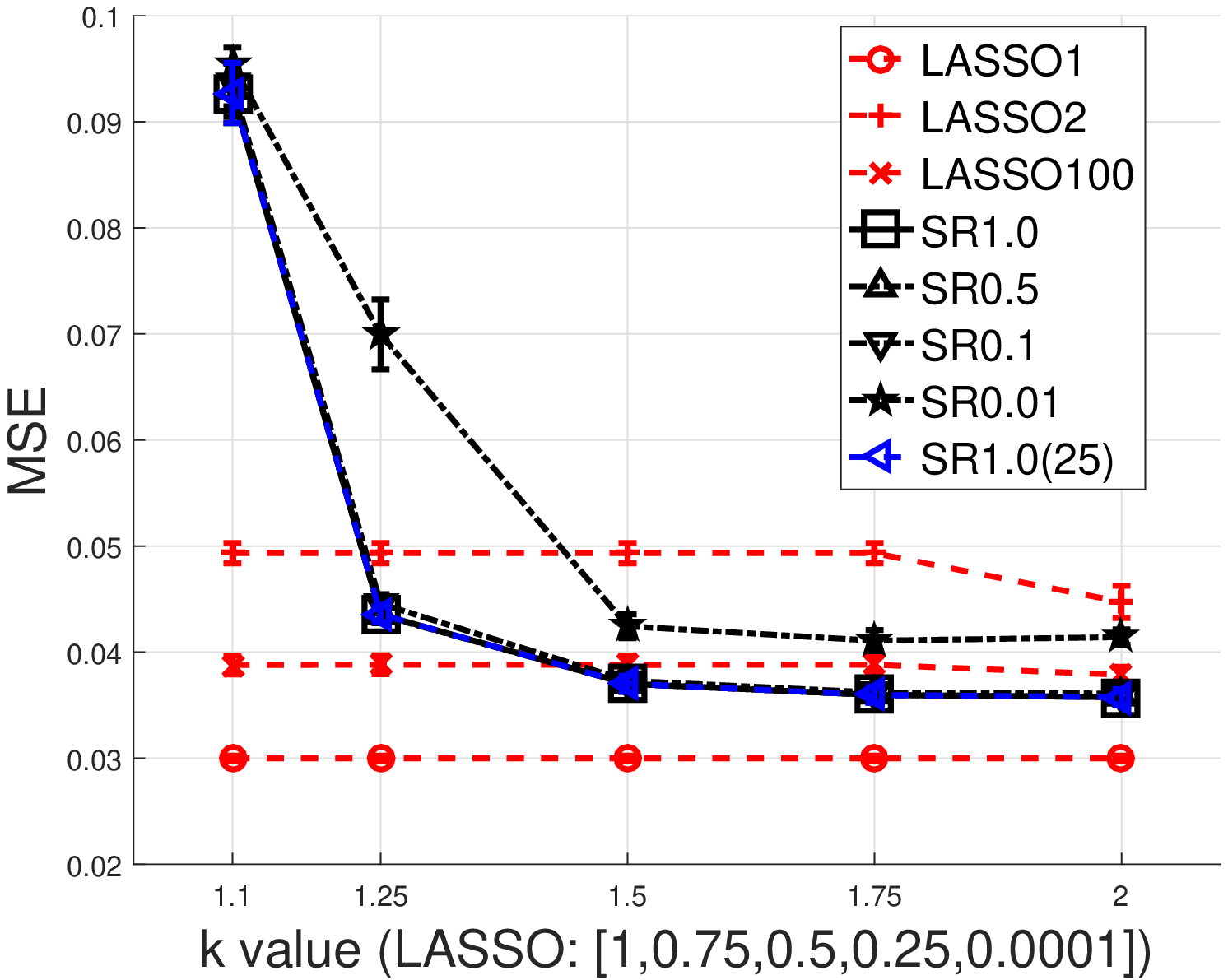} &
  \epsfxsize=6.8cm
  \epsfysize=3.9cm
  \hspace{-2mm}\epsffile[22     0   444   342]{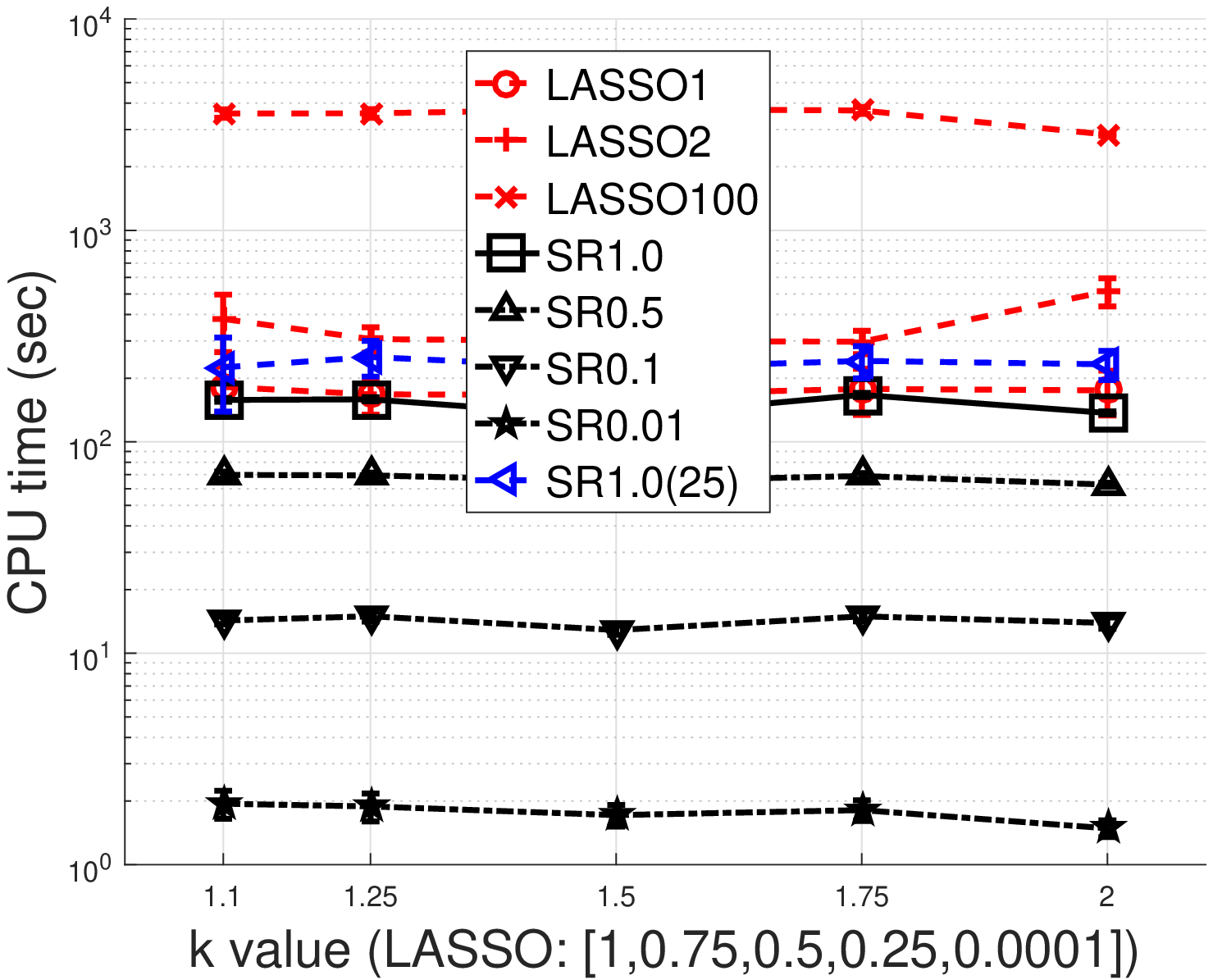}
  \\  \small{Gaze: (i) MSE}  & \small{(j) CPU} \\*[2mm]
\end{tabular}
  \caption{Regression: MSE and CPU plots for SR at $k\in\{1.1,1.25,1.5,1.75,2\}$ and LASSO
  at \texttt{Alpha}$\in\{1,0.75,0.5,0.25,0.0001\}$.}
  \label{fig_results_Acc_regression}
  \end{center}
\end{figure}

\textbf{(i) Regression Problems: }
Fig.~\ref{fig_results_Acc_regression}(a),(c),(e),(g),(i) show the accuracy results in
terms of MSE for the SR and the LASSO on the regression data sets. Due to the relatively
similar results at different regularization settings, only the results at ${c}=10^2$ are
shown in these figures to avoid cluttering of plots, apart from the result for the
non-regularized case using \eqref{eqn_k_norm_solution} (denoted as SR1.0(25)). For the
Prostate and Diabetes data sets as shown in Fig.~\ref{fig_results_Acc_regression}(a) and
(c), a comparable MSE result is observed for both the SR and the LASSO except at LASSO1
setting. For the Corn data set in Fig.~\ref{fig_results_Acc_regression}(e), the square
root of MSE is adopted to reduce the many leading zeros of the MSE representation. These
results show comparable performance between SR and LASSO. The MSE results for the TFIDF
data set in Fig.~\ref{fig_results_Acc_regression}(g) show a different behavior comparing
with that of the Corn data set for the SR where the error bars show large standard
deviations for most $k$ values. In contrast, LASSO shows a relatively stable MSE at all
\texttt{Alpha} settings. Finally, the MSE results in terms of radian gaze angle in
Fig.~\ref{fig_results_Acc_regression}(i) show comparable performance of SR with that of
LASSO only at large $k$-values ($k \geq 1.5$).

\begin{figure}[hhh]
  \begin{center}
\begin{tabular}{ccc}
  \epsfxsize=6.8cm
  \epsfysize=3.9cm
  \hspace{-2mm}\epsffile[25     0   446   341]{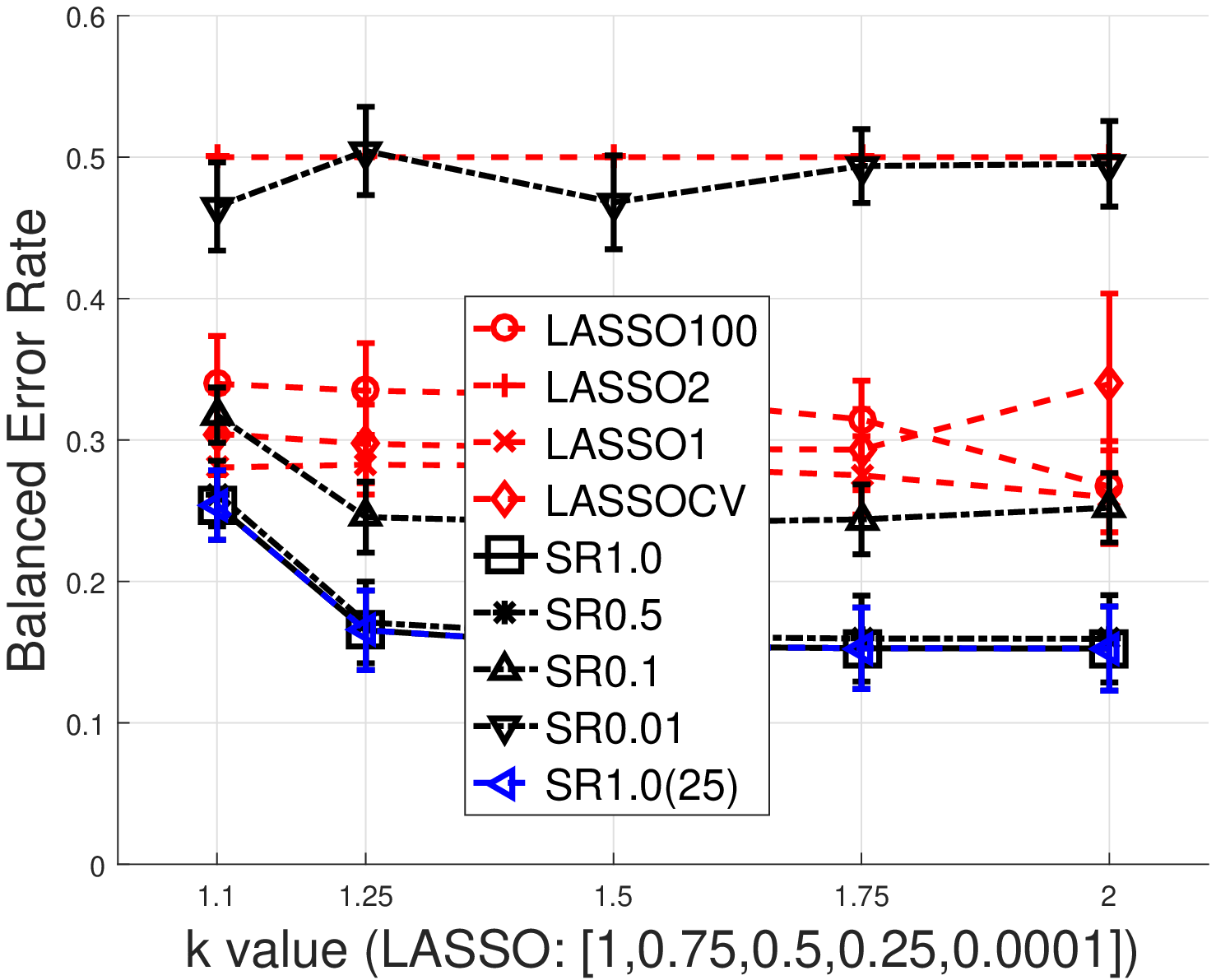} &
  \epsfxsize=6.8cm
  \epsfysize=3.9cm
  \hspace{-2mm}\epsffile[20     0   446   340]{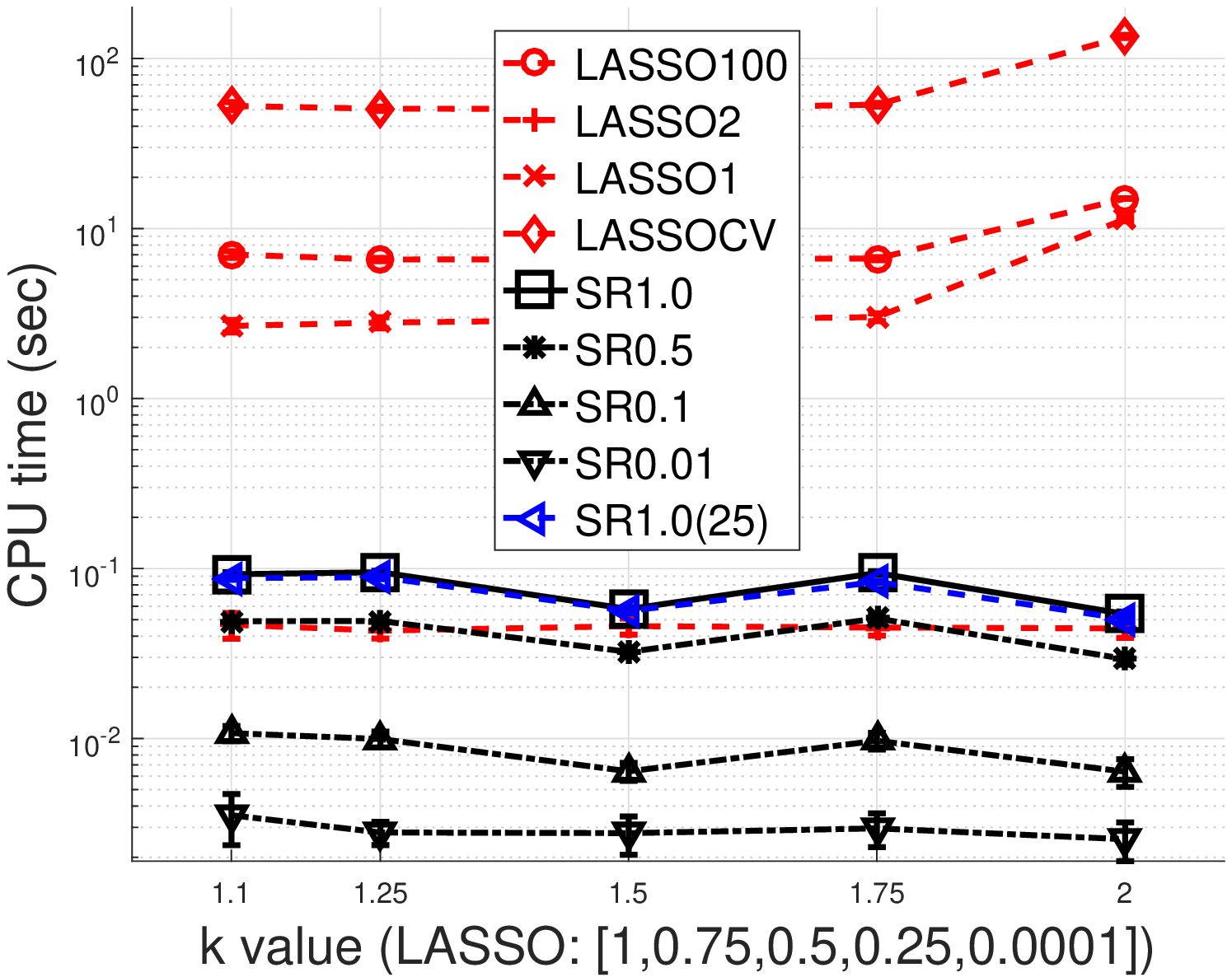}
  \\  \small{Arcene: (a) BER}  & \small{(b) CPU} \\*[2mm]
  \epsfxsize=6.8cm
  \epsfysize=3.9cm
  \hspace{-2mm}\epsffile[25     0   445   340]{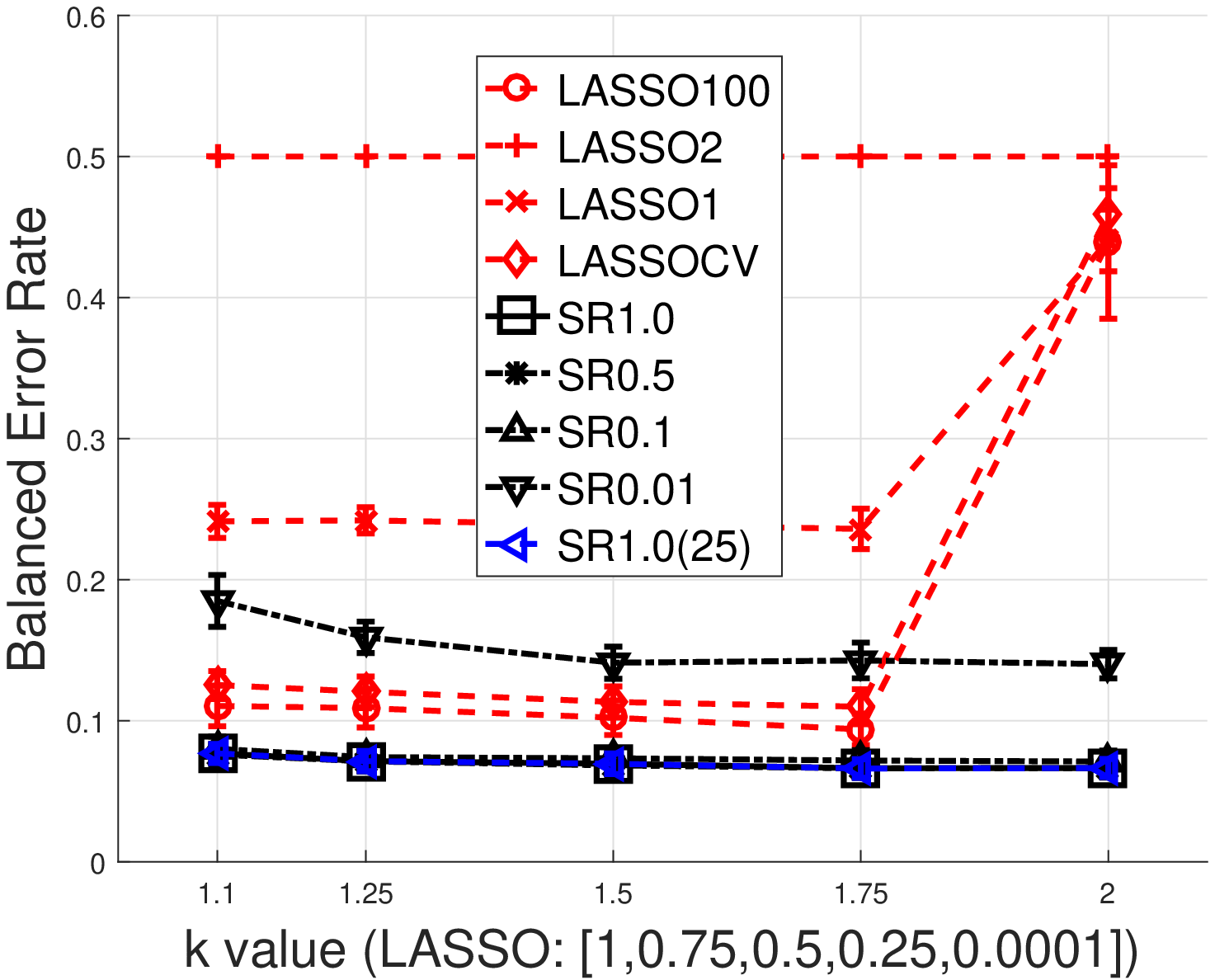} &
  \epsfxsize=6.8cm
  \epsfysize=3.9cm
  \hspace{-2mm}\epsffile[20     0   445   342]{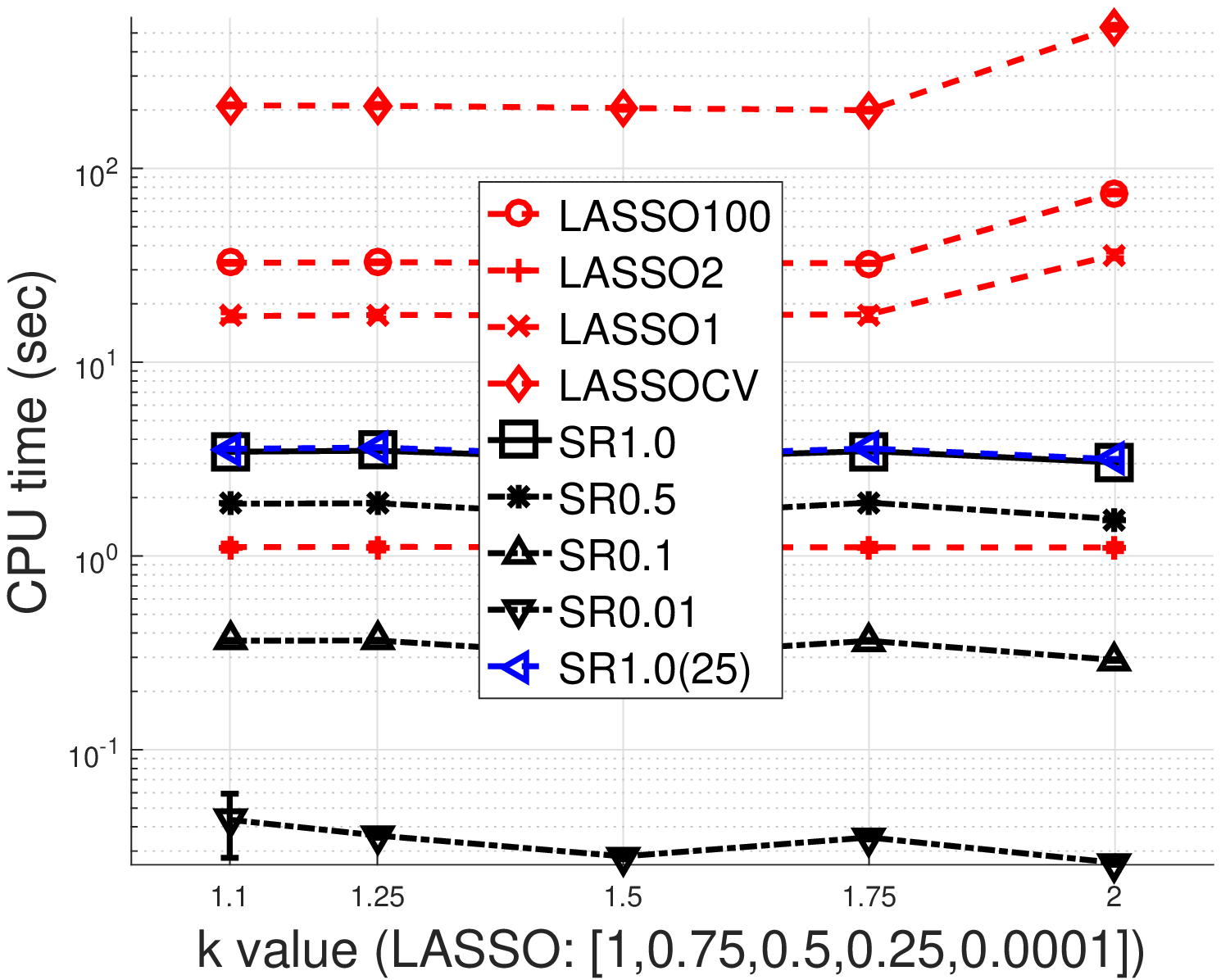}
  \\  \small{Dexter: (c) BER}  & \small{(d) CPU} \\*[2mm]
  \epsfxsize=6.8cm
  \epsfysize=3.9cm
  \hspace{-2mm}\epsffile[19     0   445   342]{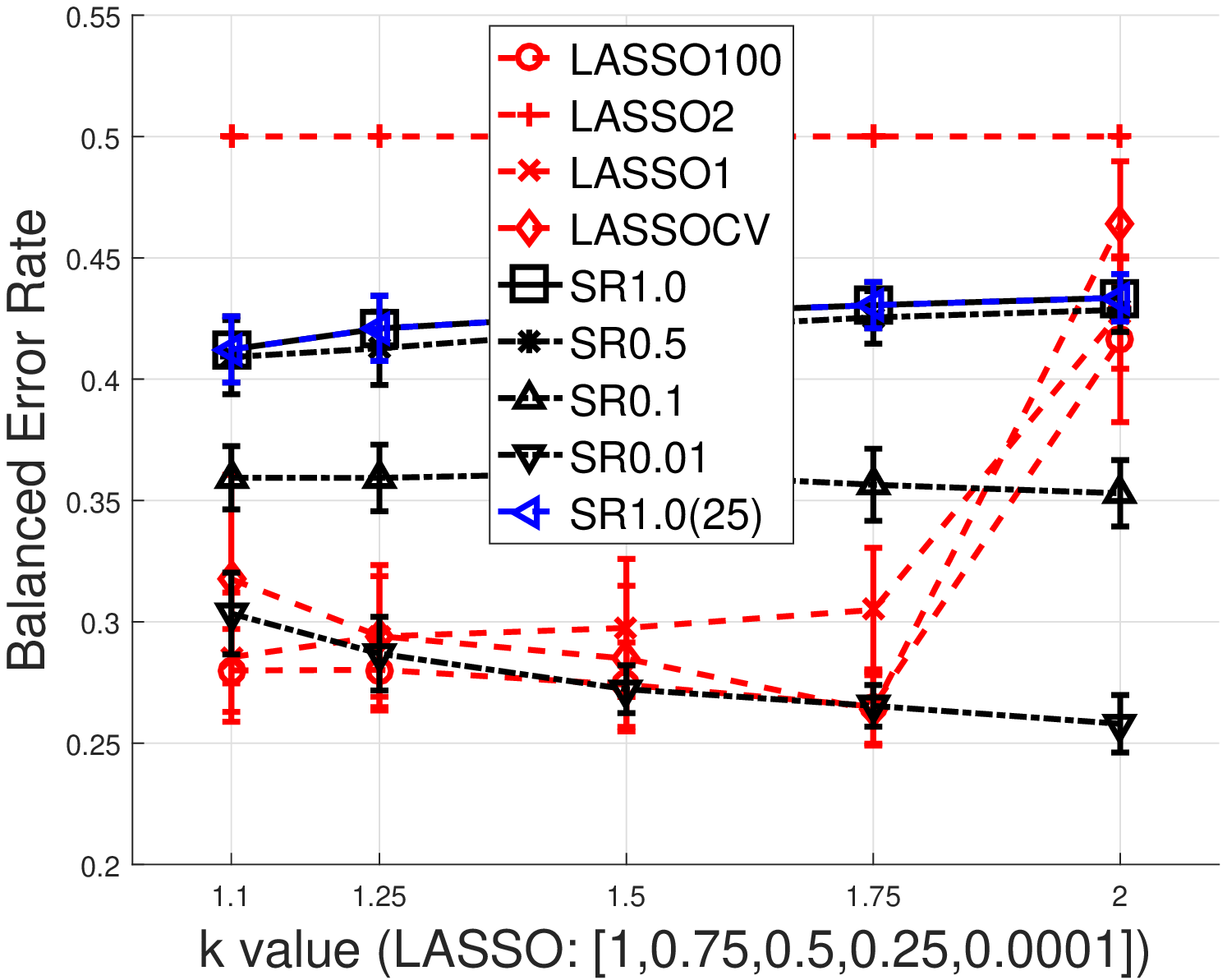} &
  \epsfxsize=6.8cm
  \epsfysize=3.9cm
  \hspace{-2mm}\epsffile[22     0   445   342]{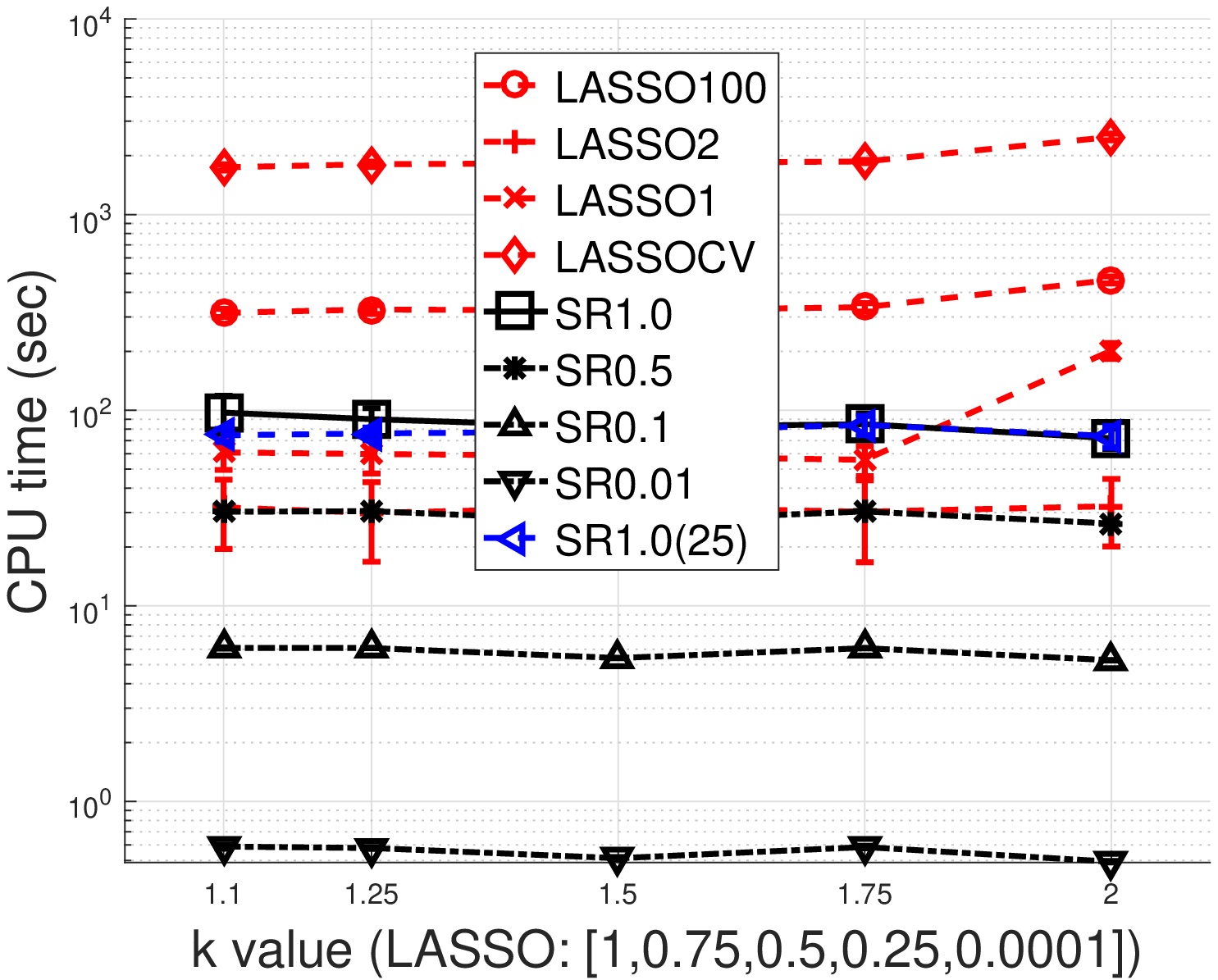}
  \\  \small{Dorothea: (e)} BER  & \small{(f) CPU} \\*[2mm]
  \epsfxsize=6.8cm
  \epsfysize=3.9cm
  \hspace{-2mm}\epsffile[19     0   445   341]{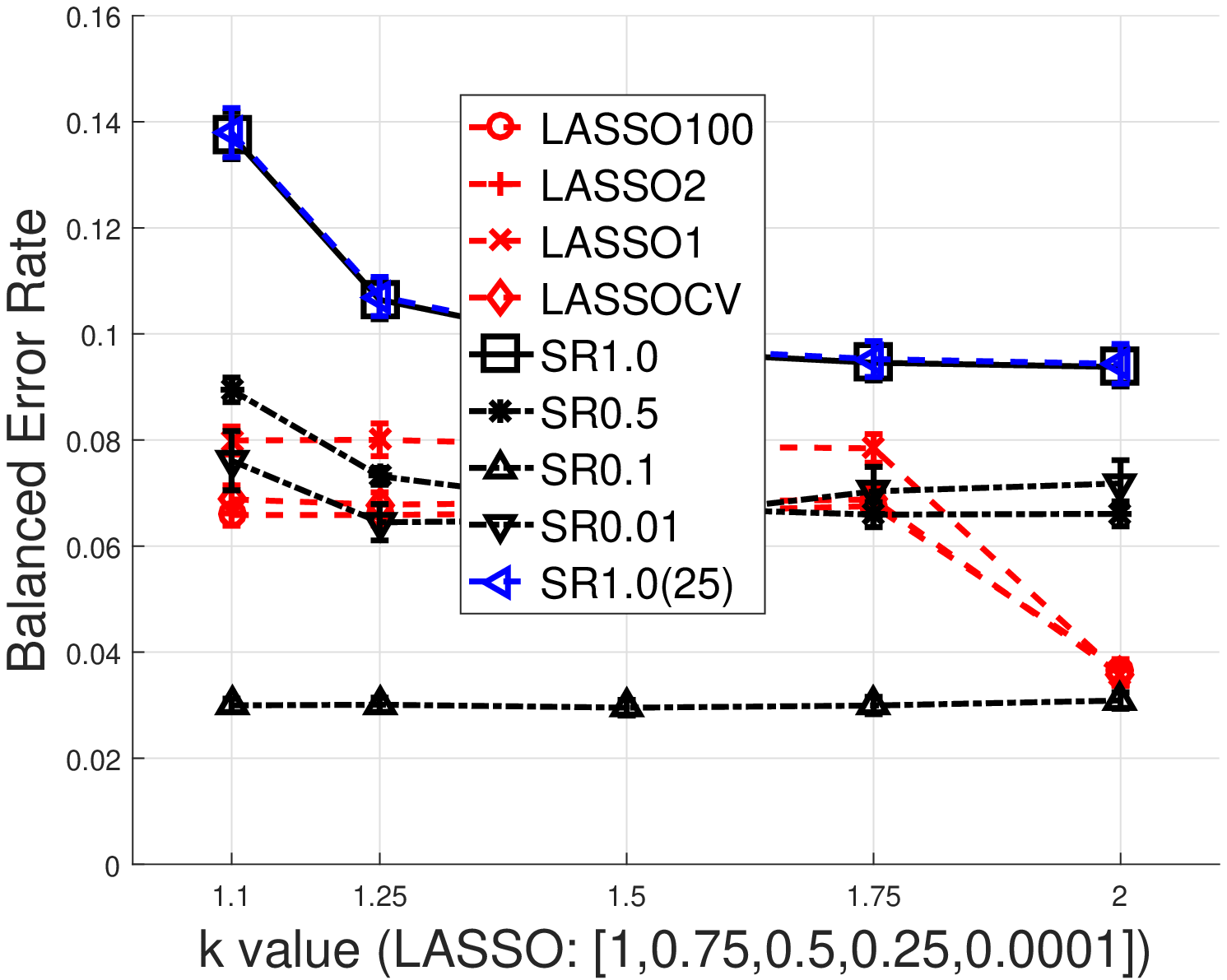} &
  \epsfxsize=6.8cm
  \epsfysize=3.9cm
  \hspace{-2mm}\epsffile[20     0   445   343]{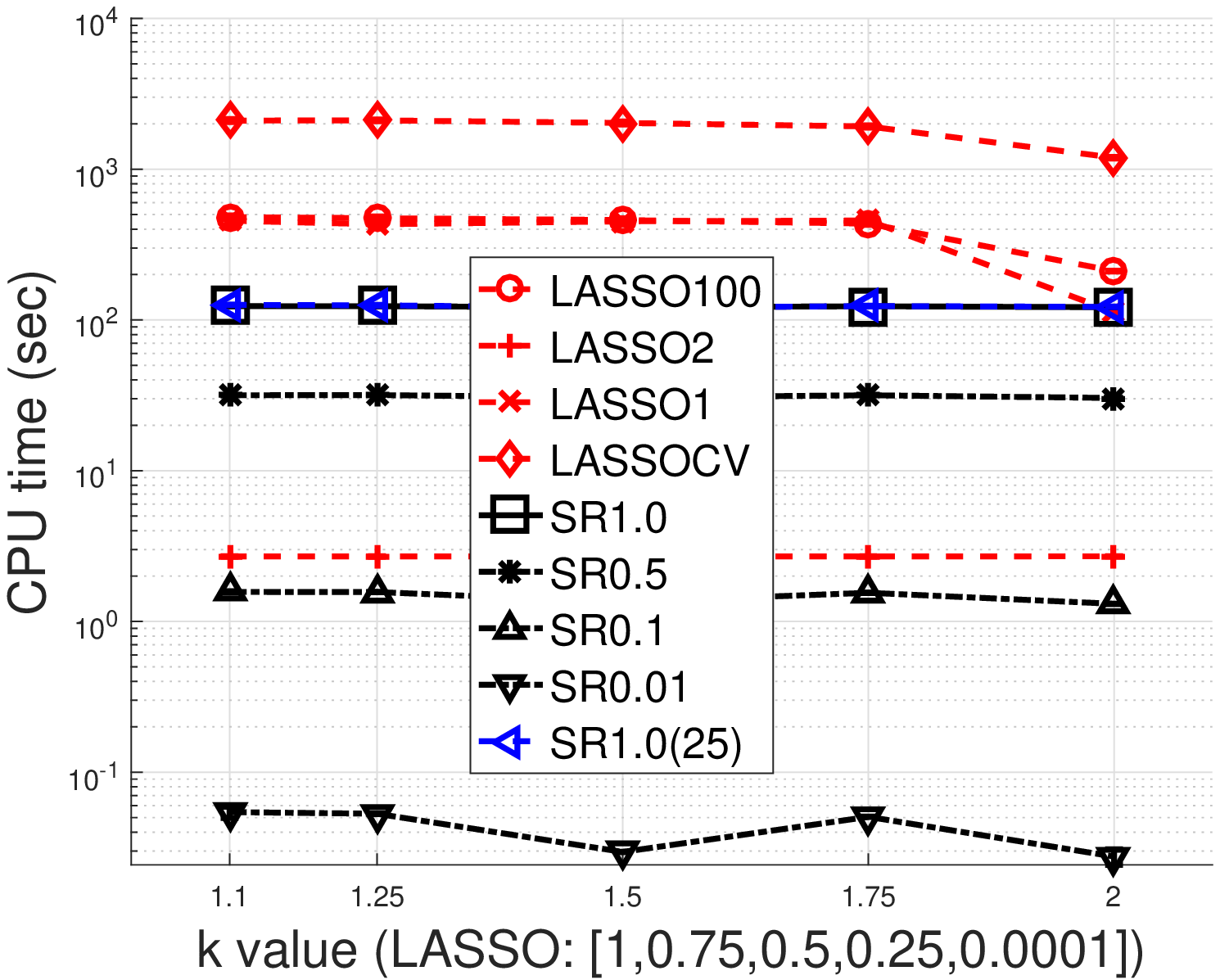}
  \\  \small{Gisette: (g) BER}  & \small{(h) CPU} \\*[2mm]
  \epsfxsize=6.8cm
  \epsfysize=3.9cm
  \hspace{-2mm}\epsffile[46     6   628   397]{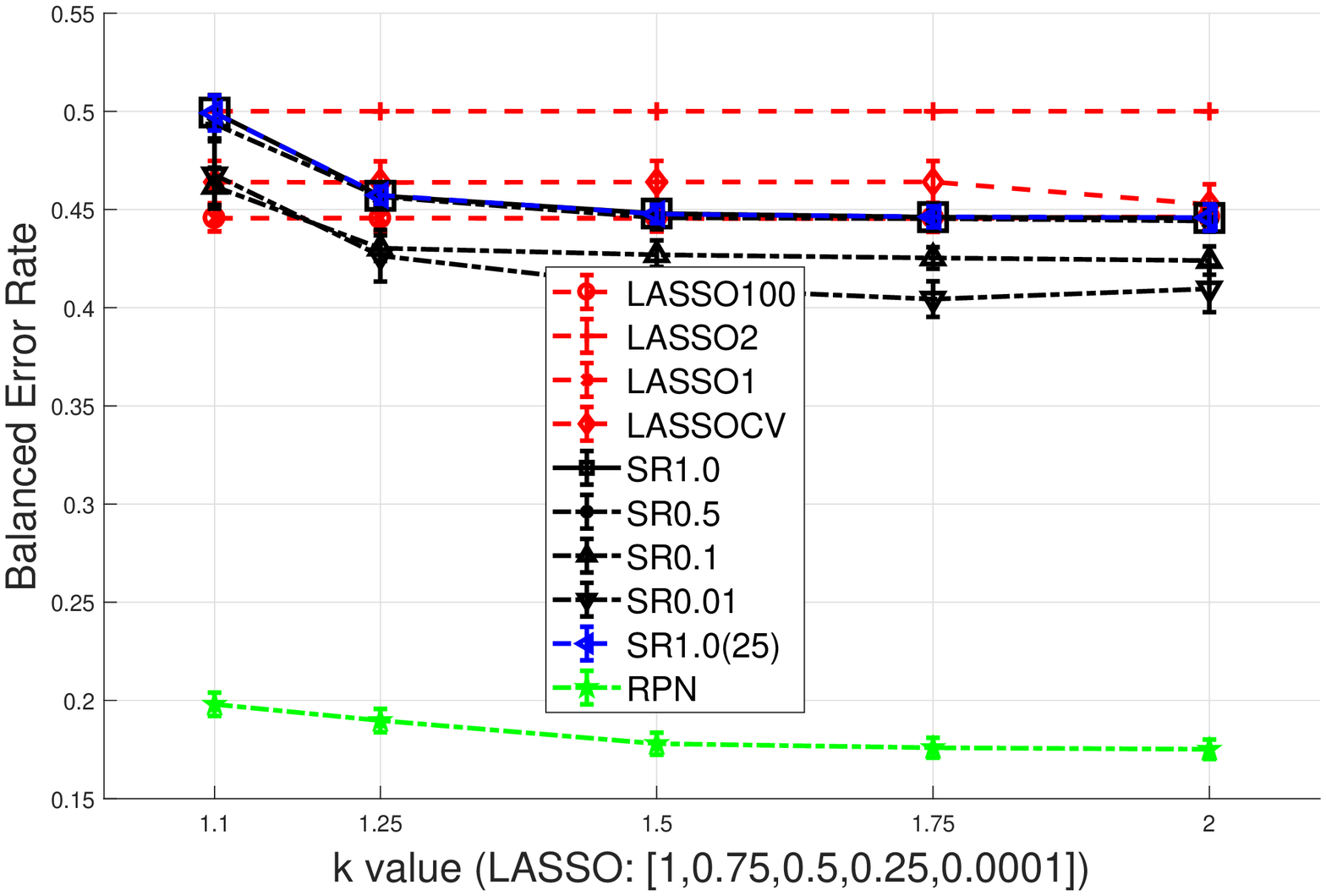} &
  \epsfxsize=6.8cm
  \epsfysize=3.9cm
  \hspace{-2mm}\epsffile[36     8   552   413]{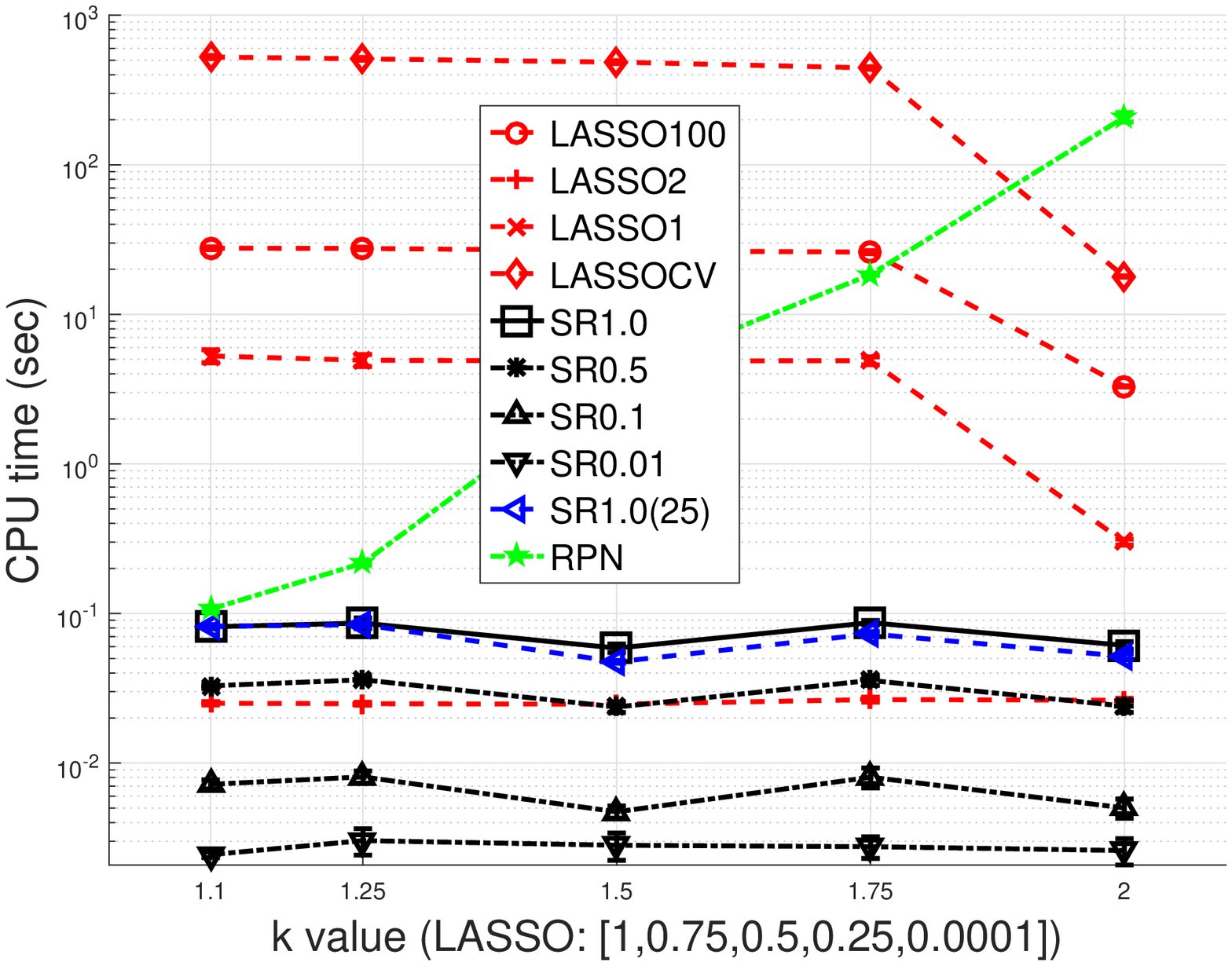}
  \\  \small{Madelon: (i) BER}  & \small{(j) CPU} \\*[2mm]
\end{tabular}
  \caption{Classification: BER and CPU plots for SR at $k\in\{1.1,1.25,1.5,1.75,2\}$ and LASSO
  at \texttt{Alpha}$\in\{1,0.75,0.5,0.25,0.0001\}$. The CPU for RPN is plotted over projection
  sizes 500, 1000, 5000, 10000 and 20000.}
  \label{fig_results_Acc_classification}
  \end{center}
\end{figure}

\textbf{(ii) Classification Problems: }
Fig.~\ref{fig_results_Acc_classification}(a),(c),(e),(g),(i) show the accuracy results in
terms of BER for the SR and the LASSO on the NIPS data sets, at ${c}=10^2$ and one using
\eqref{eqn_k_norm_solution} for similar reasons. The classification accuracy for the
Arcene data set in terms of BER is shown in Fig.~\ref{fig_results_Acc_classification}(a)
where the proposed SR is evaluated at various $k$ values ($k\in\{1.1,1.25,1.5,1.75,2\}$)
at different sparseness levels namely, SR1.0, SR0.5, SR0.1 and SR0.01. The results of
LASSO for various \texttt{Alpha} values (elastic net mixing value) at {four}
\texttt{NumLambda} settings (LASSO1, LASSO2, LASSO100 {and LASSOCV}) are plotted in the
same figure for comparison. In terms of BER, the proposed SR outperforms LASSO at various
feature densities (SR1.0, SR0.5 and SR0.1) except for SR0.01. The prediction accuracy
results for Dexter is shown in Fig.~\ref{fig_results_Acc_classification}(c) where the BER
trend is apparently similar to that of the Arcene example with the lowest performance
goes to SR0.01 and the higher performances go to SR1.0, SR0.5 and SR0.1. This shows at
least 10\% of the parameters are useful for estimation in SR learning for this case.
Different from the first two examples, in the Dorothea example of
Fig.~\ref{fig_results_Acc_classification}(e), the highly sparse SR (SR0.01) gives the
best BER performance compared with that of LASSO. For the Gisette data set in
Fig.~\ref{fig_results_Acc_classification}(g), SR0.1 achieves the best performance while
SR1.0 exhibits the worst performance among the compared classifiers.
Fig.~\ref{fig_results_Acc_classification}(i) show the BER performance for SR and LASSO at
various settings on the Madelon data set. Here, although some of the SR settings show
better {BER performance} compared to LASSO, the results are far from satisfactory (only
around {45\% BER}) due to noninformative features. To validate the need of a nonlinear
decision hyperplane on relevant features, an additional experiment is carried out on
correlated features selected according to the dot product between the input vector and
the target vector. An empirically selected eight features are then projected to high
dimensions (500, 1000, 5000, 10000 and 20000) using a random projection network (RPN)
\cite{Toh77}. The results show a significant improvement of {BER} for over 20\%. This
verifies the need of correlated features with nonlinear mapping capability beyond the
evaluated linear SR and LASSO for this data set.

\begin{figure}[hhh]
  \begin{center}
  \begin{tabular}{cc}
  \epsfxsize=5.8cm
  \hspace{-5mm}\epsffile[29     8   506   381]{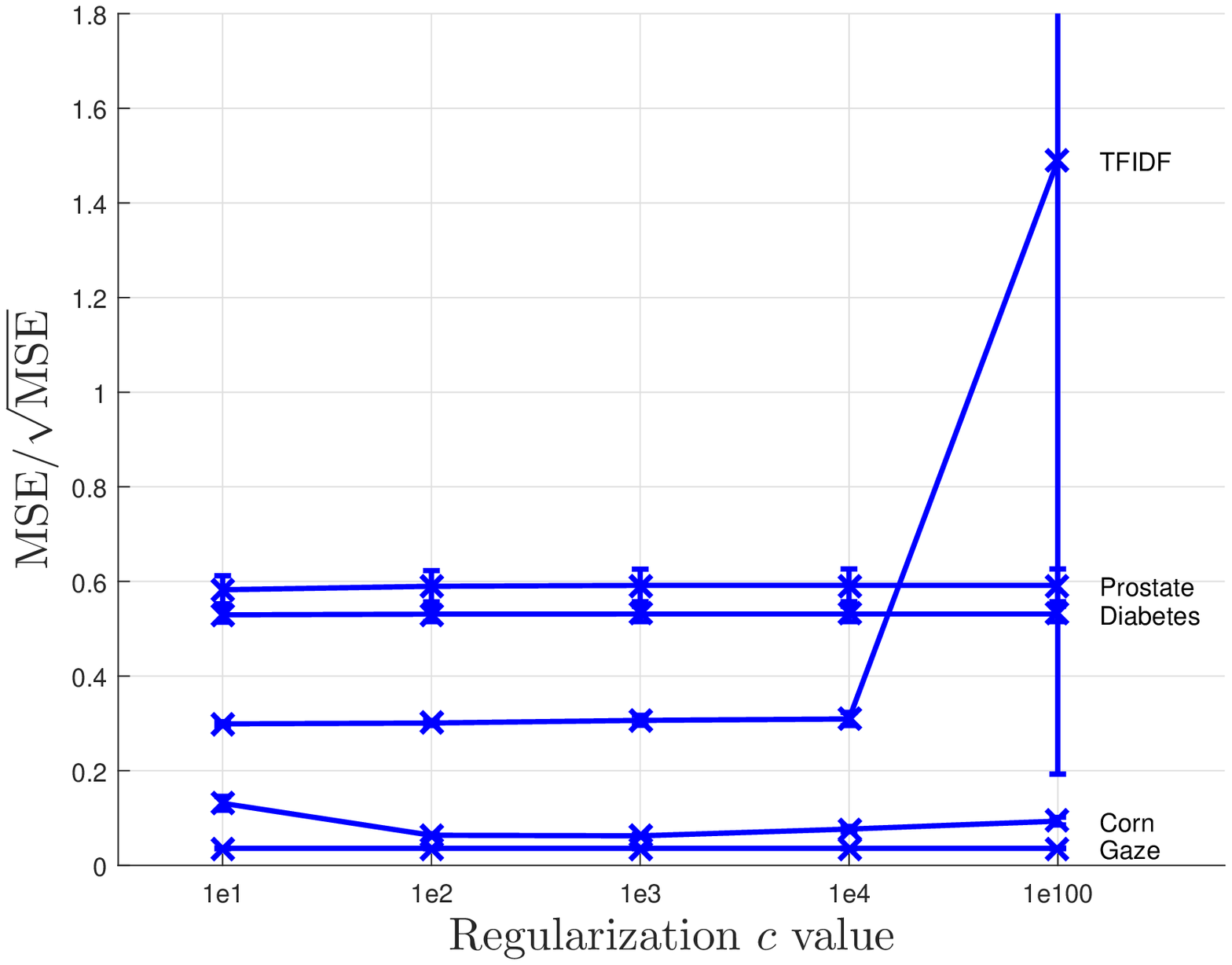} &
  \epsfxsize=5.8cm
  \hspace{8mm}\epsffile[10     2   383   295]{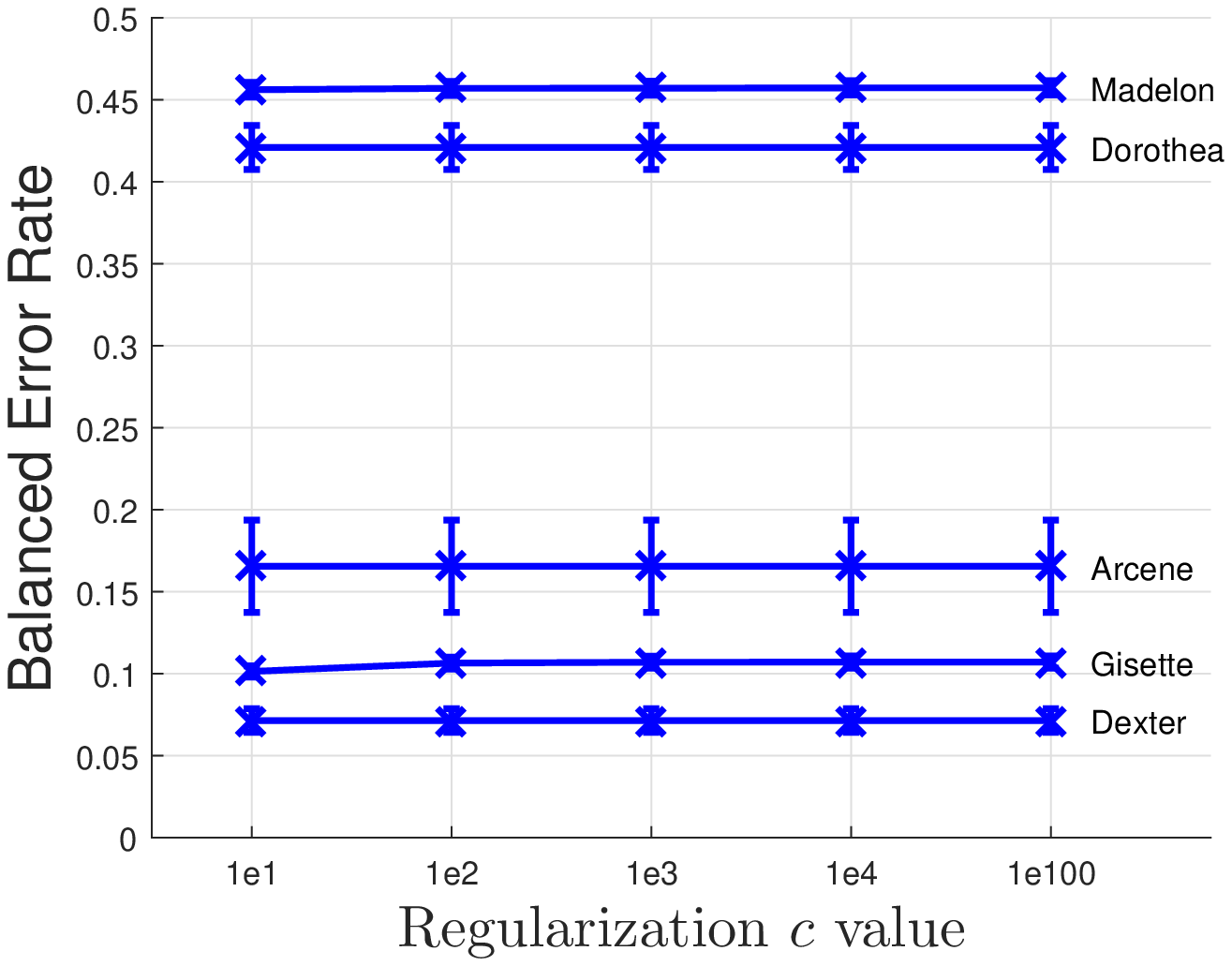}
  \\*[1mm]  (a) Regression data   & (b) Classification data
  \end{tabular}
  \caption{Error rates versus regularization ${c}$ values
  at ${k}=1.25$ and SR1.0: (a) MSE/$\sqrt{\rm MSE}$, (b) BER.}
  \label{fig_reg_study}
  \end{center}
\end{figure}

\textbf{(B) Effect of ${c}$ values}: Next, we observe the impact of regularization
settings by varying the ${c}$ values in $\{10,10^2,10^3,10^4,10^{100}\}$ at ${k}=1.25$.
Fig.~\ref{fig_reg_study} shows the plots of prediction accuracies for both the regression
and the binary classification problems. These results show that the estimations are
relatively insensitive to the variation of ${c}$ except for the TFIDF data set where a
significantly large variation of the error rate is observed at ${c}=10^{100}$.

\subsection{Processing Time}

\textbf{(i) Regression Problems: }
Fig.~\ref{fig_results_Acc_regression}(b),(d),(f),(h),(j) show the average training CPU
times in seconds for the SR and the LASSO on respective regression data sets. For the
Prostate data set, Fig.~\ref{fig_results_Acc_regression}(b) shows that SR is up to three
thousand times faster than LASSO in terms of training time. For the Diabetes data set,
Fig.~\ref{fig_results_Acc_regression}(d) shows that SR is up to nine thousand times
faster than LASSO. For the Corn data set, Fig.~\ref{fig_results_Acc_regression}(f) shows
no significant difference between SR1.0 and LASSO1 in execution time. However, the
execution time of LASSO2, LASSO100 and LASSOCV slows down significantly due to inclusion
of a penalty search in their implementation. For the TFIDF data set,
Fig.~\ref{fig_results_Acc_regression}(h) shows that the execution time of SR1.0 is
observed to be nearly twice slower than that of LASSO1. However, when a penalty search is
involved, the speed of LASSO slows down significantly. Finally, for the Gaze data set,
LASSO100 shows significantly longer search times compared with all other algorithmic
settings as seen from Fig.~\ref{fig_results_Acc_regression}(j). LASSOCV was not included
in this experiment due to its excessively long search time. In summary, we see that LASSO
depends much on the \texttt{NumLambda} values search process whereas the proposed SR
depends on the selected training feature dimension.

\textbf{(ii) Classification Problems:}
Fig.~\ref{fig_results_Acc_classification}(b),(d),(f),(h),(j) show the average training
CPU times in seconds for the SR and the LASSO on respective classification data sets. For
the Arcene data set, Fig.~\ref{fig_results_Acc_classification}(b) shows a comparable
speed between SR and LASSO1. However, when the \texttt{NumLambda} values are set beyond
1, the search process in LASSO significantly slows down its training speed. The training
CPU time in Fig.~\ref{fig_results_Acc_classification}(d) for the Dexter data set shows a
trend similar to that of the Arcene example. For the Dorothea data set, the training CPU
time as seen from Fig.~\ref{fig_results_Acc_classification}(f) for SR is comparable with
that of LASSO except for LASSO100 and LASSOCV which have the largest training CPU times.
This is due to the large operating matrix dimension of this data set. For the Gisette
data set, the training CPU times for all SR settings and LASSO1 in
Fig.~\ref{fig_results_Acc_classification}(h) are seen to be significantly lower than
those of LASSOCV, LASSO100 and LASSO2. Fig.~\ref{fig_results_Acc_classification}(j) shows
the processing times for the Madelon data set. The increasing trend of processing time
for the RPN is due to the increasing re-projected network sizes.

\subsection{Summary of Results and Discussion}

The experiments on regression problems show comparable MSE performance between SR and
LASSO. In terms of execution time, SR is on par with LASSO without penalty search.
However, when penalty search is included in LASSO (i.e., when \texttt{NumLambda}$>1$),
the efficiency of SR becomes obvious.

\begin{figure}[hhh]
  \begin{center}
\begin{tabular}{cc}
  \epsfxsize=5.8cm
  \epsfysize=5.3cm
  \epsffile[87    19   612   341]{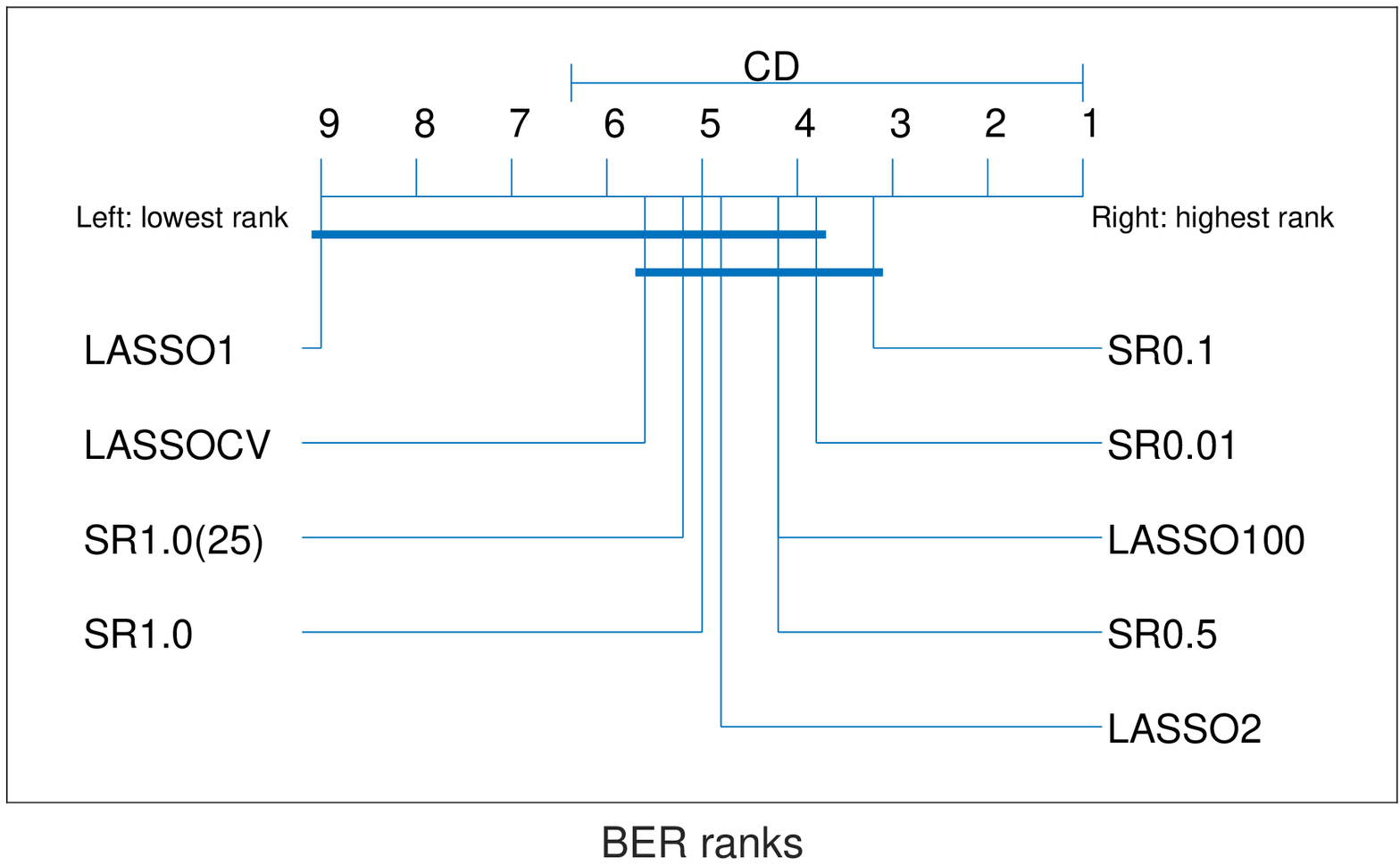} &
  \epsfxsize=5.8cm
  \epsfysize=5.3cm
  \hspace{2.8mm}\epsffile[87    19   610   340]{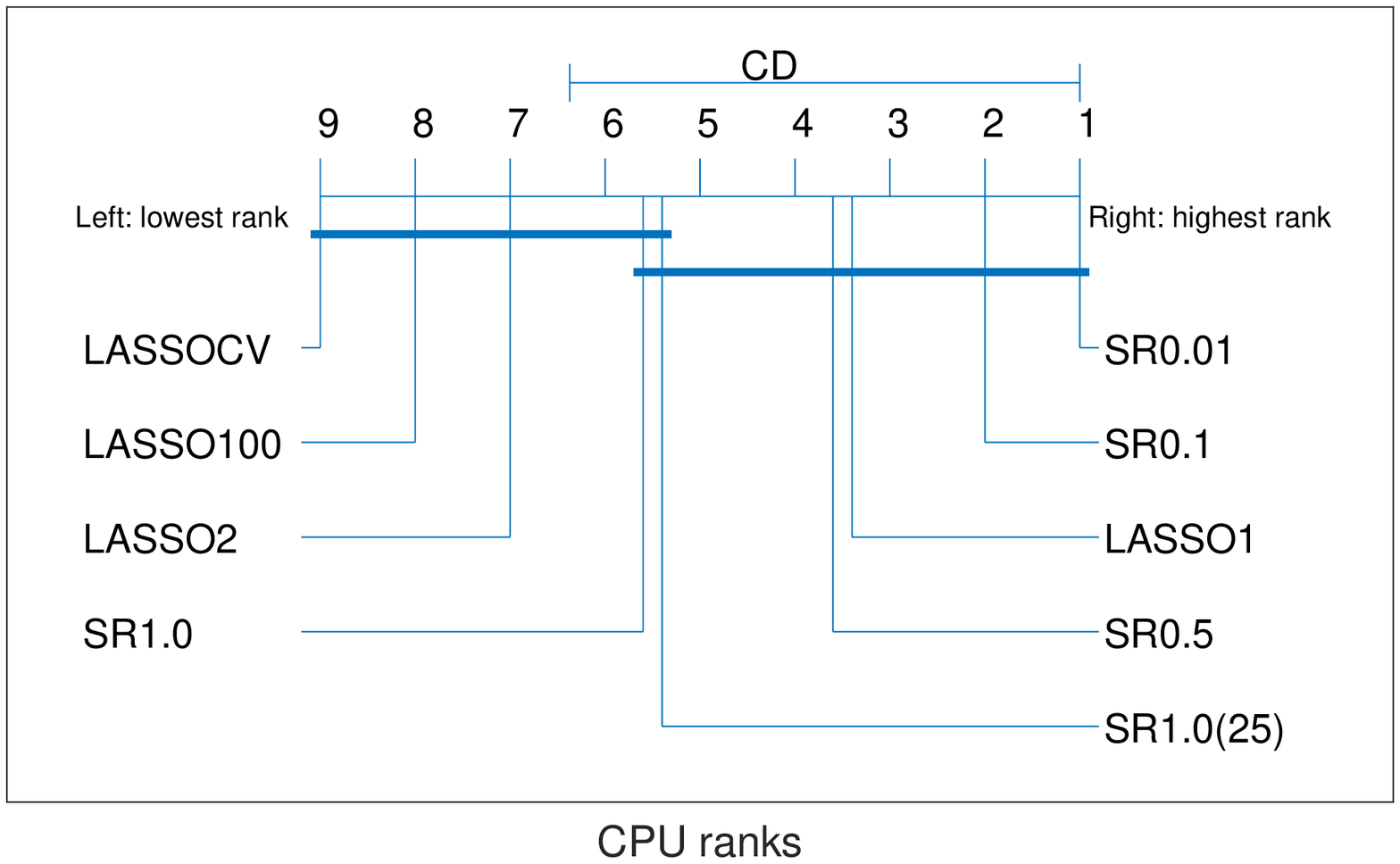}
  \\*[-1mm]  {\footnotesize (a) Nemenyi test for BER results}  &
    {\footnotesize (b) Nemenyi test for CPU results.}
\end{tabular}
  \caption{Nemenyi tests for binary classification problems.}
  \label{fig_Nemenyi_tests}
  \end{center}
\end{figure}

For the extension of regression to binary classification problems, because a complete set
of experiments for each algorithmic setting is available, we are able to perform a
Friedman test to see whether the comparisons in Fig.~\ref{fig_results_Acc_classification}
are significantly different in statistical sense. The confidence levels obtained for each
comparison metric are respectively, BER with $p=0.0595$ and CPU with $p=4.2142\times
10^{-6}$. This suggests that we cannot reject the null hypothesis that all compared
algorithms have similar performance in terms of BER where the observed differences are
merely random at $95\%$ confidence level. However, from the Nemenyi test \cite{Demsar1}
plots as shown in Fig.~\ref{fig_Nemenyi_tests}(a) and (b), we observe a significant
difference between the training CPU time of LASSO and SR when penalty search is involved
in LASSO (such as that in {LASSOCV,} LASSO100 and LASSO2). Here, consistent with the
Friedman test, the {BER} performance of penalty search based LASSOs {do not differ
significantly} from those of SR. Without inclusion of penalty
search, LASSO1 shows significantly lower BER performance %(ACC, AUC and BER)
than SR.

In summary, \emph{these experiments show the applicability of the proposed SR to
real-world problems}. Particularly the ${k}$ values are found to produce good accuracy at
${k}\geq 1.25$ and ${k}\geq 1.1$ respectively for regression and classification problems.
It is noted that the closer is the ${k}$ value to one, the higher is the compression.
However, due to the numerical stability of the term $1/(k-1)$ when $k\rightarrow 1$, the
accuracy can be affected. Hence, in practice, the ${k}$ value can be chosen based on the
lowest possible value near one with an accuracy performance comparable with that when
${k}=2$. The observed impact of regularization on the estimation error rate is not
significant when ${c}$ is chosen within the range $[10,10^4]$. While an in-depth study of
noise tolerance would be a valuable topic for future work, we observe here a
significantly faster training time of SR over that of LASSO with a penalty search.
\emph{Under relatively simple settings, the proposed SR shows predictive accuracy
comparable with that of LASSO for both regression (in terms of MSE) and binary
classification (in terms of BER) problems.}

\section{Conclusion} \label{sec_conclusion}

In this paper, a stretchy regression was proposed for learning of linear parametric
models beyond existing means. Essentially, a closed-form solution with feature warping
capability was derived in primal and dual forms, analogous to that of ridge regression.
Because the solution was operated upon positive and real input values, an exponential
transformation was proposed to convert the inputs to the positive real axes. Under the
space responsible for compressed estimation, it turned out that the solution enabled a
seamless combination of feature extraction and learning within a single framework. Our
experiments validated the feasibility of such regression for both synthetic and physical
data learning.

\section*{Acknowledgment}
We thank Zhengguo Li and Geok-Choo Tan for insightful discussions and Bernd Burgstaller
for proof reading the manuscript.

\bibliographystyle{IEEEtran2003}

\appendix

\section{Proof of Lemma~\ref{lemma_convex} on convexity of $k$-measure}
\begin{proof}
Based on the convexity of $f$ on each element $\alpha_i$, $i=0,1,...,D-1$ (of the
parameter vector $\balpha$), we have
\begin{equation}\label{eqn_convexity_of_f}
    f(\lambda\alpha_{i1}+(1-\lambda)\alpha_{i2})
        \leq \lambda f(\alpha_{i1}) + (1-\lambda) f(\alpha_{i2}),
        \ \ \ 0\leq\lambda\leq 1.
\end{equation}
Suppose $h(\theta):=\theta^{k}$ with $k,\theta>0$ where we know that $h$ is nondecreasing
and convex on $\theta$ since $dh/d\theta=k\theta^{k-1}\geq 0$ and
$d^2h/d\theta^2=k(k-1)\theta^{k-2}\geq 0$, $\forall k\geq 1$ (see also \cite{Hardy1}).
Using \eqref{eqn_convexity_of_f} plus the fact that $h$ is nondecreasing and convex, we
have for each $i=0,1,...,D-1$,
\begin{eqnarray}
    h(f(\lambda\alpha_{i1}+(1-\lambda)\alpha_{i2}))
        &\leq & h(\lambda f(\alpha_{i1}) + (1-\lambda) f(\alpha_{i2}))
        \nonumber \\
        &\leq & \lambda h(f(\alpha_{i1})) + (1-\lambda) h(f(\alpha_{i2})),
        \ \ \ 0\leq\lambda\leq 1 . \label{eqn_convexity_of_hf1}
\end{eqnarray}
Since summation of convex functions preserves the convexity, we have
\begin{eqnarray}
   \sum^{D-1}_{i=0} h(f(\lambda\alpha_{i1}+(1-\lambda)\alpha_{i2}))
        \leq \sum^{D-1}_{i=0} \lambda h(f(\alpha_{i1})) + (1-\lambda) h(f(\alpha_{i2})),
  \ \ 0\leq\lambda\leq 1,
\label{eqn_convexity_of_hf4}
\end{eqnarray}
which means convexity of $\sum^{D-1}_{i=0} h(\alpha_i) = \sum^{D-1}_{i=0} f(\alpha_i)^k =
\kmeask{\balpha}$ on $\balpha=[\alpha_0,\alpha_1,...,\alpha_{D-1}]^T$. Hence the proof.
\end{proof}

\section{Proof of Lemma~\ref{lemma3} on power of matrix-vector product factorization}
\begin{proof}
By direct multiplication, we have {\small
\begin{eqnarray}
  && \hspace{-1cm}{\bf A}(({\bf A}^T){\bf b})^{\elementwise{k}} = \nonumber\\
  &&  \left[ \begin{array}{cccc}
        a_{11} & a_{12} & \cdots & a_{1d} \\
        a_{21} & a_{22} & \cdots & a_{2d} \\
         \vdots & \vdots & \ddots & \vdots \\
        a_{m1} & a_{m2} & \cdots & a_{md} \\
        \end{array} \right]
        \left(
        \left[ \begin{array}{cccc}
        a_{11} & a_{21} & \cdots & a_{m1} \\
        a_{12} & a_{22} & \cdots & a_{m2} \\
        \vdots & \vdots & \ddots & \vdots \\
        a_{1d} & a_{2d} & \cdots & a_{md} \\
        \end{array} \right]
        \left[ \begin{array}{c}
        b_{1} \\
        b_{2} \\
        \vdots \\
        b_{m} \\
        \end{array} \right]
        \right)^{\elementwise{k}}  \nonumber\\
   &=&
   \left[ \begin{array}{cccc}
        a_{11} & a_{12} & \cdots & a_{1d} \\
        a_{21} & a_{22} & \cdots & a_{2d} \\
        \vdots & \vdots & \ddots & \vdots \\
        a_{m1} & a_{m2} & \cdots & a_{md} \\
        \end{array} \right]
        \left[ \begin{array}{c}
        (\sum^{m}_{i=1} a_{1i} b_{i} )^k \\
        (\sum^{m}_{i=1} a_{2i} b_{i} )^k \\
        \vdots \\
        (\sum^{m}_{i=1} a_{di} b_{i} )^k \\
        \end{array} \right]  \nonumber\\
   &=&  \left[ \begin{array}{c}
        \sum^{d}_{j=1}a_{1j} (\sum^{m}_{i=1} a_{ji} b_{i} )^k \\
        \sum^{d}_{j=1}a_{2j} (\sum^{m}_{i=1} a_{ji} b_{i} )^k \\
        \vdots \\
        \sum^{d}_{j=1}a_{mj} (\sum^{m}_{i=1} a_{ji} b_{i} )^k \\
        \end{array} \right]  \nonumber\\
   &=&  \left[ \begin{array}{c}
        \sum^{d}_{j=1}a_{1j}\sum^{m}_{i=1} a_{ji}^k b_{i}^k \\
        \sum^{d}_{j=1}a_{2j}\sum^{m}_{i=1} a_{ji}^k b_{i}^k \\
        \vdots \\
        \sum^{d}_{j=1}a_{mj}\sum^{m}_{i=1} a_{ji}^k b_{i}^k \\
        \end{array} \right]
        \circ \left[ \begin{array}{c}
        \frac{ \sum^{d}_{j=1}a_{1j} (\sum^{m}_{i=1} a_{ji} b_{i} )^k }{ \sum^{d}_{j=1}a_{1j}\sum^{m}_{i=1} a_{ji}^k b_{i}^k }\\
        \frac{ \sum^{d}_{j=1}a_{2j} (\sum^{m}_{i=1} a_{ji} b_{i} )^k }{ \sum^{d}_{j=1}a_{2j}\sum^{m}_{i=1} a_{ji}^k b_{i}^k } \\
        \vdots \\
        \frac{ \sum^{d}_{j=1}a_{mj} (\sum^{m}_{i=1} a_{ji} b_{i} )^k }{ \sum^{d}_{j=1}a_{mj}\sum^{m}_{i=1} a_{ji}^k b_{i}^k } \\
        \end{array} \right]  \nonumber\\
   &=& ({\bf A}(({\bf A}^T)^{\elementwise{k}} {\bf b}^{\elementwise{k}})) \circ \bm{s} \nonumber\\
   &=& ({\bf A}({\bf A}^T)^{\elementwise{k}} {\bf b}^{\elementwise{k}}) \circ \bm{s}, \textrm{by\
   associative\ law\
}\end{eqnarray} } \noindent for all $\sum^{d}_{j=1}a_{lj}\sum^{m}_{i=1} a_{ji}^k
b_{i}^k\neq 0$, $l=1,...,m$. Hence the result.
\end{proof}

\section{Derivations for Analysis of Variance} \label{app_AoV}

The analysis is based on ${\bf y} = {\bf P}\bm{\alpha} + \bm{e}$ where $\bm{e}$ is an
independently and identically distributed noise of zero mean with covariance matrix ${\bf
C}$.

For the under-determined system, the expectation of $\hat{\bm{ \alpha}}$ based on
\eqref{eqn_k_norm_solution_reg} is given by {\footnotesize
\begin{eqnarray*}
   E [\hat{\bm{ \alpha}}]
   &=&
     E \left [
    \left({\bf P}^T\right)^{\elementwise{\frac{1}{k-1}}}\left[{\bf P}
        ({\bf P}^T)^{\elementwise{\frac{1}{k-1}}} + \frac{1}{{c}k}{\bf I}\right]^{-1}
    ({\bf P}\bm{\alpha} + \bm{e}) \right ]\\
    &=&
     E \left [
    \left({\bf P}^T\right)^{\elementwise{\frac{1}{k-1}}}\left[{\bf P}
        ({\bf P}^T)^{\elementwise{\frac{1}{k-1}}} + \frac{1}{{c}k}{\bf I}\right]^{-1}
    ({\bf P}\bm{\alpha}) \right ] ,   \\
    &=&
          \left({\bf P}^T\right)^{\elementwise{\frac{1}{k-1}}}\left[{\bf P}
        ({\bf P}^T)^{\elementwise{\frac{1}{k-1}}} + \frac{1}{{c}k}{\bf I}\right]^{-1}
    {\bf P}\bm{\alpha} .
\end{eqnarray*} }
which is equal to \eqref{eqn_expect1}. The corresponding covariance matrix of $\hat{\bm{
\alpha}}$ is derived as {\footnotesize
\begin{eqnarray*} E [ (\hat{\bm{ \alpha}}-E [\hat{\bm{ \alpha}}])
(\hat{\bm{ \alpha}}-E [\hat{\bm{ \alpha}}])^T ]
  &=&
   E \left [  \left \{
    \left({\bf P}^T\right)^{\elementwise{\frac{1}{k-1}}}\left[{\bf P}
        ({\bf P}^T)^{\elementwise{\frac{1}{k-1}}} + \frac{1}{{c}k}{\bf I}\right]^{-1}
    \bm{e} \right \} \right .\\
   && \qquad \times
   \left .  \left\{
    \left({\bf P}^T\right)^{\elementwise{\frac{1}{k-1}}}\left[{\bf P}
        ({\bf P}^T)^{\elementwise{\frac{1}{k-1}}} + \frac{1}{{c}k}{\bf I}\right]^{-1}
    \bm{e} \right \}^T
    \right ]\\
   &=&
  \left \{
    \left({\bf P}^T\right)^{\elementwise{\frac{1}{k-1}}}\left[{\bf P}
        ({\bf P}^T)^{\elementwise{\frac{1}{k-1}}} + \frac{1}{{c}k}{\bf I}\right]^{-1}%\right \}
    \right \} {\bf C} \\
    &&   \qquad \qquad \times
     \left \{
    \left({\bf P}^T\right)^{\elementwise{\frac{1}{k-1}}}\left[{\bf P}
        ({\bf P}^T)^{\elementwise{\frac{1}{k-1}}} + \frac{1}{{c}k}{\bf I}\right]^{-1}
   \right \}^T .
\end{eqnarray*} }
which gives rise to \eqref{eqn_covar1}.

For the over-determined system, the expectation of $\hat{\bm{ \alpha}}$ is given by
\eqref{eqn_expect2} based on the following derivation: {\footnotesize
\begin{eqnarray*}
   E [\hat{\bm{ \alpha}}]
   &=&
     E \left [
    \left[({\bf P}^T)^{\elementwise{\frac{1}{k-1}}}{\bf P}
         + \frac{1}{{c}k}{\bf I}\right]^{-1}
        \left({\bf P}^T\right)^{\elementwise{\frac{1}{k-1}}}
    ({\bf P}\bm{\alpha} + \bm{e}) \right ]\\
    &=&
     E \left [
    \left[({\bf P}^T)^{\elementwise{\frac{1}{k-1}}}{\bf P}
         + \frac{1}{{c}k}{\bf I}\right]^{-1}
        \left({\bf P}^T\right)^{\elementwise{\frac{1}{k-1}}}
    ({\bf P}\bm{\alpha}) \right ].
\end{eqnarray*} }
The corresponding covariance matrix of $\hat{\bm{ \alpha}}$ is {\footnotesize
\begin{eqnarray*}
E [ (\hat{\balpha}-E [\hat{\balpha}]) (\hat{\bm{ \alpha}}-E [\hat{\bm{ \alpha}}])^T ]
  &=&
   E \left [  \left\{
    \left[({\bf P}^T)^{\elementwise{\frac{1}{k-1}}}{\bf P}
         + \frac{1}{{c}k}{\bf I}\right]^{-1}
        \left({\bf P}^T\right)^{\elementwise{\frac{1}{k-1}}}
    \bm{e} \right \} \right .\\
   && \qquad \times
   \left .  \left\{
\left[({\bf P}^T)^{\elementwise{\frac{1}{k-1}}}{\bf P}
         + \frac{1}{{c}k}{\bf I}\right]^{-1}
         \left({\bf P}^T\right)^{\elementwise{\frac{1}{k-1}}}
         \bm{e} \right \}^T
    \right ]
   \\ &=&
  \left\{
  \left[({\bf P}^T)^{\elementwise{\frac{1}{k-1}}}{\bf P}
         + \frac{1}{ck}{\bf I}\right]^{-1}
         \left({\bf P}^T\right)^{\elementwise{\frac{1}{k-1}}}
         \right \} {\bf C} \\
    &&   \qquad \qquad \times
     \left \{
     \left[({\bf P}^T)^{\elementwise{\frac{1}{k-1}}}{\bf P}
         + \frac{1}{ck}{\bf I}\right]^{-1}
         \left({\bf P}^T\right)^{\elementwise{\frac{1}{k-1}}}
       \right \}^T .
\end{eqnarray*} }
which gives rise to \eqref{eqn_covar2}.

\end{document}